\documentclass[sn-apa,iicol]{sn-jnl}

\usepackage{subcaption}
\usepackage{bm}

\hypersetup{hidelinks}
\usepackage{doi}

\renewcommand{\doi}[1]{\url{#1}}

\jyear{2021}%

\theoremstyle{thmstyleone}%
\theoremstyle{thmstyletwo}%

\theoremstyle{thmstylethree}%

\begin{document}

\title[NORPPA: NOvel Ringed seal re-identification by Pelage Pattern Aggregation]{NORPPA: NOvel Ringed seal re-identification by Pelage Pattern Aggregation}

\author*[1]{\fnm{Ekaterina} \sur{Nepovinnykh}}

\author[2]{\fnm{Ilia} \sur{Chelak}}

\author[1]{\fnm{Tuomas} \sur{Eerola}}

\author[1]{\fnm{Heikki} \sur{K\"{a}lvi\"{a}inen}}

\affil*[1]{\orgdiv{Computer Vision and Pattern Recognition Laboratory, Department of Computational Engineering, School of Engineering Science}, \orgname{Lappeenranta-Lahti University of Technology LUT}, \orgaddress{\street{P.O.Box 20}, \city{Lappeenranta}, \postcode{53851}, \country{Finland}}}

\affil[2]{\orgdiv{Department of Computer Science, Faculty of Science}, \orgname{University of Helsinki}, \orgaddress{\street{P.O. Box 4}, \city{Helsinki}, \postcode{00100}, \country{Finland}}}


\abstract{We propose a method for Saimaa ringed seal (\emph{Pusa hispida saimensis}) re-identification. Access to large image volumes through camera trapping and crowdsourcing provides novel possibilities for animal monitoring and conservation and calls for automatic methods for analysis, in particular, when re-identifying individual animals from the images. The proposed method NOvel Ringed seal re-identification by Pelage Pattern Aggregation (NORPPA) utilizes the permanent and unique pelage pattern of Saimaa ringed seals and content-based image retrieval techniques. First, the query image is preprocessed, and each seal instance is segmented. Next, the seal’s pelage pattern is extracted using a U-net encoder-decoder based method. Then, CNN-based affine invariant features are embedded and aggregated into Fisher Vectors. Finally, the cosine distance between the Fisher Vectors is used to find the best match from a database of known individuals. We perform extensive experiments of various modifications of the method on a new challenging Saimaa ringed seals re-identification dataset. The proposed method is shown to produce the best re-identification accuracy on our dataset in comparisons with alternative approaches.}


\maketitle

\section{Introduction}

Animal biometrics, especially image-based individual re-identification, has recently gained extensive attention due to the availability of large volumes of wildlife image data gathered via automatic game cameras and citizen science projects. The benefits of automated re-identification methods are evident as they allow valuable data for conservation efforts to be obtained, for example, accurate population size estimates and novel information about animal migration and behavior patterns~\citep{mccoy2018long,araujo2020getting}. Compared to traditional methods such as tagging, which may cause stress and change the behavior of the animal, image-based re-identification offers a non-invasive technique for monitoring of endangered species~\citep{norouzzadeh2018automatically}. 

The Saimaa ringed seal (\emph{Pusa hispida saimensis}) is an endangered species native to Lake Saimaa, Finland. Seals of this species have a distinct ring-like pelage pattern, which is both permanent and unique for each individual, providing a basis for re-identification. An ongoing conservation effort~\citep{koivuniemi2016photo, koivuniemi2019mark, kunnasranta2021sealed} uses image-based re-identification to study animal migration and behavior. Currently, however, this re-identification work is carried out manually, which in view of the large number of images is very labour intensive and time consuming. Automated computer vision-based re-identification would clearly be of great benefit when carrying out this task.

A variety of methods for animal re-identification exist that utilize distinct characteristics in fur, feather and skin patterns~\citep{hotspotter, berger2015ibeis, moskvyak2019robust, li2019amur}, and methods originally developed for human face re-identification have been successfully applied to animals~\citep{deb2018face, crouse2017lemur, agarwal2019triplet}. Visual animal re-identification can be formulated as a task of finding a match for the given query image from a database of known individuals, which is equivalent to a content-based image retrieval (CBIR) problem~\citep{smeulders2000content} where an image is searched from a database based on the image content. However, despite the clear similarity between CBIR and re-identification tasks, utilizing utilization of CBIR approaches for animal re-identification has remained largely unstudied.

Saimaa ringed seals introduce additional challenges to the re-identification that make the task more difficult compared to many other animals for which re-identification has already been successfully applied. First, the image data is extremely biased. The majority of images are collected using static game cameras producing images with the same viewing angle and background and a limited set of possible seal locations and poses in the frame. At the same time, the high site fidelity (a tendency to return to previously visited locations) and low sociality of Saimaa ringed seals often result in a large portion of images of one individual seal being captured by only one game camera. Machine learning models trained on this kind of data tend to learn features that do not generalize to new datasets (e.g., data from a different year with different game camera locations). Moreover, as only a small portion of Saimaa ringed seal habitat can be covered with game cameras, datasets for seal identification are usually complemented with DSLR camera images, as well as images obtained via citizen science projects (e.g., mobile phone camera pictures). This image heterogeneity introduces a domain shift and due to the fact that different individuals are often captured with different cameras, it also contributes to the database bias problem. Finally, re-identifying Saimaa ringed seals from images is very challenging per se because of: (i) the large variation in possible poses, which is further exacerbated by the deformable nature of the seals, (ii) the non-uniform pelage patterns, limiting the size of the regions that can be used for the re-identification task, and (iii) the low contrast between the ring pattern and the rest of the pelage, as well as the varying appearance (e.g., wet and dry fur). Re-identification of Saimaa ringed seals is therefore considerably more difficult than, for example, zebra re-identification, where there are clearly visible patterns and limited variation in the pose of the torso.

\begin{figure*}[ht!]
	\centering
	\includegraphics[width=0.9\linewidth]{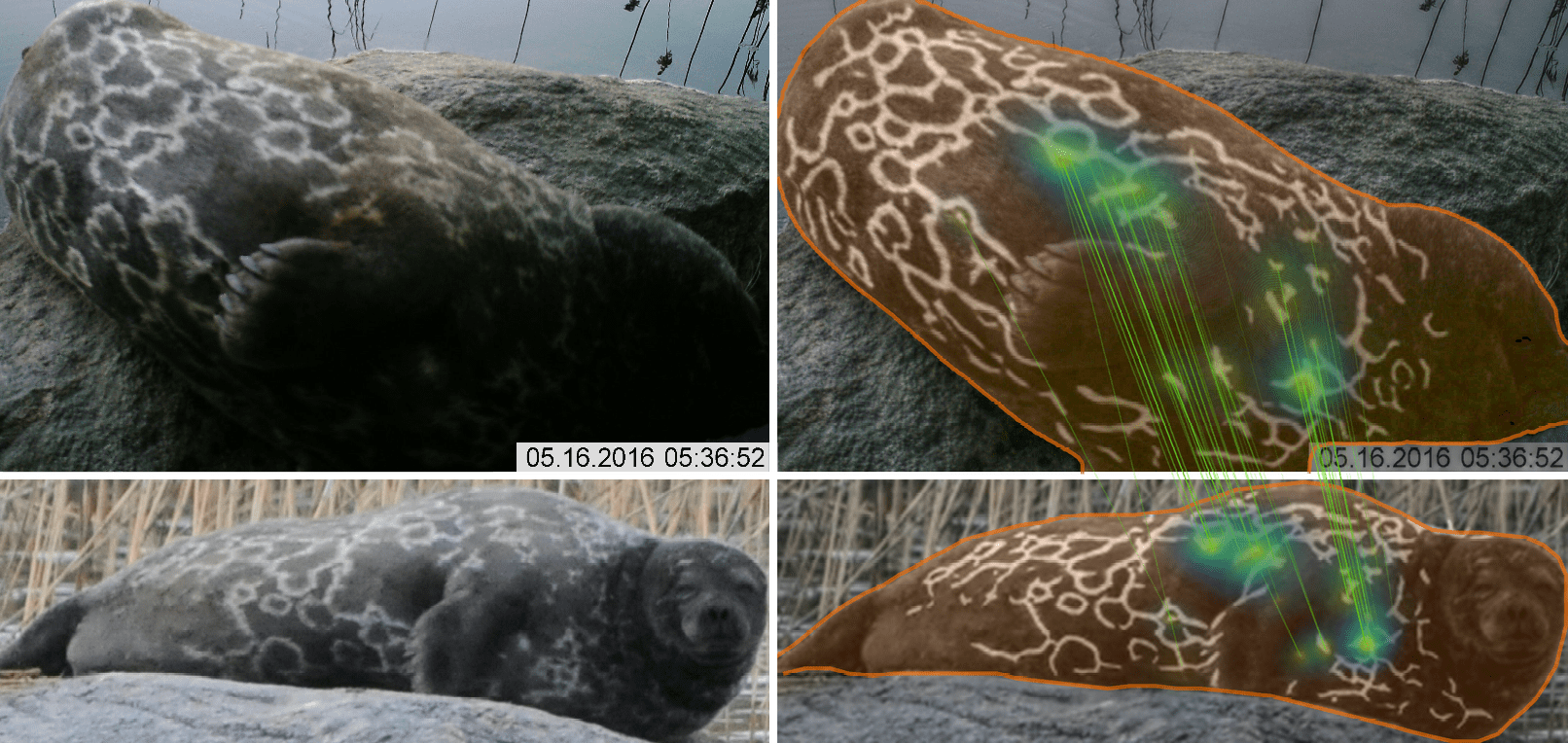}
	\caption{Visualisation of the proposed re-identification method. Input pictures are on the left, the results are on the right. The seal is segmented (orange outline), and matching regions of the pelage pattern are highlighted and connected with lines. The intensity of the highlights corresponds to the similarity of the matched regions.}
	\label{fig:method}
\end{figure*}

In this paper, we address the above challenges by proposing the NOvel Ringed seal re-identification by Pelage Pattern Aggregation (NORPPA) method for automatic Saimaa ringed seal re-identification (Fig.~\ref{fig:method}). The method is inspired by CBIR methods and builds on earlier work~\citep{nepovinnykh2020siamese} where Siamese networks were utilized to learn a similarity metric for local patches of pelage patterns. We further develop this approach by proposing an improved pattern feature embedding, which is done by utilizing affine invariant local CNN features and aggregating them into a fixed size embedding vector describing global features. The input image is first preprocessed using tone mapping and then segmented to detect and separate the seals from the background. The pelage pattern is further extracted using a U-net encoder-decoder~\citep{ronneberger2015u} based method. Affine invariant features are extracted and aggregated into a descriptor. Finally, the re-identification is performed by finding a descriptor with the minimum distance from the database of known individuals.

In the experimental part of the work, we show that the proposed method outperforms previously developed re-identification methods for Saimaa ringed seals as well as HotSpotter~\citep{hotspotter}, a popular species agnostic pattern-based re-identification approach, on the challenging task of Saimaa ringed seal re-identification. In addition, different variations of the method are comprehensively evaluated to find the best pattern feature embeddings for the task. The main contribution of this paper can be summarized as follows: 
(i) a novel Saimaa ringed seal re-identification method (NORPPA) inspired by content based image retrieval methods, 
(ii) a novel combination of local affine-covariant region learning and CNN-based descriptors and feature aggregation to obtain a single fixed size pattern embedding vector with high discrimination power, and 
(iii) extensive evaluation of the method and its modifications on a challenging Saimaa ringed seal dataset. 
While the method was developed for Saimaa ringed seals, it is also possible to apply it to other patterned species as shown by~\citet{badreldeen2021metric}.

\section{Related work}

\subsection{Animal re-identification}	

Animal re-identification is a broad term referring to the process of identifying an individual animal based on its features. The features are based on biological traits, and they can be captured in a number of ways, for example, acoustically~\citep{hartwig2005individual,pruchova2017cues} or visually in the form of images~\citep{vidal2021perspectives} or videos~\citep{freytag2016Chimpanzee}. Currently, image-based approaches are the most widely utilized approach due to the relative ease of data acquisition and manual analysis~\citep{schneider2019past}.

\begin{figure*}[ht]
	\centering
	\subfloat[][]
	{
		\includegraphics[width=0.23\linewidth]{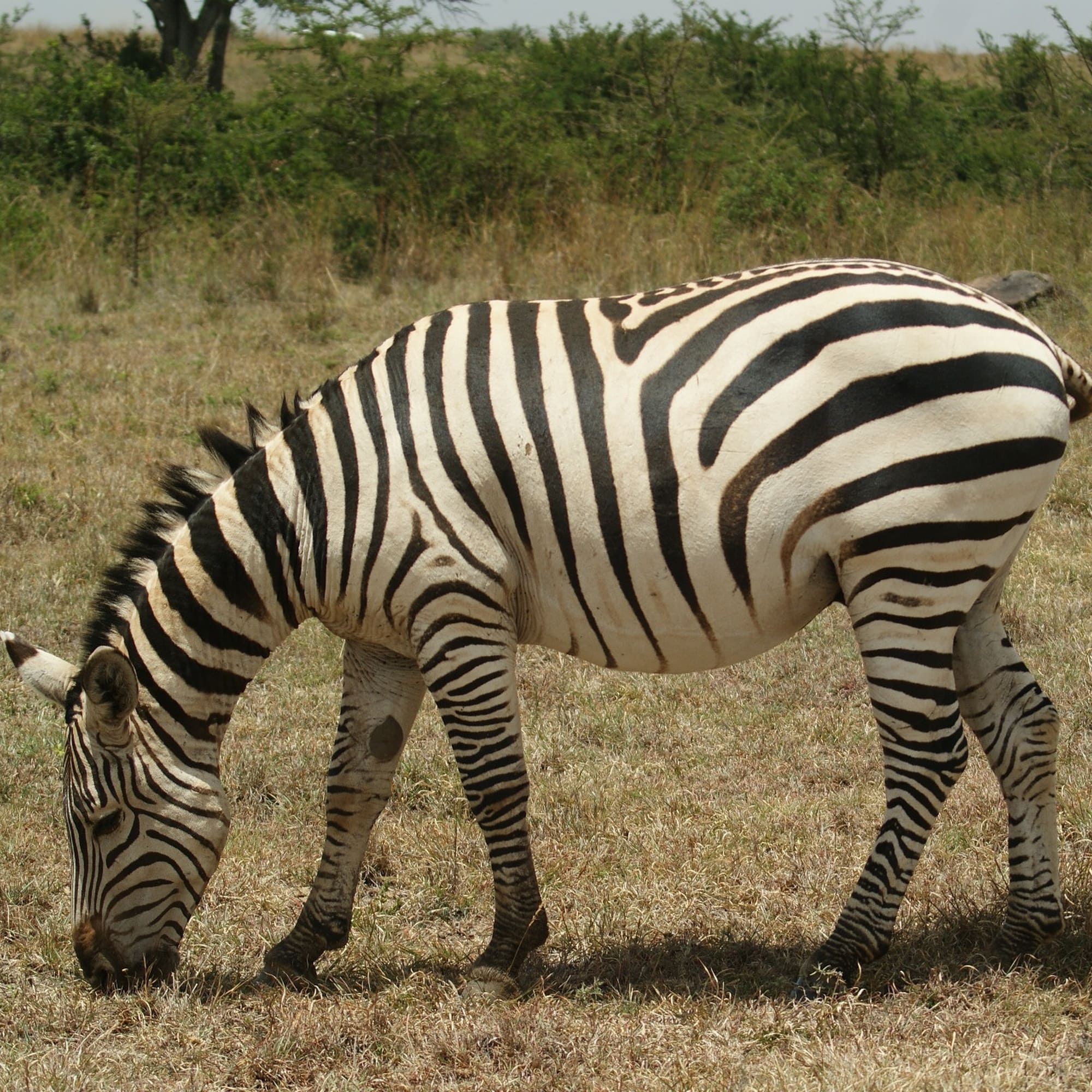}
		\label{subfig:zebra}
	}
	\subfloat[][]
	{
		\includegraphics[width=0.23\linewidth]{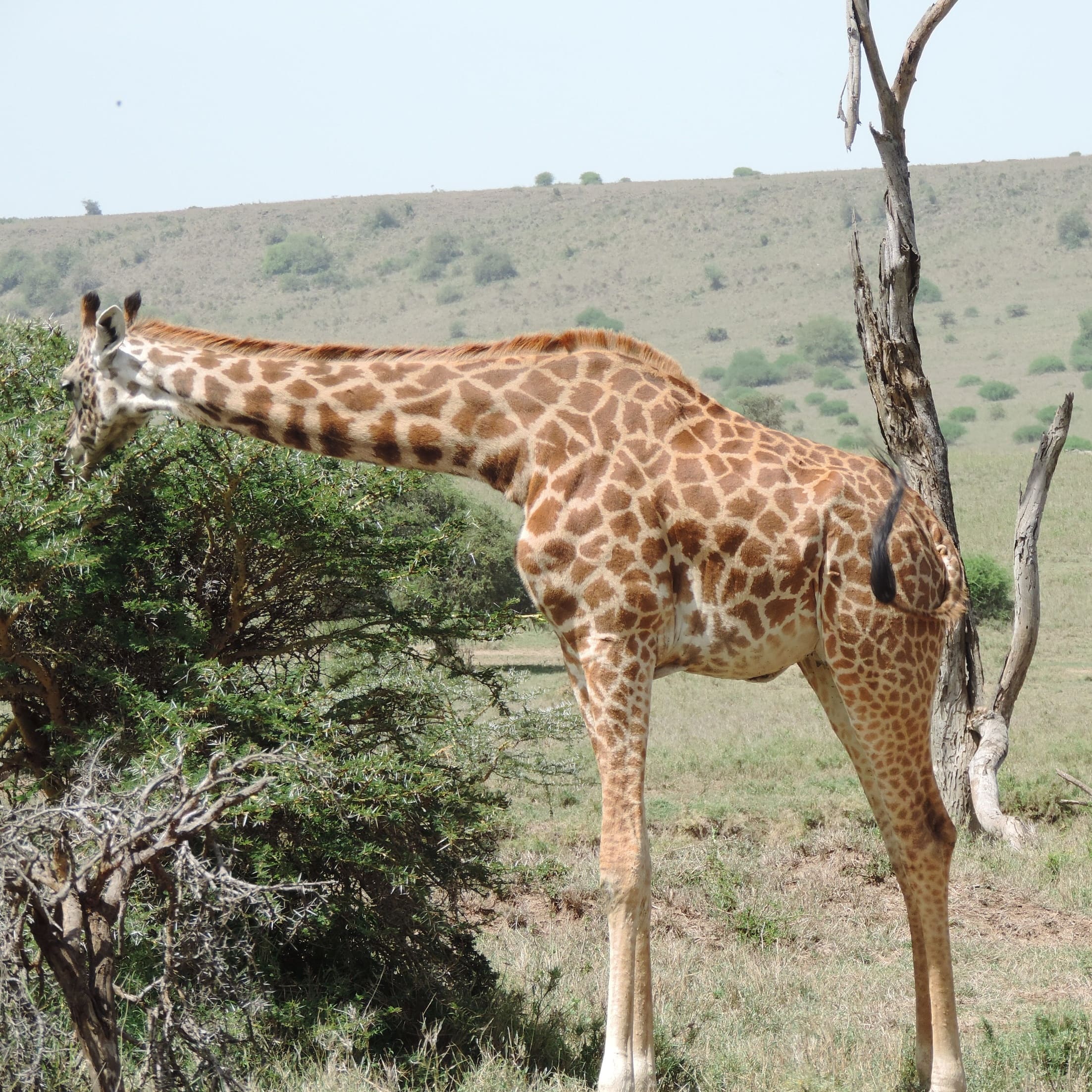}
		\label{subfig:giraffe}
	}
	\subfloat[][]
	{
		\includegraphics[width=0.23\linewidth]{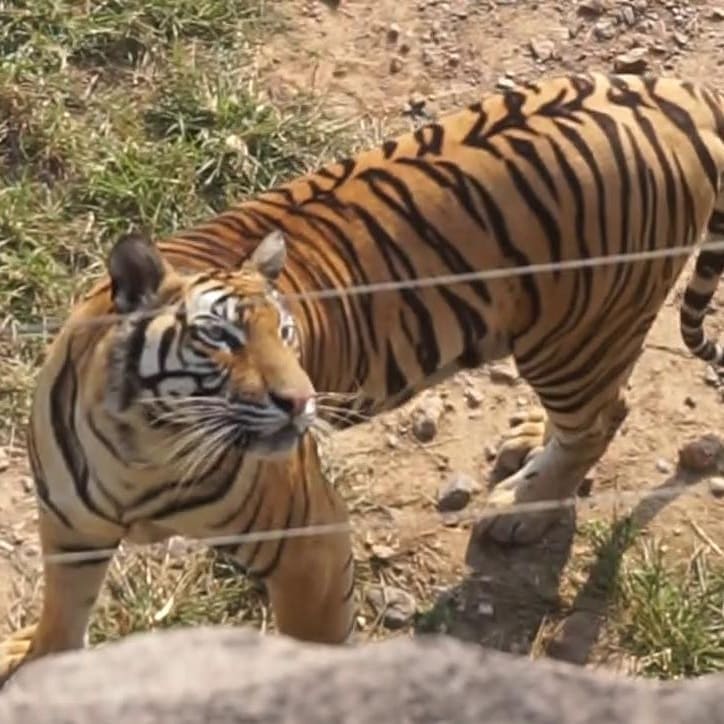}
		\label{subfig:tiger}
	}
	\subfloat[][]
	{
		\includegraphics[width=0.23\linewidth]{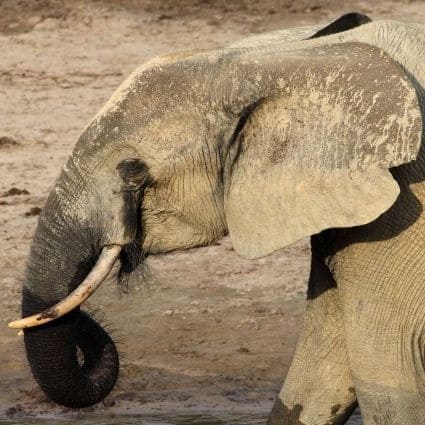}
		\label{subfig:elephant}
	}\\
	\subfloat[][]
	{
		\includegraphics[width=0.23\linewidth]{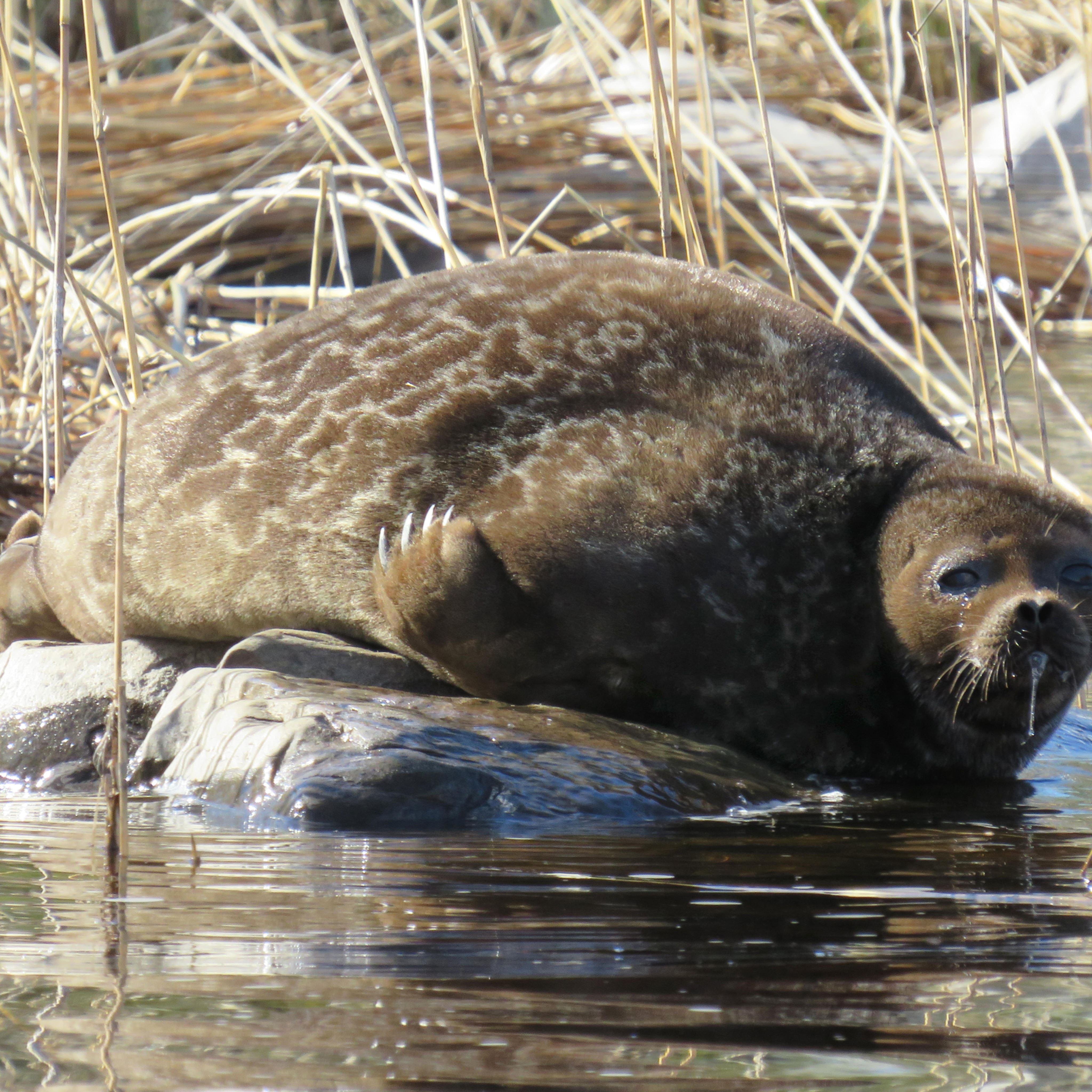}
		\label{subfig:norppa}
	}
	\subfloat[][]
	{
		\includegraphics[width=0.23\linewidth]{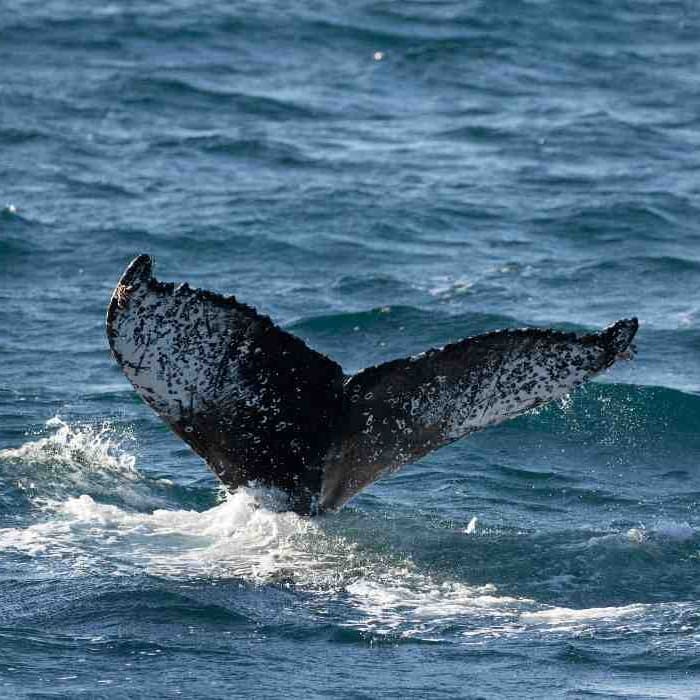}
		\label{subfig:humpbackwhale}
	}
	\subfloat[][]
	{
		\includegraphics[width=0.23\linewidth]{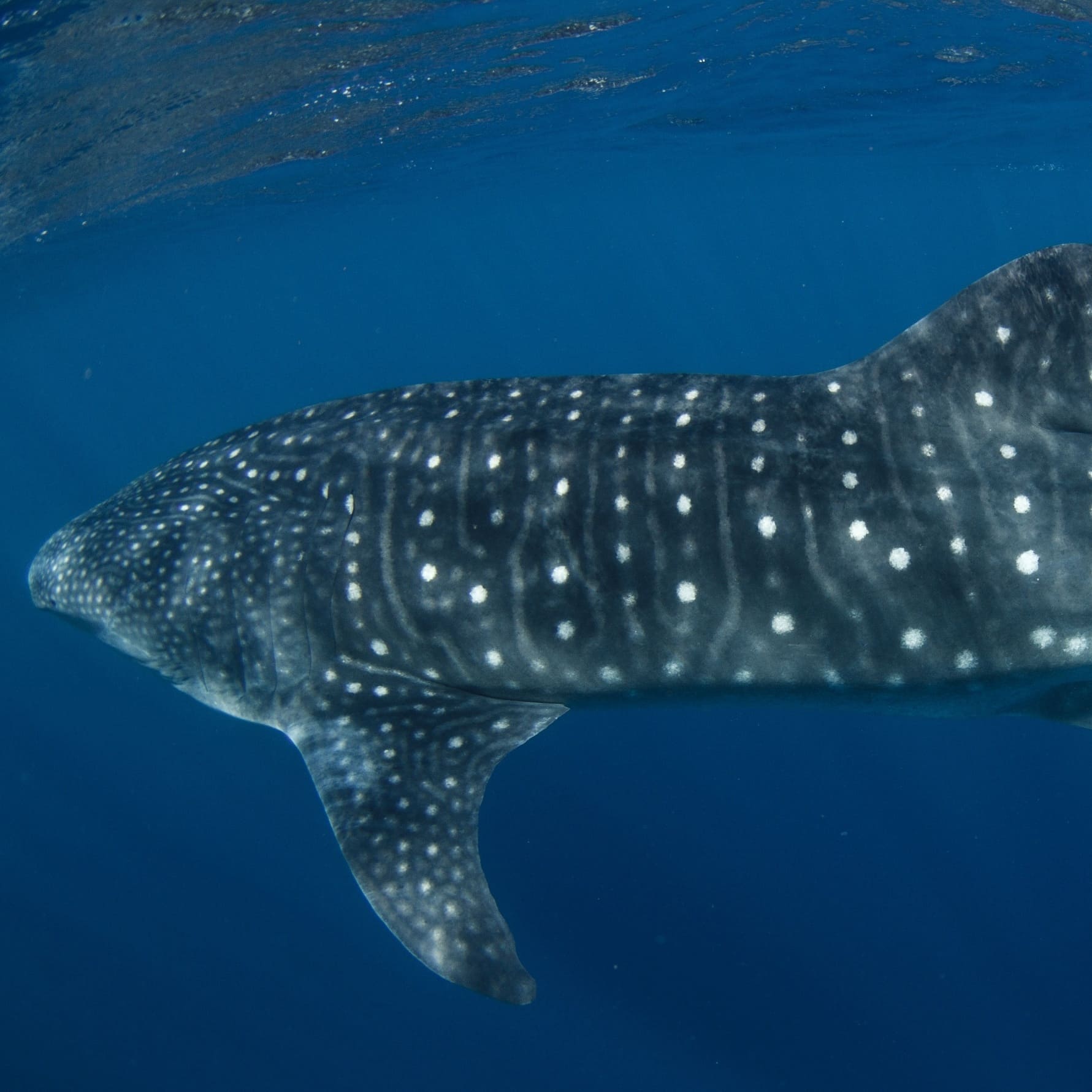}
		\label{subfig:whaleshark}
	}
	\subfloat[][]
	{
		\includegraphics[width=0.23\linewidth]{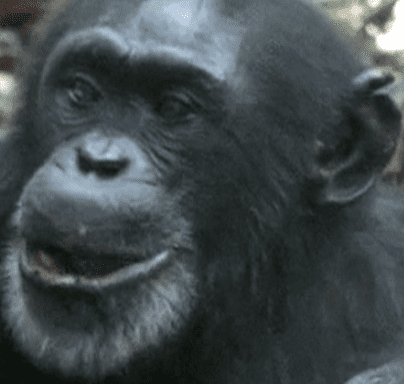}
		\label{subfig:chimpface}
	}
	
	\caption[]{Example images of the main identifiable features from publicly available re-identification data sets: (\subref{subfig:zebra}) Plains zebra (\emph{Equus quagga})~\citep{parham2017animal}: stripe fur pattern; (\subref{subfig:giraffe}) Masai giraffe (\emph{Giraffa tippelskirchi})~\citep{parham2017animal}: spot fur pattern; (\subref{subfig:tiger}) Amur tiger (\emph{Panthera tigris})~\citep{li2019amur}: stripe fur pattern;
	(\subref{subfig:elephant}) African elephant (\emph{Loxodonta africana})~\citep{korschens2019elpephants}: head shape;  (\subref{subfig:norppa}) Saimaa Ringed seal (\emph{Pusa hispida saimensis})~\citep{dataset}: ringed fur pattern; (\subref{subfig:humpbackwhale}) Humpback whale (\emph{Megaptera novaeangliae})~\citep{cheeseman2017happywhale}: fluke shape;  (\subref{subfig:whaleshark}) Whale shark (\emph{Rhincodon typus})~\citep{holmberg2009estimating}: skin spot pattern;
	(\subref{subfig:chimpface}) Chimpanzee (\emph{Pan troglodytes})~\citep{freytag2016Chimpanzee}: face}
	\label{fig:visual_features}
\end{figure*}

Various animal species can be re-identified by different types of visually unique biological traits such as fur pattern, face or fin shape. Examples of such traits are presented in Fig.~\ref{fig:visual_features}. Algorithmically, the methods can be divided into classification and metric-based approaches~\citep{vidal2021perspectives}. Classification-based approaches assume that the database of known individuals is known and finite, and the final algorithm can only identify individuals from that database. Metric-based methods, on the other hand, aim to learn a similarity metric between the input images. The re-identification is then performed by clustering or matching based on the similarity, which means that metric-based approaches are not limited by the initial database and can be applied to new individuals without retraining. Re-identification algorithms also differ in the feature extraction approaches used, which can be manual or semi-manual, where user input is required to extract salient regions, or automatic, where input images are fully processed by the method. Fully automatic methods are of most interest as they would allow efficient analysis of large data volumes.

One of the largest wildlife re-identification projects, Wildbook~\citep{berger2017wildbook} uses different kinds of algorithms for edge-based or pattern-based re-identification. The most efficient algorithms for deep learning edge-based re-identification are CurvRank~\citep{weideman2017integral}, finFindR~\citep{thompson2019finfindr}, and OC/WDTW~\citep{bogucki2019applying}, which have been applied to marine mammals such as bottlenose dolphins (\emph{Tursiops truncatus}), humpback whales (\emph{Megaptera
novaeangliae}), right whales (\emph{Eubalaena glacialis}), and use the unique shape of tail or fins to identify the animals.

Wildbook uses PIE and Hotspotter metric-based algorithms to re-identify animals by pattern. 
PIE~\citep{moskvyak2019robust} is a deep learning-based method for matching of individuals invariantly to the pose. It receives shape embedding and pose embedding separately and normalizes the shape to match the individual regardless of the specific pose. PIE was originally developed for manta rays~\citep{moskvyak2019robust}, but in Wildbook it is also used for humpback whale flukes, orcas, and right whales. HotSpotter~\citep{hotspotter} is a SIFT-based~\citep{lowe1999object} species agnostic algorithm that uses viewpoint invariant descriptors and a scoring mechanism which emphasizes the most distinctive key points, called “hot spots,” on an animal pattern. The HotSpotter algorithm has been successfully used for re-identification of zebras (\emph{Equus quagga})~\citep{hotspotter} and giraffes (\emph{Giraffa tippelskirchi})~\citep{parham2017animal}, jaguars (\emph{Panthera onca})~\citep{hotspotter} and ocelots (\emph{Leopardus pardalis})~\citep{nipko2020identifying}.

Most recent methods for animal re-identification utilize deep learning, particularly convolutional neural networks (CNNs)~\citep{schneider2019past, schneider2020similarity}. CNNs have been successfully applied for re-identification of primate faces~\citep{deb2018face, brust2017towards} and for pattern-based re-identification and recognition of Amur tigers (Panthera tigris)\citep{li2019amur, Liu_2019_ICCV_Workshops, Liu_2019_ICCV}, cattle muzzle \citep{kumar2018deep}, zebras (\emph{Equus quagga}) and giraffes (\emph{Giraffa tippelskirchi}) \citep{badreldeen2021metric}. In order to improve re-identification accuracy, pose estimation and key point alignment have been proposed~\citep{yeleshetty20203d,yu2021ap, moskvyak2021keypoint}.

\subsection{Ringed seal re-identification}

A number of methods for the re-identification of Saimaa ringed seals have been developed~\citep{zhelezniakov2015segmentation,chehrsimin2017automatic,nepovinnykh2018identification,chelak2021eden}. Generally, the methods start with preprocessing steps, including seal segmentation, and then proceed to analyzing the unique pelage pattern to generate a descriptor for each individual seal. A seal segmentation method utilizing superpixel classification was proposed in~\citep{zhelezniakov2015segmentation}. The re-identification method employs common texture features extracted from the segmented seal and a Bayesian classifier. Additional color normalization and contrast enhancement steps were applied in~\citep{chehrsimin2017automatic} to make the pattern more visible. The actual re-identification was performed using the Hotspotter algorithm~\citep{hotspotter}.

The first attempt to utilize CNNs for Saimaa ringed seal identification was done in~\citep{nepovinnykh2018identification}. The individual re-identification was reformulated as a classification problem where each class corresponds to a unique individual, and transfer learning was utilized to train an individual classifier. While the performance is good on a small dataset, the method is only able to reliably perform the re-identification if there is a large set of example images for each individual. Furthermore, the whole system needs to be retrained if a new seal individual is introduced. Finally, it is unclear if the high accuracy is due to the method’s ability to learn the necessary features from the pelage pattern, or if it also learns features such as pose, size, or illumination, which separate individuals in the used dataset but do not provide the means to generalize the method to other datasets.

In order to address these issues, a one-shot approach was proposed in~\citep{nepovinnykh2020siamese}. The method starts with CNN-based segmentation of the seal. The pelage pattern is extracted utilizing a Sato tubeness filter-based method. For the re-identification, the whole pattern image is divided into patches, which are then fed into an embedding CNN. The CNN is trained using a triplet loss and essentially provides a metric that measures the visual similarity between the patches. Re-identification is then performed based on this similarity by using topology-preserving projections. The main advantage of using a triplet CNN is the ability to easily add new individuals into the database since no retraining is necessary.

The pattern embedding step is crucial for any re-identification method as distinctive but compact embedding that captures the characteristics of the pattern forms the basis for successful re-identification. The pattern embedding step was considered in more detail in~\citep{chelak2021eden}, where EDEN, a new pooling layer, was proposed to account for the spatial distribution of pattern features. It was shown that the proposed pooling layer increases the matching accuracy of the pattern patches.

Another version of the re-idenfitication algorithm was proposed and applied to the sister species of Saimaa ringed seals, Ladoga ringed seals (\emph{Pusa hispida ladogensis}) in~\citep{ladoga}. Ladoga ringed seals have a similar pattern to Saimaa ringed seals, which means that the same re-identification algorithm is applicable to both species. Two new steps were introduced into the pipeline: individual grouping and Fisher Vector computation. The individual grouping step focuses on finding multiple instances of the same individual from an image sequence. This rather simple image retrieval-based method was shown to attain high accuracy in matching individuals within an image sequence producing sets of images of each seal to be re-identified. This was shown to be beneficial for the re-identification as it helps to compensate for poor image quality, which often results in an inability to extract patterns from some images, and it allows a larger portion of the pattern to be captured as the seal changes its pose between images. The Fisher Vector~\citep{perronnin2007fisher, perronnin2010large, perronnin2010improving} is used to aggregate patch descriptors from an individual seal into a single image descriptor. The vector further allows the patch descriptors from multiple images to be aggregated, providing a straightforward tool for utilize utilization of the image sets produced by the grouping step. Aggregated image descriptors are used to find a match from the database of known individuals by calculating distances. Promising results were obtained on Ladoga ringed seal re-identification.

\subsection{Content based image retrieval}
The task of visual animal re-identification can be formulated as a task of finding the most similar image from the database to the given query image. This formulation matches the definition of content-based image retrieval (CBIR)~\citep{smeulders2000content} and motivates study of the suitability of CBIR methods for animal re-identification.

CBIR methods usually consist of two main steps: feature extraction and feature aggregation. The feature extraction problem can be solved using standard hand-crafted features, such as Scale Invariant Feature Transform (SIFT)~\citep{lowe2004distinctive, arandjelovic2012three}, or extraction by convolutional neural networks (see e.g.~\citep{HardNet2017}). Then, feature aggregation creates a descriptor for each image that can be used to find the most similar image from the database. Traditional methods such as Bag of Words (BOW)~\citep{sivic2003video}, Vector of Locally Aggregated Descriptors (VLAD)~\citep{jegou2010aggregating} and the Fisher Vector~\citep{perronnin2007fisher, perronnin2010large, perronnin2010improving} do the aggregation using a specially constructed vocabulary. The vocabulary is usually created by an unsupervised clustering algorithm. For example, k-means~\citep{macqueen1967some} is used for VLAD and a Gaussian Mixture Model (GMM)~\citep{mclachlan1988mixture} is used for the Fisher Vector. Finally, fixed-size descriptors are created for each image based on the vocabulary and extracted features. The distance between these descriptors is inversely proportional to the visual similarity.

Due to the availability of data and the convenience of end-to-end approaches, deep learning-based methods for CBIR are becoming increasingly more popular. The advantage of CNN-based methods is that the two main steps, feature extraction and feature aggregation, are naturally implemented as a part of the network architecture, with the first part of the network being a feature extractor and a final specialized layer doing the feature aggregation. For example, there have been several attempts to create deep analogues of traditional methods such as NetVLAD~\citep{arandjelovic2016netvlad} where a generalized VLAD layer is used to aggregate CNN-extracted features. 

In~\citep{gong2014multi, babenko2014neural}, fully-connected layers are used to generate the final descriptor, which is a standard approach for CNNs. In~\citep{razavian2016visual}, a global max pooling approach is introduced which produces the final descriptor from the activation maps by taking a maximum value from each filter activation, resulting in a descriptor of the same size as the number of filters. Different variants of global pooling operations have also been studied. These include integral max-pooling~\citep{tolias2015particular}, sum pooling~\citep{babenko2015aggregating} and generalized mean pooling~\citep{radenovic2018fine}. Integral max-pooling~\citep{tolias2015particular} is particularly interesting since it creates the final descriptor by applying max-pooling to the overlapping image regions, which also allows spatial information to be encoded.

\section{Method}

\begin{figure*}[ht!]
	\centering
	\includegraphics[width=\linewidth]{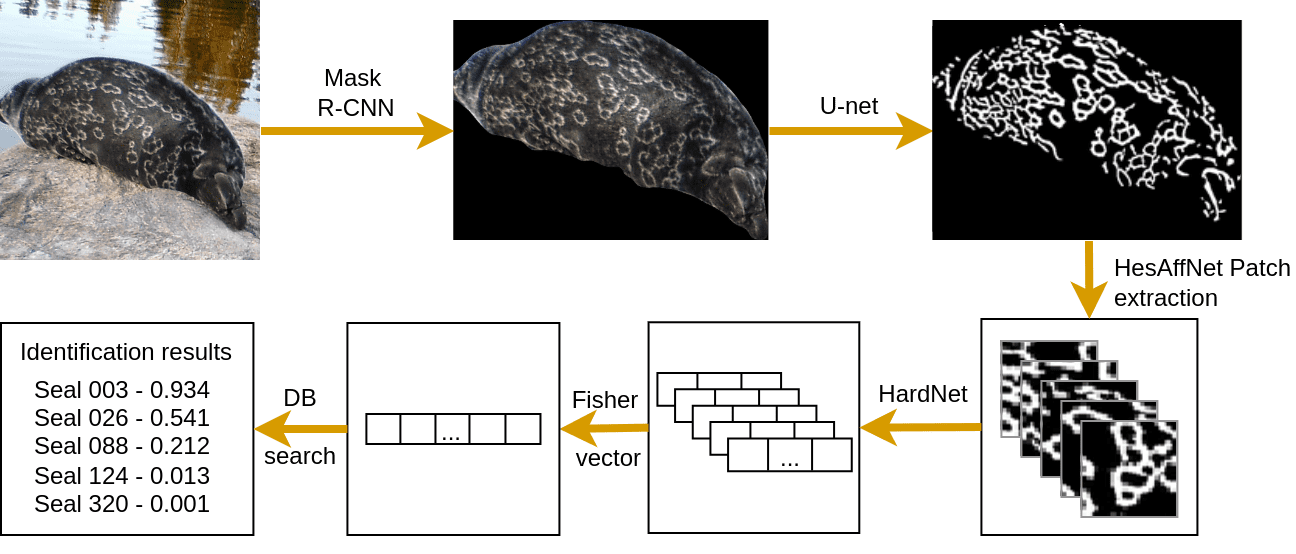}
	\caption{NORPPA re-identification pipeline.}
	\label{fig:pipeline}
\end{figure*}

The proposed NORPPA method consists of six steps: 1) image prepossessing, 2) seal instance segmentation, 3) pelage pattern extraction, 4) feature extraction, 5) feature aggregation and 6) individual re-identification (see Fig.~\ref{fig:pipeline}).

\subsection{Image preprocessing}
\begin{figure}[ht!]
		\centering
		\minipage{0.499\linewidth}
		\includegraphics[width=\linewidth]{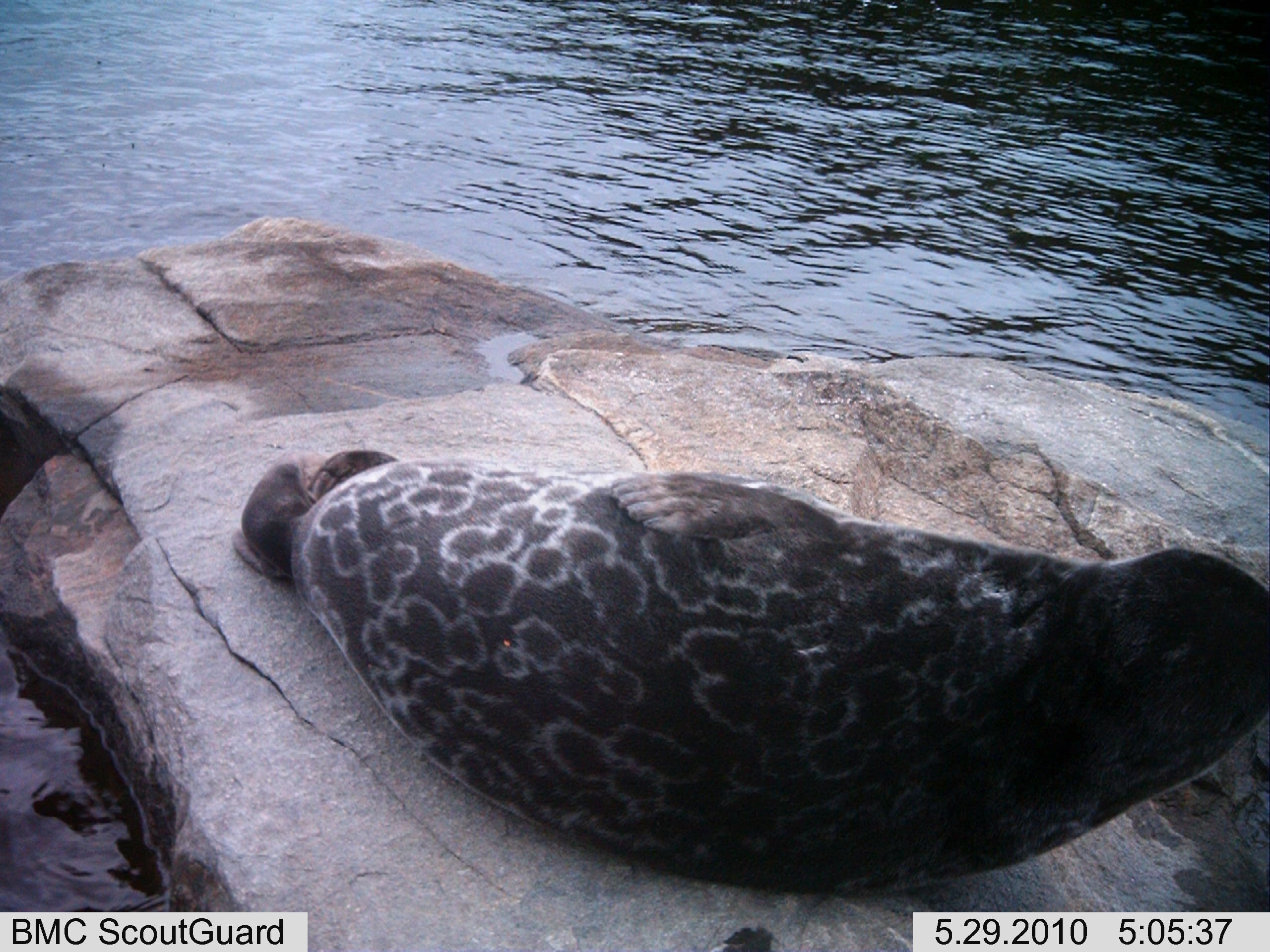}
		\endminipage\hfill
		\minipage{0.499\linewidth}
		\includegraphics[width=\linewidth]{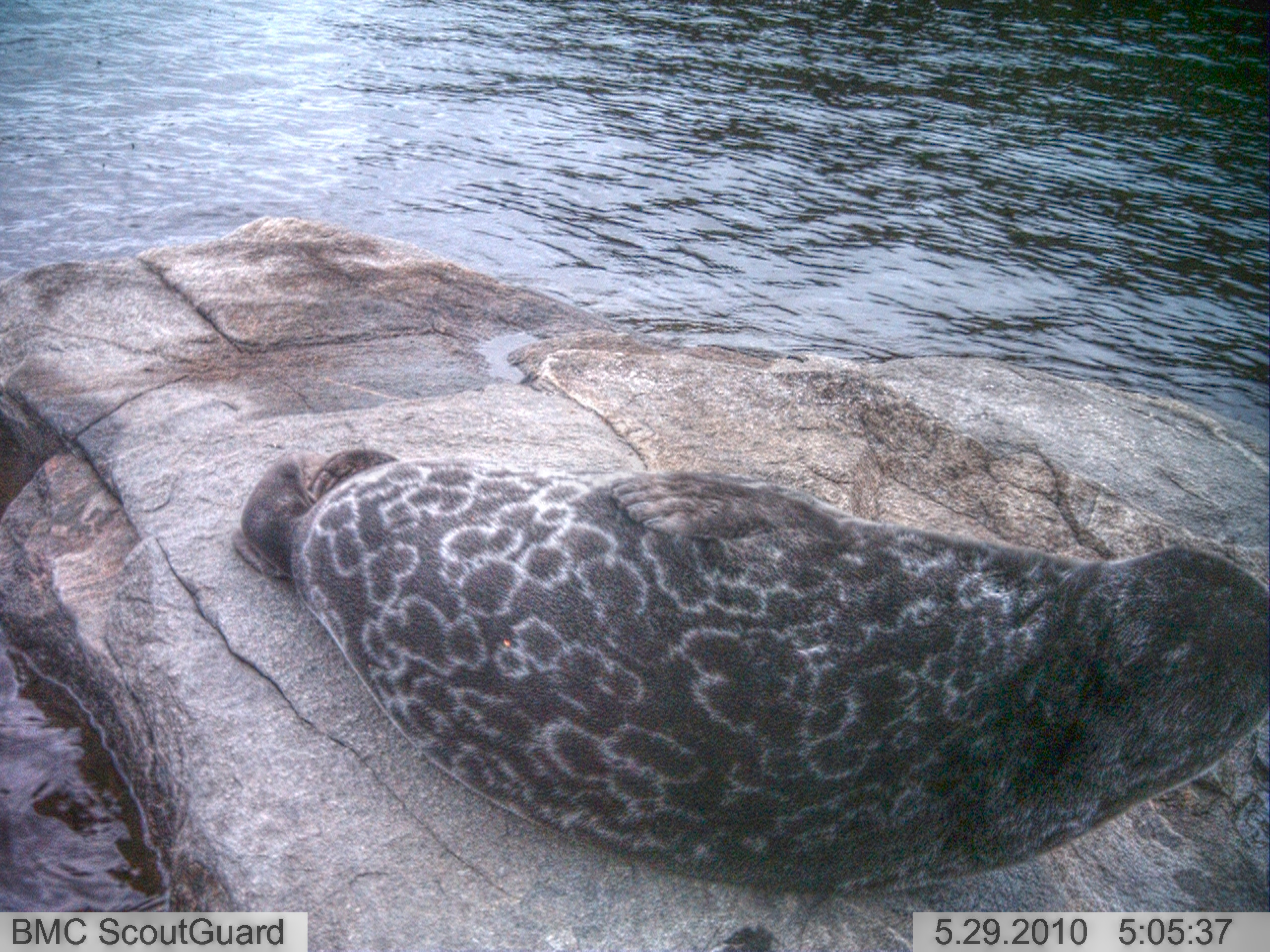}
		\endminipage\hfill
		\minipage{0.499\linewidth}
		\includegraphics[width=\linewidth]{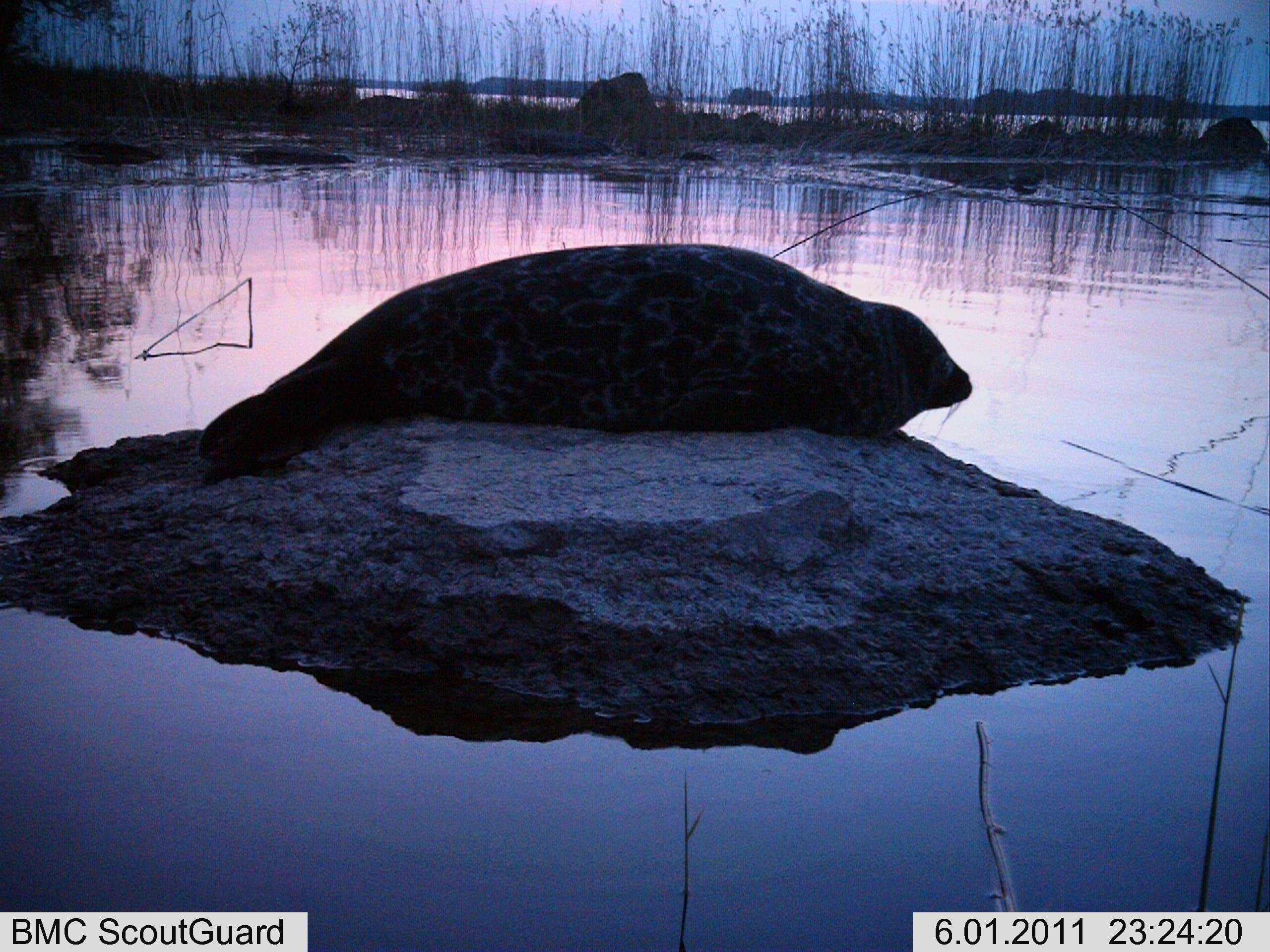}
		\endminipage\hfill
		\minipage{0.499\linewidth}
		\includegraphics[width=\linewidth]{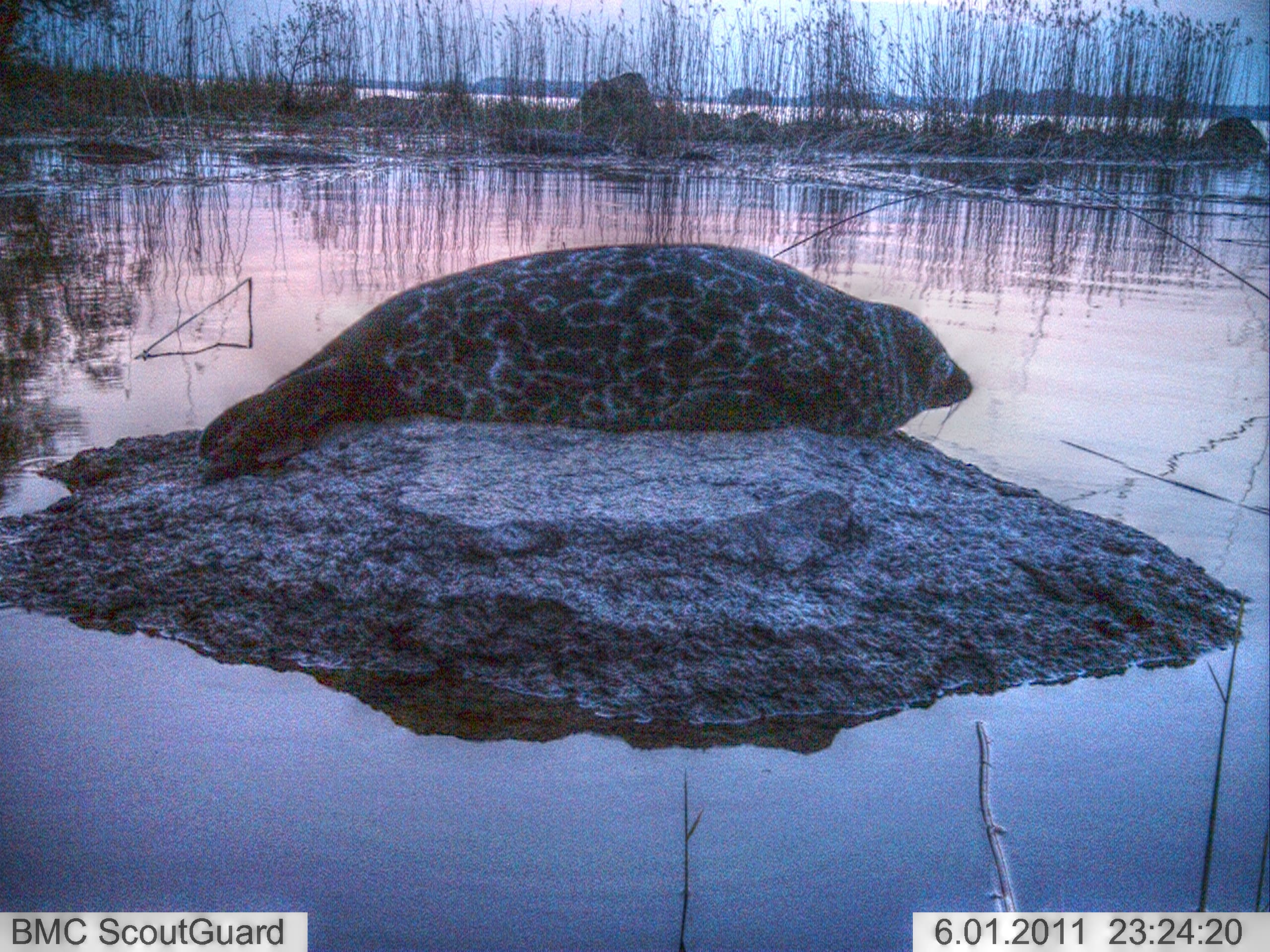}
		\endminipage
		
		\caption{Examples of the image processing of camera trap images. Images on the left are the originals. The right column demonstrates the result of the tone-mapping.}
		\label{fig:tonemapping_examples}
	\end{figure}

Depending on illumination conditions variation in the contrast of the images can be rather high. This could lead to a loss of detail in the region of interest, i.e. the seal and its pelage pattern. In order to rectify this issue, we employ the tone-mapping approach to equalize the contrast in dark and bright image regions. The algorithm proposed by~\citep{mantiuk_perceptual_2006} is used due to its ability to produce realistic tone-mapped images without introducing visual artifacts. This method considers contrast on multiple spatial frequencies while using gradient methods with some additional extensions to ensure that the global brightness levels are not reversed and low-frequency details are properly reconstructed. 
Examples of images before and after prepossessing are presented in Fig.~\ref{fig:tonemapping_examples}.

\subsection{Seal instance segmentation}

Seal instance segmentation step is important since most of the images are obtained using static camera traps. This together with the fact that seal individuals tend to use same sites or areas inter-annually cause one seal individual to be very often captured with the same camera (same background). This increases the risk that the supervised identification algorithm learns to identify the background instead of the actual seal if the full image or the bounding box around the seal is used. Consequently, algorithm behavior may result in a system that is unable to identify the seal in a new environment.

Instance segmentation is performed using Mask R-CNN~\citep{he2017mask}. A segmentation model trained for Ladoga ringed seals from~\citep{ladoga} is utilised. This is possible due to the two species being visually almost indistinguishable. Ladoga ringed seals are more numerous than Saimaa ringed seals and they are often captures in large groups which makes it easier to collect and annotate large training data for the segmentation. For more details about the instance segmentation model and training procedure see~\citep{ladoga}.

\begin{figure}[ht!]
		\centering
		\minipage{0.33\linewidth}
		\includegraphics[width=\linewidth]{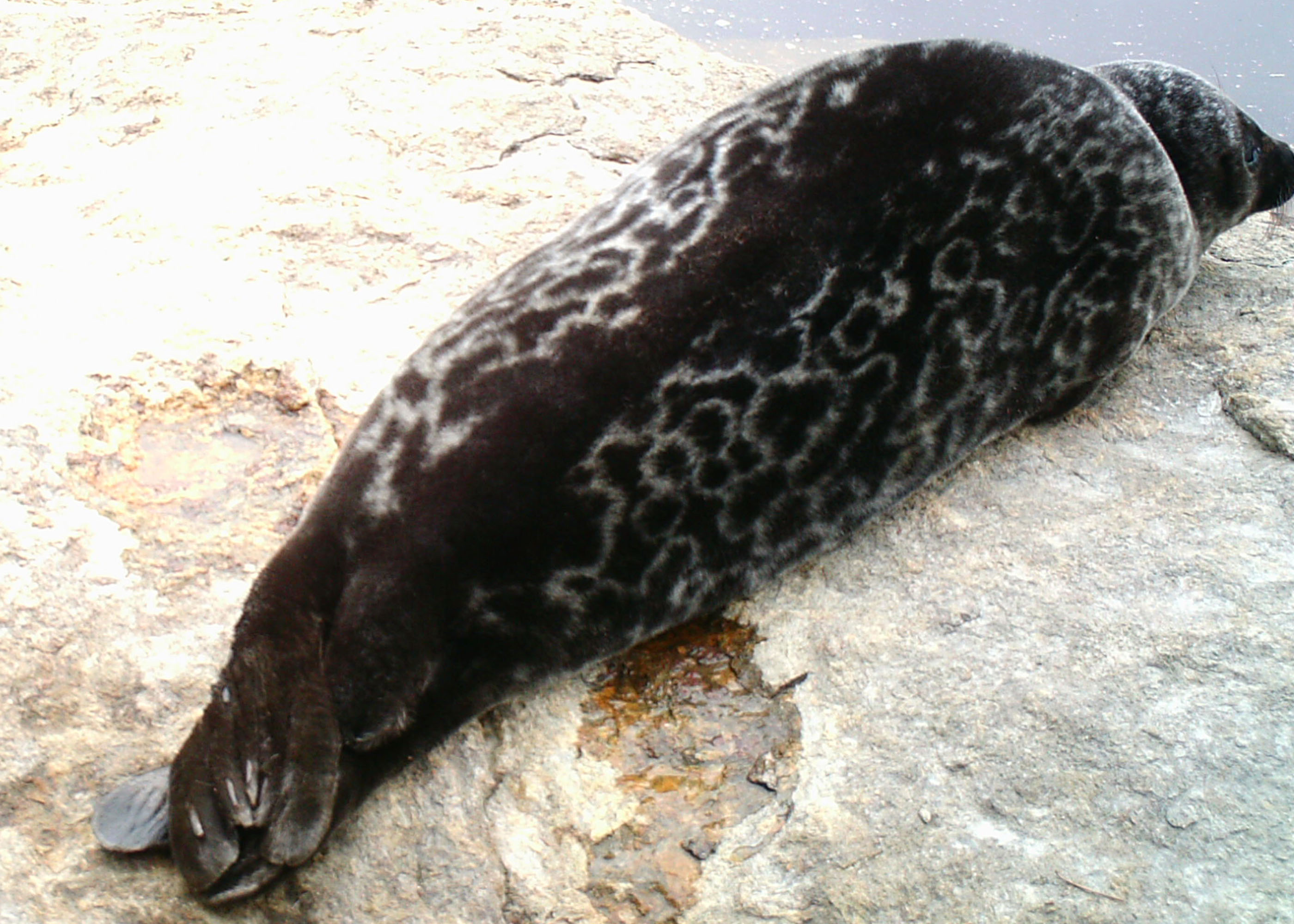}
		\endminipage\hfill
		\minipage{0.33\linewidth}
		\includegraphics[width=\linewidth]{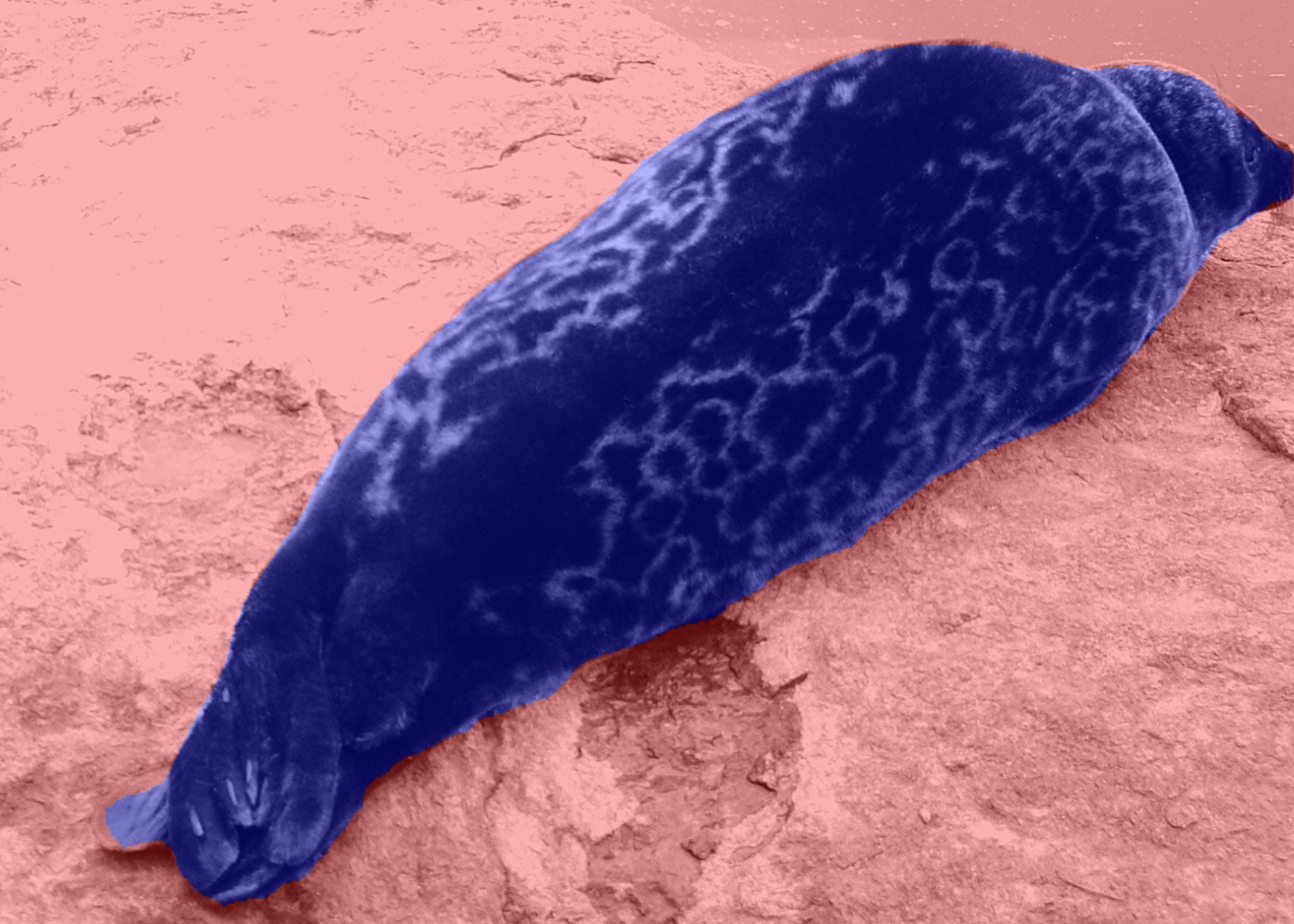}
		\endminipage\hfill
		\minipage{0.33\linewidth}
		\includegraphics[width=\linewidth]{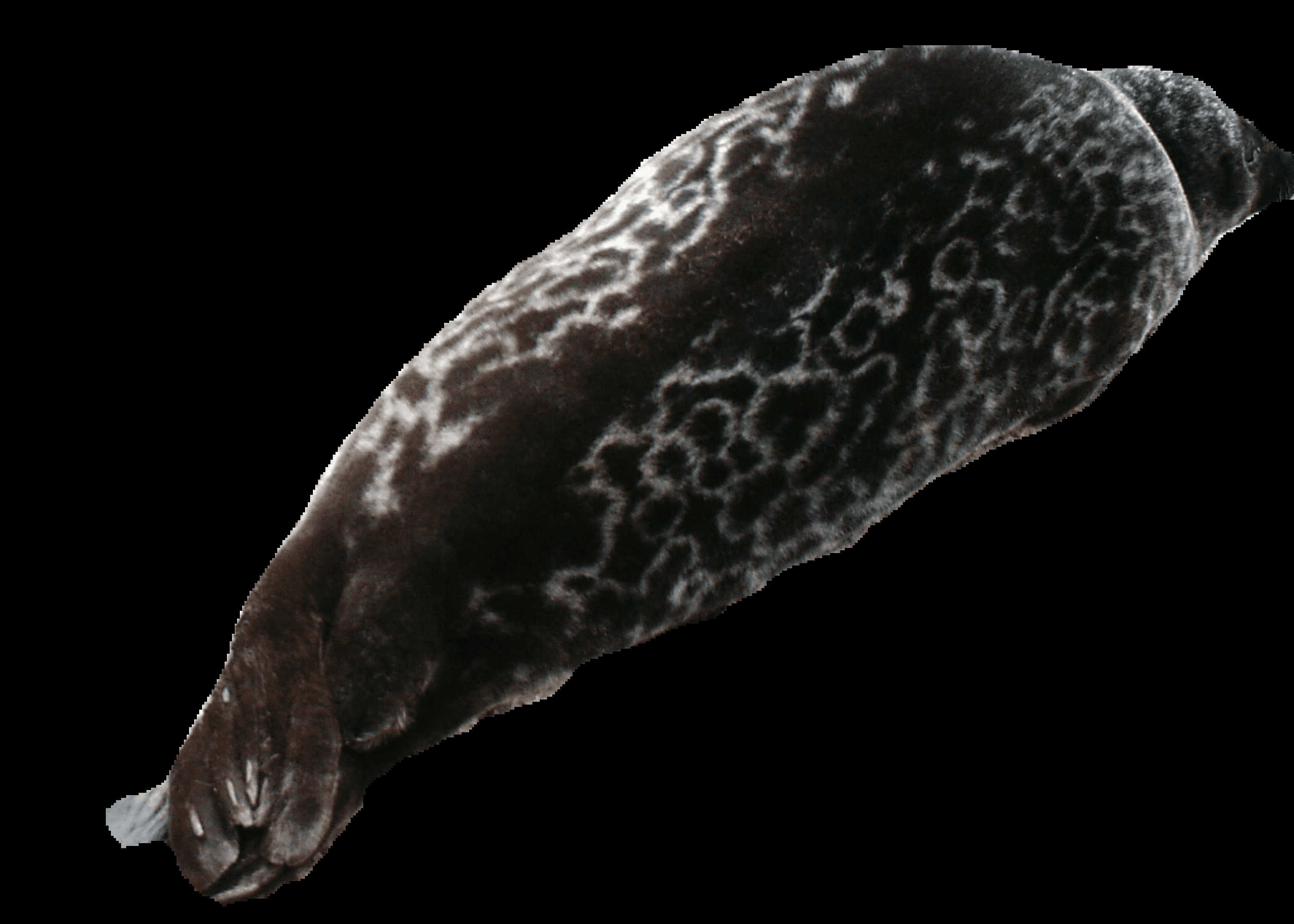}
		\endminipage\hfill
		\minipage{0.33\linewidth}
		\includegraphics[width=\linewidth]{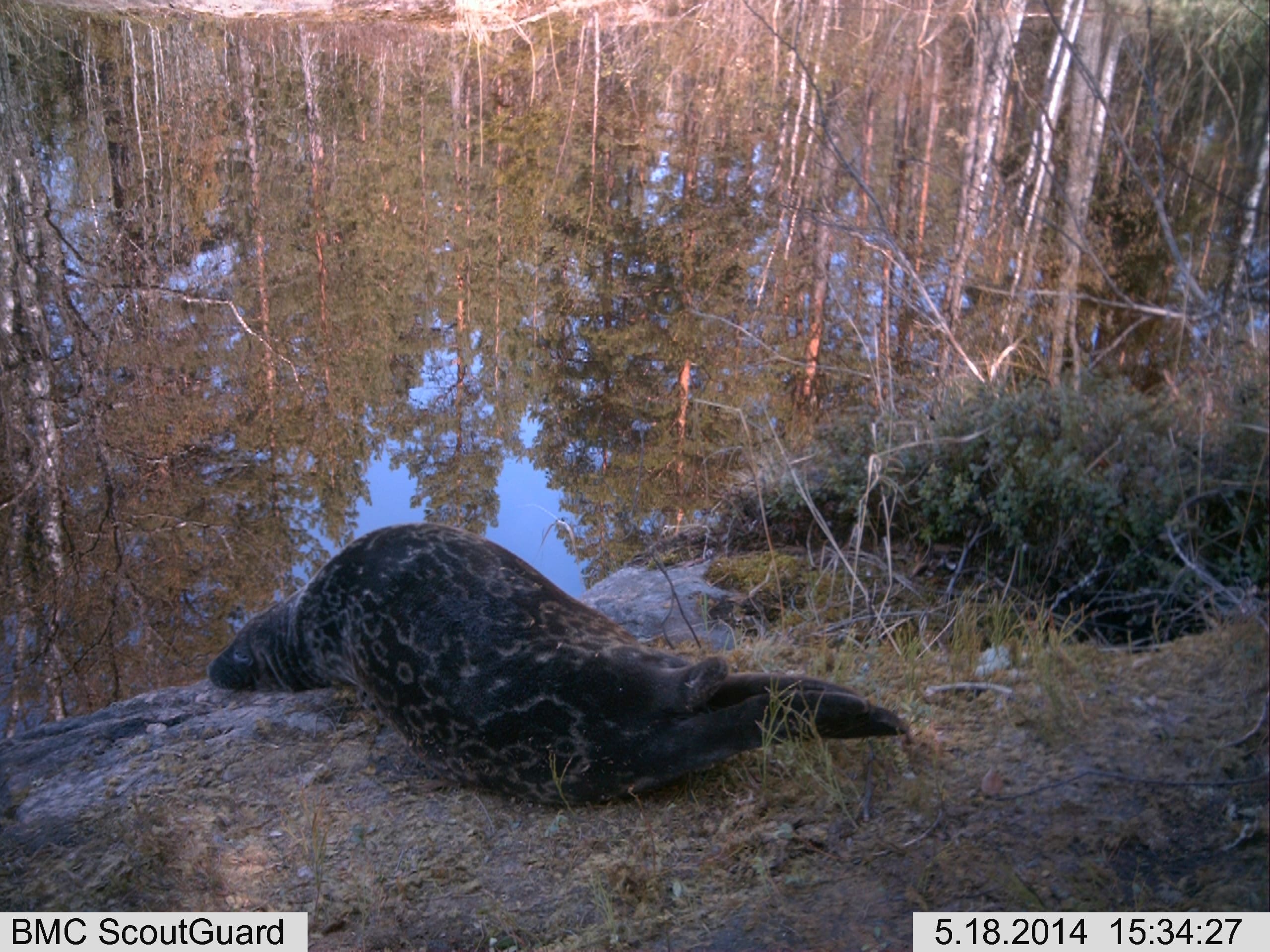}
		\endminipage\hfill
		\minipage{0.33\linewidth}
		\includegraphics[width=\linewidth]{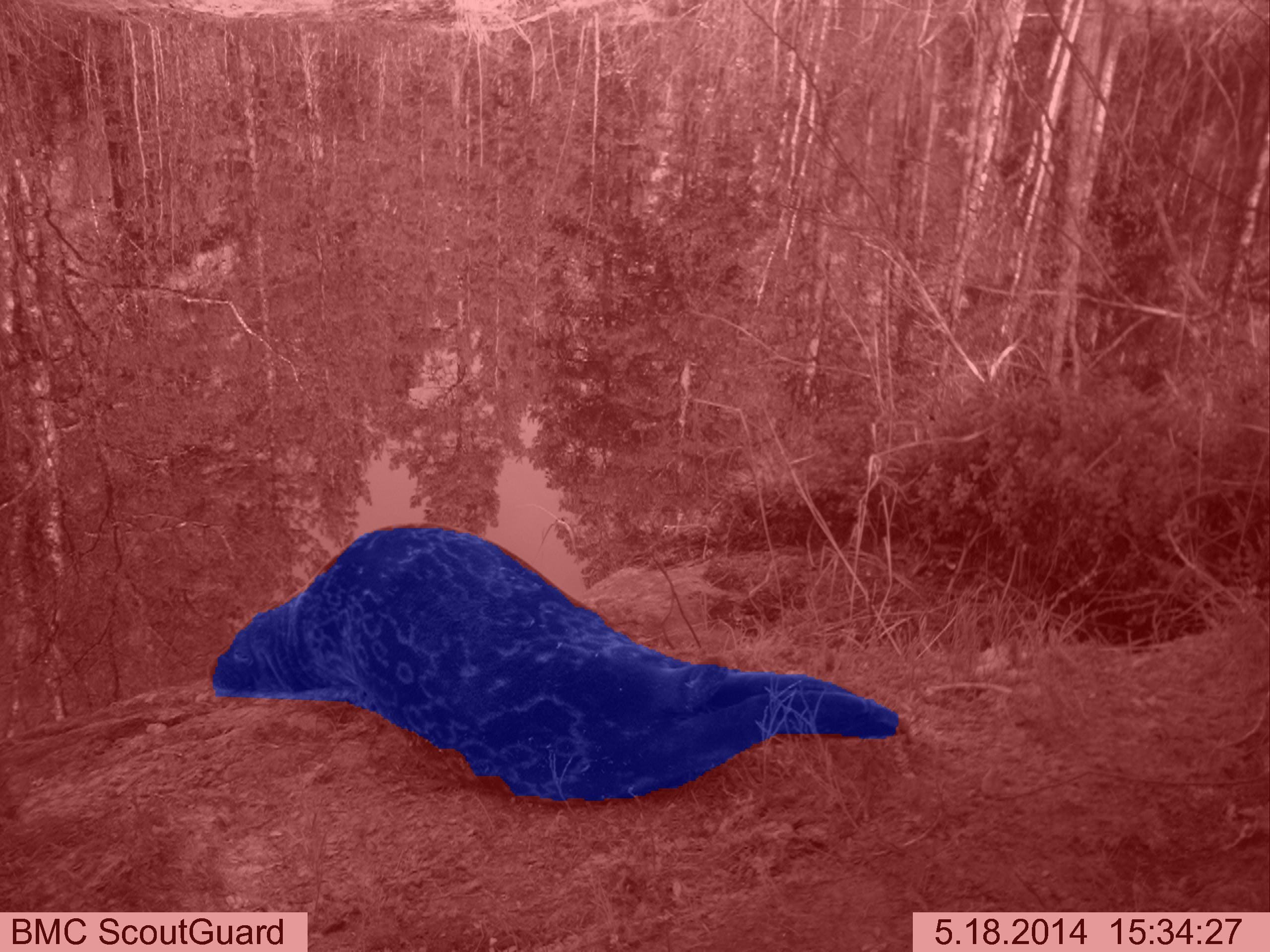}
		\endminipage\hfill
		\minipage{0.33\linewidth}
		\includegraphics[width=\linewidth]{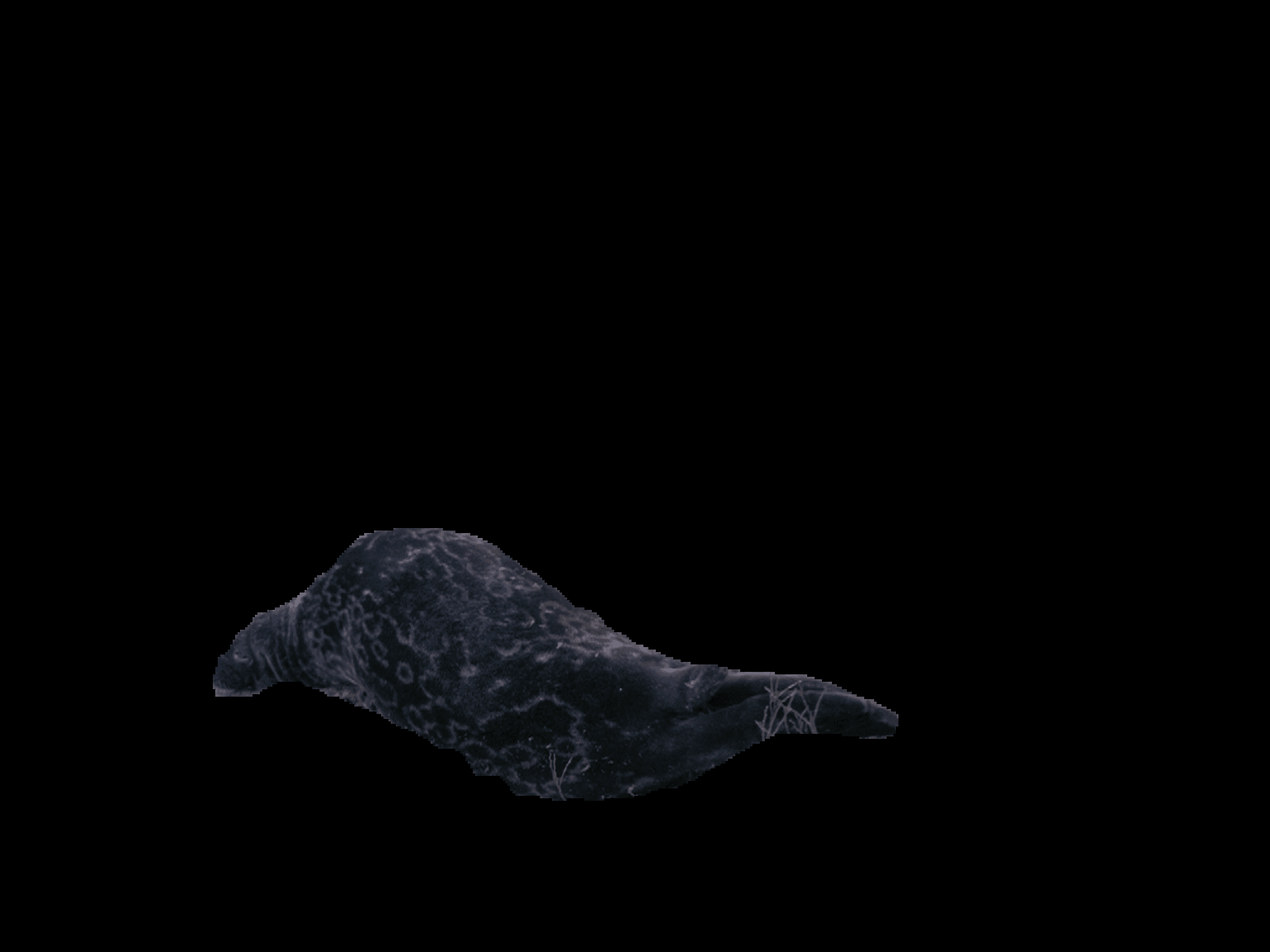}
		\endminipage
		\caption{Examples of the segmentation masks. Images on the left are the originals. The mask is highlighted in blue and the background is highlighted in red on the middle images. The last column shows the result of the segmentation.}
		\label{fig:segmentation_examples}
	\end{figure}

After the segmentation masks are obtained, additional morphological operations are applied to close the holes and smooth the borders by using morphological closing and opening. The examples of segmentation results are presented in Fig.~\ref{fig:segmentation_examples}.
	
\subsection{Pelage pattern extraction}
\begin{figure}[ht!]
	\centering
	\includegraphics[width=1.0\linewidth]{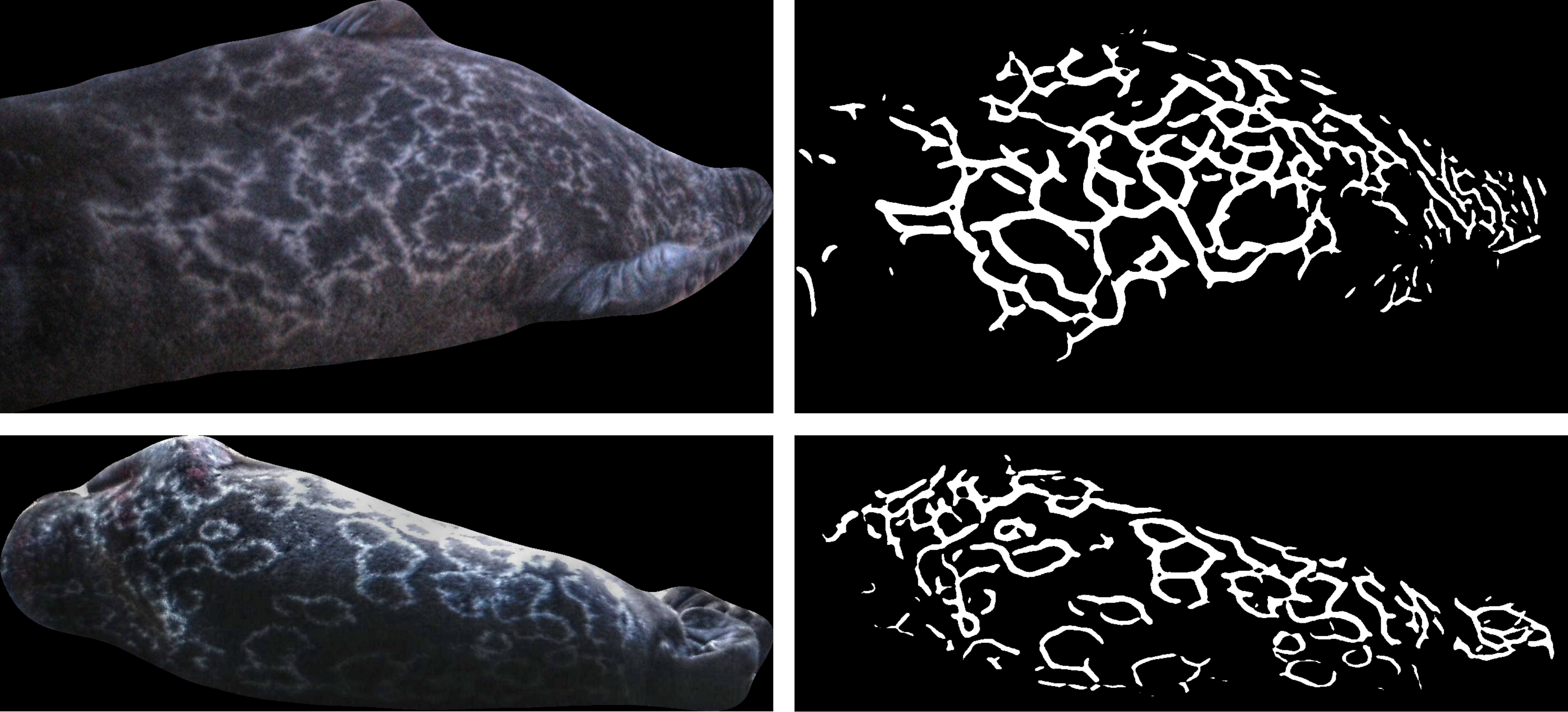}
	\caption{Example of pattern extraction output.}
	\label{fig:pattern_extraction}
\end{figure}

The main distinguishing feature of a seal is its pelage pattern, which is both permanent and unique to each seal allowing the identification of individuals over their whole lifetime. The pelage pattern forms the basis for the proposed re-identification method. In order to focus the attention on the pattern and discard irrelevant information causing database bias such as illumination and other visual factors (e.g., wet fur looks different from the dry fur), the pattern is segmented. This is done using CNN based method utilizing U-net encoder-decoder architecture~\citep{ronneberger2015u}. The output of the method is a binarized image of the pelage pattern (see Fig.~\ref{fig:pattern_extraction}). The pattern image is further post-processed to remove small noise by using unsharp masking and morphological opening. All images are then resized in such way that the mean width of the pattern lines is the same for all images, bringing them into the same scale. This is necessary because the images are obtained from various sources and the image resolution has a large variation. For more detailed explanation of the pattern extraction step, as well as the comparison to other methods, see~\citep{zavialkin2020cnn}. 

\subsection{Feature extraction}
Seals can be found in a variety of poses. The deformable nature of seals body results in distorted and warped patterns on images. While the pattern as a whole is transformed in a non-linear way, it can be argued that small local regions experience close to affine transformations, making an affine invariant feature extractor suitable for the task. For this purpose a combintaion HesAffNet~\citep{AffNet2018} detector and HardNet~\citep{HardNet2017} descriptor is used.

The combination of a Hessian-Affine detector~\citep{mikolajczyk2004scale} with RootSIFT~\citep{arandjelovic2012three} used to be considered a gold standard for local feature extraction and description. However, with the increasing size of available datasets and rapidly developing field of deep learning, CNN-based methods are now able to outperform previous handcrafted features. The combination of HesAffNet~\citep{AffNet2018} and HardNet~\citep{HardNet2017} is able to provide state-of-the-art results in image retrieval tasks, which makes those methods particularly useful for animal re-identification as well.

HesAffNet is a modification of the classical Hessian Affine Region detector~\citep{mikolajczyk2002affine, mikolajczyk2004scale}, where the shape estimation step is done by the AffNet CNN. The detector is based on  the Harris cornerness measure~\citep{Harris1988ACC}, which uses a second moments matrix to find regions of interest by estimating the most prominent gradient directions. This method is combined with the multiscale approach from \citep{lindeberg1998feature} which uses Laplacian of Gaussian to find extrema in the scale space. The same concept can be further extended to all affine transformations, not just the scale. However, the degree of freedom is much higher for affine transformations, which complicates the process and requires a special shape adaptation algorithm. The original Hessian Affine detector used Baumberg iteration~\citep{Baumberg2000}, which is replaced by an AffNet CNN in HesAffNet. 

AffNet and HardNet are closely related, sharing the architecture and similar training procedure. During the training of HardNet, batches of matching patch pairs are chosen, each containing an anchor $a_i$ and positive match $p_i$. Each pair correspond to a different location, i.e. there are no other matches except for the ones in each pair. Each patch is encoded by the network, and a matrix of pair-wise distances between all anchors and positive matches are computed. For each pair, a closest non-matching descriptor from the batch is chosen, and a final hard negative margin loss is computed as
\begin{equation}
    \begin{split}
        L = \frac{1}{n} \sum_{i=1}^n & \max(0, 1+d(a_i, p_i)) \\ &- \min(d(a_i, p_{j\min}), d(a_{j\min}, p_i)),
    \end{split}
\end{equation}
where $p_{j\min}$ is the closest non-matching positive to $a_i$, and $a_{j\min}$ is the closest non-matching anchor to $p_i$.  

\begin{figure}[htp]
	\centering
	\subfloat[][]
	{
		\includegraphics[width=0.45\linewidth]{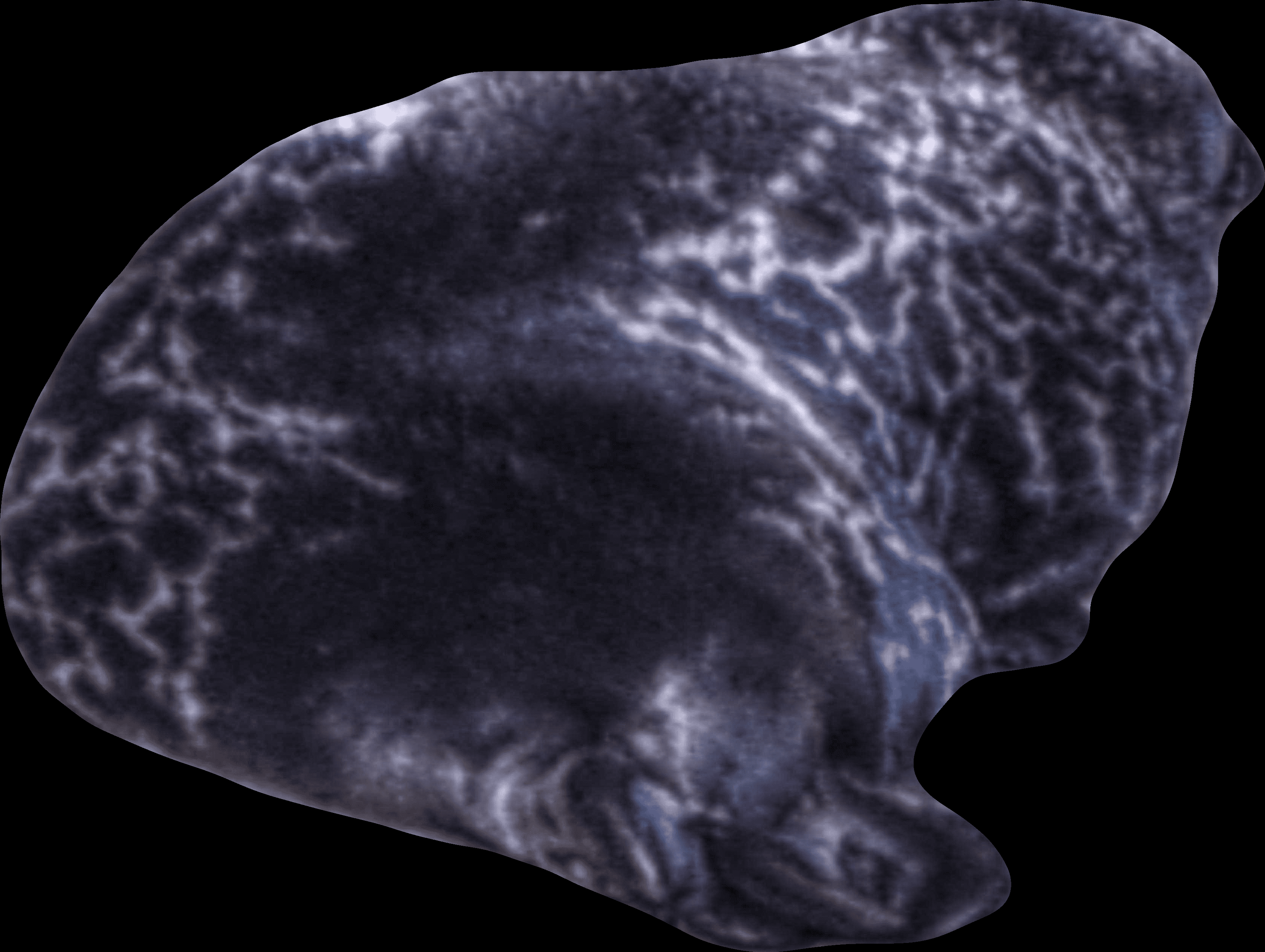}
		\label{subfig:patch_original}
	}
	\subfloat[][]
	{
		\includegraphics[width=0.45\linewidth]{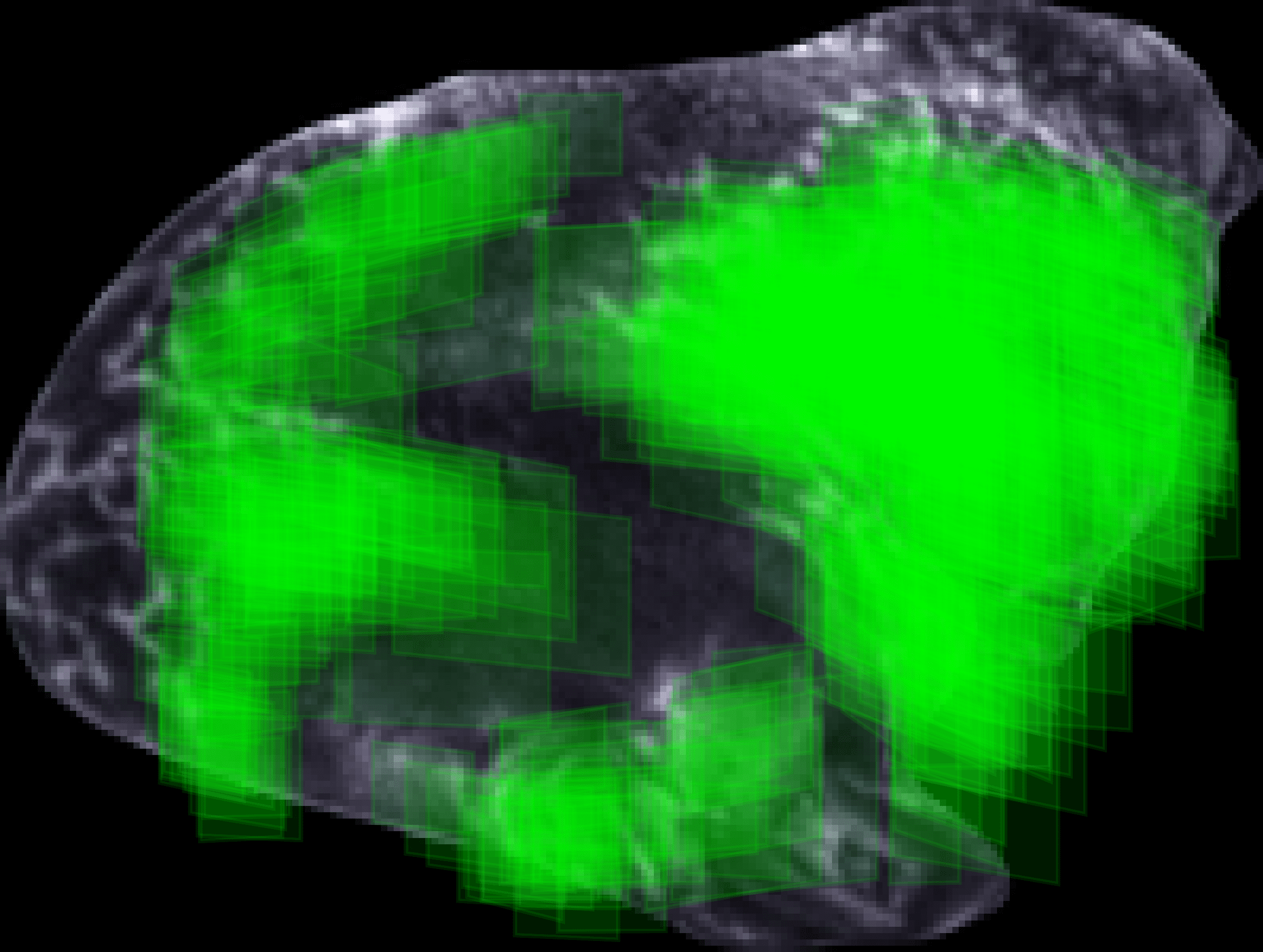}
		\label{subfig:patch_hesaff}
	}
	\caption[]{Visualisation of Hessian Affine patch extraction: (\subref{subfig:patch_original}) segmented image;  (\subref{subfig:patch_hesaff}) HesAffNet-based patch extraction.
	Extracted regions are highlighted in green.}
	\label{fig:patch_vis}
\end{figure}

AffNet utilizes a slightly different training procedure, the main difference being that the derivative for the negative term in the loss is set to 0. This loss is called hard negative-constant and helps avoid the situations where positive samples cannot be moved closer together because of a negative sample lying between them in the metric space.
The training procedure for AffNet is also more complicated, since it is learning affine shapes and not just a distance metric. Therefore, spatial transformers are used to transform input patches according to the predicted shape, which are then fed into a descriptor network, e.g. HardNet, and only then is the loss calculated and backpropagated through both networks. The example of HesaffNet application to a preprocessed image is visualised in Figure~\ref{fig:patch_vis}.

\subsection{Feature aggregation}

Features are aggregated using Fisher Vector~\citep{perronnin2007fisher, perronnin2010large, perronnin2010improving}. 
First, Principal Component Analysis (PCA) is applied to the resulting the feature embeddings to decorrelate the features and reduce the dimensionality. This is an important for Fisher Vectors, which are known to produce large descriptors. The images in the database of known individuals are used to learn principal components. Next, a visual vocabulary is constructed by applying Gaussian Mixture Model (GMM) to the features from the database. Then, Fisher Vectors are created for each image by computing the partial derivatives of the log-likelihood function with respect to the GMM parameters and concatenating them. Kernel PCA~\citep{scholkopf_nonlinear_1998} is applied to further reduce the dimensionality of the resulting image descriptors which helps to reduce the storage requirements for the database, as well as speed up the database search for the re-identification.

\subsubsection{Fisher Vector}

Let $X = \{x_t, t=1, \ldots, T\}$ be a sample of $T$ observations and $u_\lambda$ be a probability density function modelling the distribution of the data, where $\lambda$ is a vector of its parameters. The score is defined as the gradient of the log-likelihood of the data on the model:
\begin{equation}
G^X_{\lambda} = \nabla_{\lambda}\log u_{\lambda}(X).
\end{equation}
This score function can be used to define the Fisher Information Matrix (FIM)~\citep{amari2000methods}:
\begin{equation} 
F_{\lambda} = E_{x\sim u_{\lambda}}[G^X_{\lambda}{G^X_{\lambda}}'], 
\end{equation}
which acts as a local metric for a parametric family of distributions. This metric can also be used to measure the similarity between 2 samples using the Fisher Kernel (FK)~\citep{jaakkola1998exploiting}:
\begin{equation} 
    \begin{split}
        K_{FK}(X,Y) &= {G^X_{\lambda}}' F^{-1}_{\lambda} G^Y_{\lambda} \\ &= {G^X_{\lambda}}' {L_{\lambda}}' L_{\lambda} G^Y_{\lambda} \\ &= {\mathscr{G}^X_{\lambda}}' \mathscr{G}^Y_{\lambda},
    \end{split}
\end{equation}
where $L_\lambda'L_\lambda$ is the Cholesky decomposition of $F^{-1}_{\lambda}$, $G^X_{\lambda}$ and $G^Y_{\lambda}$ are the Fisher Vectors of samples $X$ and $Y$ respectively. By using Fisher Vectors, it is possible to calculate the kernel as simple dot product, which can efficiently be utilized by linear classifiers.
When constructing a Fisher Vector for an image, a set of local features is assumed to be independent, meaning that the final descriptor can be constructed as a sum of Fisher Vectors for each local feature, i.e.
\begin{equation}
G^X_{\lambda} = \sum_{t=1}^T L_\lambda \nabla_{\lambda}\log u_{\lambda}(X).
\end{equation}
Usually, Gaussian Mixture Model (GMM) is used as $u_\lambda$, since it can be used to approximate any continuous distribution with arbitrary precision~\citep{titterington1985statistical}. Then, the vector of parameters $\lambda$ contains mixture weights $w_k$, mean vectors $\mu_k$ and covariance matrices $\Sigma_k$ for each Gaussian $u_k, k=1,\ldots,K$.
Using the assumption that the assignment of each feature to mixture components is almost hard, i.e. each feature is assigned to only one cluster, it could be inferred~\citep{sanchez2013image} that the FIM is diagonal, which means that $L_\lambda$ is just a coordinate-wise normalization of the gradient vectors. The final normalized gradients are then defined as follows
\begin{equation}
    \mathscr{G}^X_{\mu_k} = \frac{1}{\sqrt{w_k}}\sum_{t=1}^{T}\gamma_t(k)(\frac{x_t-\mu _k}{\sigma_k}),
\end{equation}
\begin{equation}
     \mathscr{G}^X_{\sigma_k} = \frac{1}{\sqrt{w_k}}\sum_{t=1}^{T}\gamma_t(k)\frac{1}{\sqrt{2}}\left[ \frac{(x_t-\mu_k)^2}{\sigma^2_k }-1\right],
\end{equation}
where $\gamma_t$ is the soft assignment function
\begin{equation} 
    \gamma_t(k)=\frac{w_k u_k(x_t)}{\sum_{j=1}^{K} w_j u_j(x_t)}.
\end{equation}
It should be noted that the gradients for the weight parameters $w_k$ are usually omitted, since they do not provide much additional information~\citep{perronnin2010improving}. Those gradients are concatenated into a vector of size $2DK$, where  $D$ is the dimensionality of samples and $K$ is the number of components in GMM. It has been shown~\citep{perronnin2010improving} that $L2$ and Power normalization generally improve the performance of the method. Therefore, it is common to apply Power and $L2$ normalization to the Fisher Vector to get the final descriptor.

\subsection{Individual re-identification}
Re-identification is done by calculating the cosine distance from the query image descriptor to each image descriptor in the database of known individuals and selecting the individual ID with the lowest distance. 
To visualize the re-identification and to provide semi-automatic tool for experts, heatmaps highlighting the similar areas in patterns of the query image and database images are computed. This is done using the following method. First, features from a query are paired with the closest database features. Then, pairs with distance larger than 10th percentile of distances are discarded. The remaining pairs are used to find the homography using Direct Linear Transform (DLT)~\citep{DLT} and Random Sample Consensus (RANSAC)~\citep{RANSAC}. The inliers of the final homography are highlighted with ellipses aligned and transformed according to the extracted affine regions. The intensity of each ellipse is inversely proportional to the distance between the local features in the corresponding pair, i.e. directly proportional to their similarity. 

\section{Experiments and results}


\subsection{Data}

\begin{figure}[ht!]
	\centering
	\includegraphics[width=\linewidth]{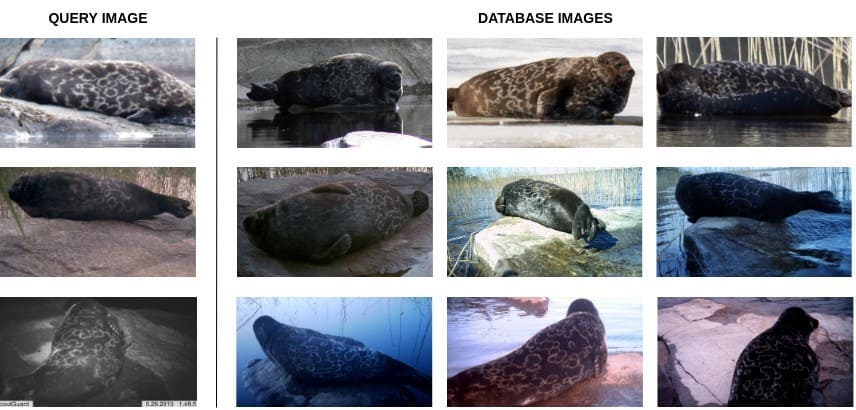}
	\caption{Examples from the database and query datasets. Every row contains images of an individual seal. For every image from the query dataset (left) there is a corresponding subset of images from the database (right). }
	\label{fig:datasets}
\end{figure}

The dataset consists of 57 individual seals with a total of 2080 images. The dataset is divided into two subsets: database and query. The database subset contains a minimal number of high-quality unique images that are enough to cover the full body pattern of each seal. The query subset contains the remaining images and contains the same individuals as in the database. It should be noted that the high-quality images were prioritized when constructing the database and, therefore, images in the query subset often have lower quality. Examples of images from both subsets are presented in Fig. \ref{fig:datasets}. The dataset has been made publicly available. For further description of the dataset, see~\citep{dataset}.

\begin{figure}[t!]
	\begin{center}
		\includegraphics[width=1\linewidth]{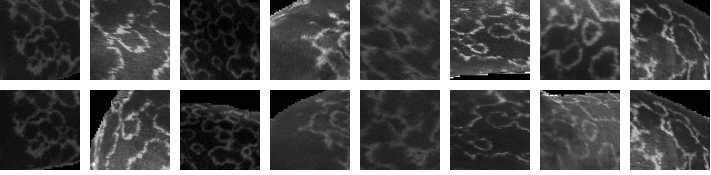}
		\caption{Examples of pattern image patches. The patches in the second row match the patches in the first row.}
		\label{fig:original_patches}
	\end{center}
\end{figure}

To train and evaluate the patch embedding (feature extraction) and matching (finding the corresponding patch in other images) a separate dataset of pattern image patches (see Fig.\ref{fig:original_patches}) was constructed~\citep{chelak2021eden}. The dataset contains, in total, 4599 images (patches of the size $256 \times 256$ pixels). The data is divided into training and testing subsets. The training subset contains 3016 images and 16 classes. The testing subset contains 1583 images and 26 classes that are different from the training classes in the training set. Each class corresponds to one manually selected location in the pelage pattern of one individual seal. Each sample from one class was extracted from different images of the same seal. For estimation of the accuracy of the method, the testing set was divided into the database and query subset with a ratio of 1 to 2. The images that were used to construct the dataset of pattern image patches are not included in the database and query subsets of the re-identification dataset.

\subsection{Feature extraction}
The feature extraction step contains two differences compared to the previous version of the Saimaa ringed seal re-identification algorithm~\citep{nepovinnykh2020siamese}. The first difference is that the region of interest detection approach uses the affine invariant regions (HesAffNet) instead of dense patches. The second difference is a switch to HardNet network to compute patch embedding. To assess the necessity of each of these changes both modifications were evaluated separately. Hyperparameters for all versions of the algorithm were chosen using the Tree Parzen Estimator~\citep{bergstra2011algorithms} algorithm. The results of the experiments are presented in Table~\ref{tab:reid}.

\begin{table*}[ht!]
    \centering
    \caption{Re-identification accuracy for different variants of the algorithm.}
    \begin{tabular}{lcccccc}
        \toprule
        \multicolumn{1}{l}{\begin{tabular}[c]{@{}l@{}}Patch\\extraction\end{tabular} } & \multicolumn{1}{c}{\begin{tabular}[c]{@{}c@{}}Patch\\ embedding\end{tabular}} & \multicolumn{1}{c}{Top-1} & \multicolumn{1}{c}{Top-2} & \multicolumn{1}{c}{Top-3} & \multicolumn{1}{c}{Top-4} & \multicolumn{1}{c}{Top-5} \\ 
        \midrule
        \multirow{3}{*}{Dense}
        & Triplet       & 52.06\%    & 56.91\%    & 60.36\%    & 63.58\%    & 65.70\%    \\ 
        & ArcFace       & 39.94\% & 45.15\% & 50.06\% & 53.58\% & 56.67\%    \\ 
        & HardNet               & 52.18\% & 57.88\% & 61.70\% & 64.61\% & 67.27\%    \\ 
        \midrule
        \multirow{3}{*}{HessAffNet} 
        & Triplet       & 60.42\%    & 65.03\%    & 69.27\%    & 71.64\%    & 73.52\%    \\ 
        & ArcFace  & 47.03\%  & 51.64\%    & 55.58\%    & 58.36\%    & 60.55\%    \\ 
        & HardNet   & \textbf{77.64\%} & \textbf{80.91\%} & \textbf{82.97\%} & \textbf{84.18\%} & \textbf{85.27\%}         
        \\
        \botrule
    \end{tabular}
    \label{tab:reid}
\end{table*}

As can be seen, both HesAffNet for region of interest detection and HardNet for patch embedding computation improve the accuracy noticeably. 
This finding leads to the conclusion that the dense patches approach cannot handle more general cases, whereas fine invariant features provide much needed robustness to various imaging conditions.

In order to evaluate the effect of the pelage pattern extraction on the algorithm's accuracy, an ablation study has been performed. The results with and without the pattern extraction step are presented in Table~\ref{tab:nopattern}. It is clear that the pelage feature extraction significantly increases the accuracy of the algorithm.

\begin{table*}[ht!]
    \centering
    \caption{Comparison of re-identification results by the NORPPA method on the SealID dataset with and without the pattern extraction step.}
	\begin{center}
		\begin{tabular}{lcccc}
        \toprule
			Input data & TOP-1 & \quad TOP-5 \quad & TOP-10 & TOP-20\\
			\midrule
            Original images & 55.03\% & 68.48\% & 76.36\% & 84.73\% \\
            \midrule
			Pattern images & 77.64\% & 85.27\% & 89.09\% & 92.18\% \\
			\botrule
		\end{tabular}
		\label{tab:nopattern}
	\end{center}
\end{table*}

\subsection{Patch embedding network}

The following experiments were conducted in order to further improve the method:
\begin{enumerate}
  \item Training and fine-tuning of HardNet on different datasets,
  \item Various architecture modifications to the HardNet model.
\end{enumerate}

\subsubsection{Training and fine-tuning}

The original HardNet was trained on the union of HPatches~\citep{balntas2019hpatches} and Brown~\citep{brown_automatic_2007} datasets.
Typically, fine-tuning a machine learning model on domain-specific training data improves the method performance in a new domain. To test this on Saimaa ringed seal re-identification, we fine-tuned the HardNet model on patches of pelage pattern images. Fine-tuned models were compared to the pretrained model, a model trained from scratch on the pattern patches, and a model trained on the union of all datasets. 

The results are presented in Table~\ref{tab:ft_result}. For the training, all hyperparameters and random seeds were taken from the original implementation of HardNet~\citep{HardNet2017}.



\begin{table*}[ht]
\centering
\caption{Comparison of results for HardNet trained and fine-tuned on various datasets. We report mean with standard deviation.}
\label{tab:ft_result}
\begin{tabular}{lcccc} 
\toprule
 Training & Fine-tuning & \begin{tabular}[c]{@{}c@{}}Patches \\top-1\end{tabular} & \begin{tabular}[c]{@{}c@{}}Full\\top-1\end{tabular} & \begin{tabular}[c]{@{}c@{}}Full\\top-5\end{tabular}  \\ 
\midrule

 Pattern patches         & -        & $86.44\pm 0.41\%$                                                    & $59.86 \pm 1.36 \%$                                                     & $71.39 \pm 1.31 \%$                                                    \\ 

Brown+HPatches        & -        & $93.02 \pm 0.51\%$                                                & $\mathbf{77.19 \pm 0.93\%}$                                             & $\mathbf{85.11 \pm 0.77\%}$                                              \\ 

Brown+HPatches        & Pattern patches        & $\mathbf{93.76 \pm 0.12 \%}$                                        & $70.69 \pm 0.41 \%$                                               & $80.46 \pm 0.42 \%$                                               \\ 

\begin{tabular}[t]{@{}l@{}}Brown+HPatches+\\Pattern patches\end{tabular}    & -       & $92.48 \pm 0.94$\%                                                & $76.99 \pm 1.19$\%                                              & $84.85 \pm 1.03$\%                                               \\
\botrule
\end{tabular}
\end{table*}

While fine-tuning on the patches dataset improved the accuracy of the patch matching, the overall accuracy of the full-image matching dropped significantly. One possible reason is that the patches dataset was created using patches of the same scale, while the patches extracted by HesAffNet during the full re-identification algorithm vary in scale, leading to a different level of detail. 

Training on the union of all datasets showed no considerable improvements. This result can be explained by the size of the pelage pattern patches dataset in comparison to the combined sizes of the Brown and HPatches datasets. In other words, since HardNet utilizes triplet sampling during the training stage, the probability of an image from the pelage pattern dataset appearing in the triplet is extremely small. 

\subsubsection{Architecture modifications}

Several further modifications to the HardNet architecture were also considered. First, a Self-Organized Operational Neural Network (Self-ONN)~\citep{malik2021self} was incorporated into the HardNet model. Self-ONNs are networks consisting of layers that are the generalizations of convolutional layers. Simply put, each value in a convolutional kernel can be seen as a linear function, and this function can be generalized through Taylor series approximation with coefficients learned by the network. Such an approach leads to great nonlinearity even with shallow networks. Other modifications include the use of an EDEN pooling layer~\citep{chelak2021eden}, as well as changes to the number of channels and the output vector size.

The following models were evaluated: 
\begin{itemize}
    \item \textbf{HardNetONN}. This model has the same architecture as HardNet in terms of layers and number of channels in each layer. The only difference is that each convolution layer is replaced by a self ONN layer with a Taylor series degree equal to 3 for all layers, which leads to three times as many parameters.
    \item \textbf{HardNetONNDrop}. This model has the same architecture as HardNetONN but the last layer has a dropout with a probability of 0.3 similarly to the original HardNet.
    \item \textbf{HardNetONN + EDEN}. This model has the same architecture as HardNetONN, albeit that the last convolutional layer kernel is downsampled from $8 \times 8$ to $3 \times 3$ so that it would be possible to apply pooling. After the pooling a vector of size 128 is fed into a fully connected layer with the output size of 128, resulting in a compact embedding.
    \item \textbf{HardNetONNSmall}. This model has the same number of parameters as HardNet. All the layers were shrunk by half and the Taylor series degree was set to 4 for all layers. Consequently, the final vector has a size of 64 instead of 128.
    \item \textbf{HardNet3\_{384}}. This model is an original HardNet with 3 times as many channels in all of the layers. Therefore, it has an output vector of size 384 instead of 128. 
    \item \textbf{HardNet3\_{128}}. This model is the same as HardNet3\_{384} but with an output vector size of 128.
\end{itemize}
A comparison of the models is presented in Table~\ref{tab:onn}.

\begin{table*}[ht]
\centering
\caption{Comparison of the proposed models.}
\label{tab:onn}
\begin{tabular}{lcccc} 
\toprule
Method          & Parameters (M) & \begin{tabular}[c]{@{}c@{}}Patches\\TOP-1\end{tabular} & \begin{tabular}[c]{@{}c@{}}Full\\TOP-1\end{tabular} & \begin{tabular}[c]{@{}c@{}}Full\\TOP-5\end{tabular}  \\ 
\midrule
HardNet         & 1.3            & $93.02 \pm 0.51\%$                                                & $77.19 \pm 0.93\%$                                             & $85.11 \pm 0.77\%$                                              \\ 

HardNetONN + EDEN      & 1.3              & $91.73 \pm 0.48\%$                                                & $76.63 \pm 0.81\%$                                             & $84.5 \pm 0.47\%$                                              \\ 

HardNetONNSmall & 1.3            &  $90.48 \pm 0.68\%$                                                & $74.40 \pm 1.03\%$            &                                 $82.06 \pm 0.91\%$                                              \\ 

HardNetONN      & 4              &  $93.41 \pm 0.43\%$                                                & $78.38 \pm 0.35\%$                                            & $86.26 \pm 0.44\%$                                             \\ 

HardNetONNDrop  & 4              &  $92.55 \pm 0.31\%$                                                & $78.18 \pm 0.98\%$                                    & $85.37 \pm 0.63\%$                                              \\ 

HardNet3\_128   & 5.7            &  $93.50 \pm 0.54\%$                                                & $77.54 \pm 0.31\%$                  & \textbf{$86.18 \pm 0.44\%$}                                     \\

HardNet3\_384        & 12             & \bm{$94.30 \pm 0.56\%$}                                       & \bm{$78.75 \pm 0.33\%$}                            & \bm{$86.28 \pm 0.44\%$}                                              \\ 
\botrule
\end{tabular}
\end{table*}

\begin{figure}[ht!]
	\centering
	\includegraphics[width=1.0\linewidth]{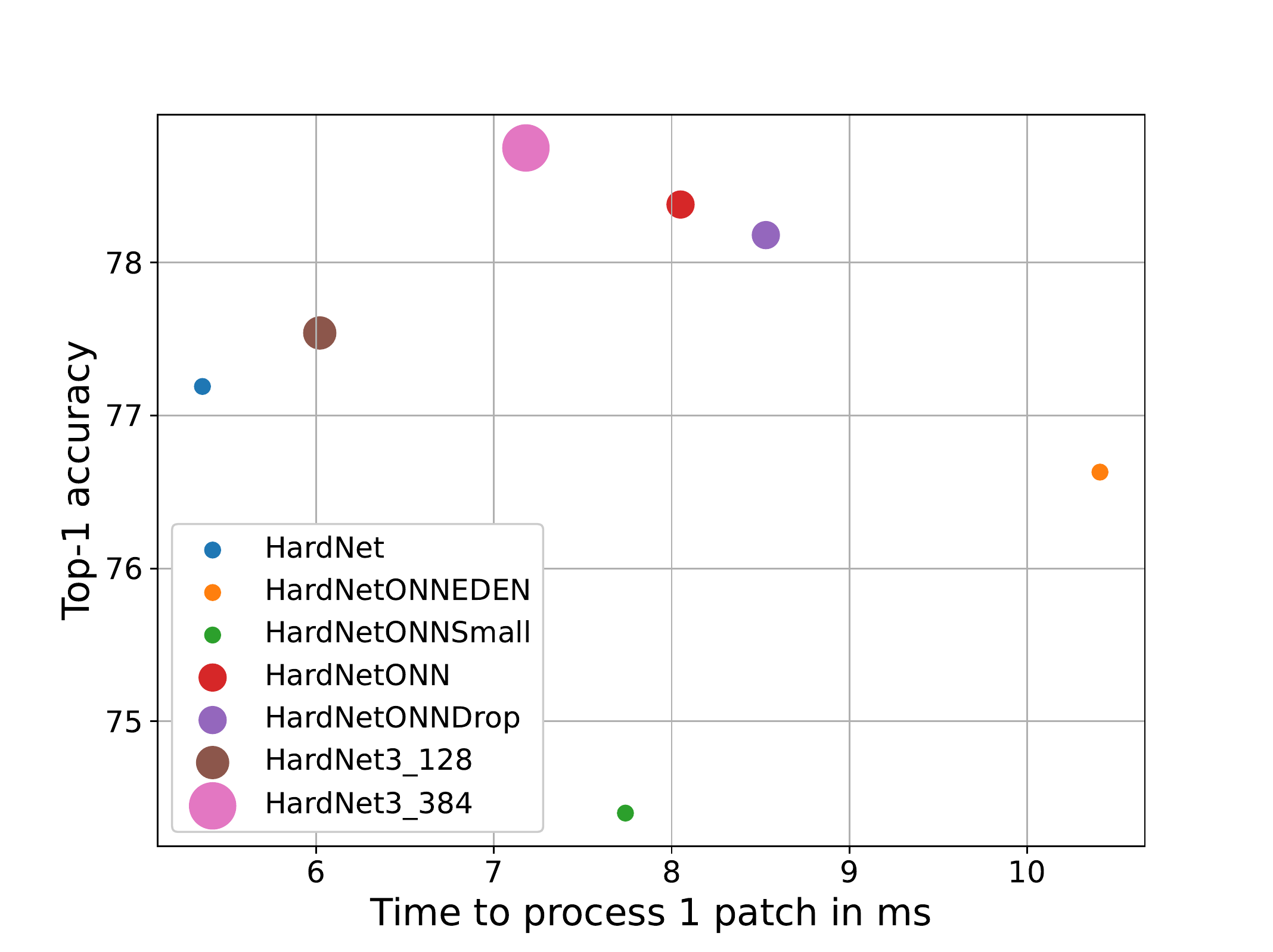}
	\caption{Plot of TOP-1 re-identification accuracy versus processing speed for all models. The circle size corresponds to the number of parameters in the model.}
	\label{fig:acc_plot}
\end{figure}

The HardNetONN and HardNet3\_{384} models show higher accuracy on both patch matching and full re-identification tasks than other versions of HardNet. Moreover, although HardNet3\_{384} has 12 million parameters and a vector size of 384, the difference of scores with HardNetONN is small, with the TOP-5 full re-identification score difference being negligible. A comparison of the processing speed of the models is presented in Fig.~\ref{fig:acc_plot}. Overall, the improvements over the baseline HardNet are rather small while result in a noticeable increase in computer time, limiting their usability in practice.


The accuracy of HardNetONNSmall is worse compared to HardNet, although it has the same number of parameters. This can be explained by the fact that the embedding vector is cut in half for HardNetONNSmall and may not contain enough information to learn a good metric. Additionally, HardNetONN + EDEN also scored lower than the original HardNet, although higher than HardNetONNSmall. The reason could lie in the redundancy of inductive bias provided by the pooling, as well as the worse convergence of the model.

\subsection{Qualitative evaluation}

\begin{figure}[ht!]
		\centering
		\includegraphics[width=\linewidth]{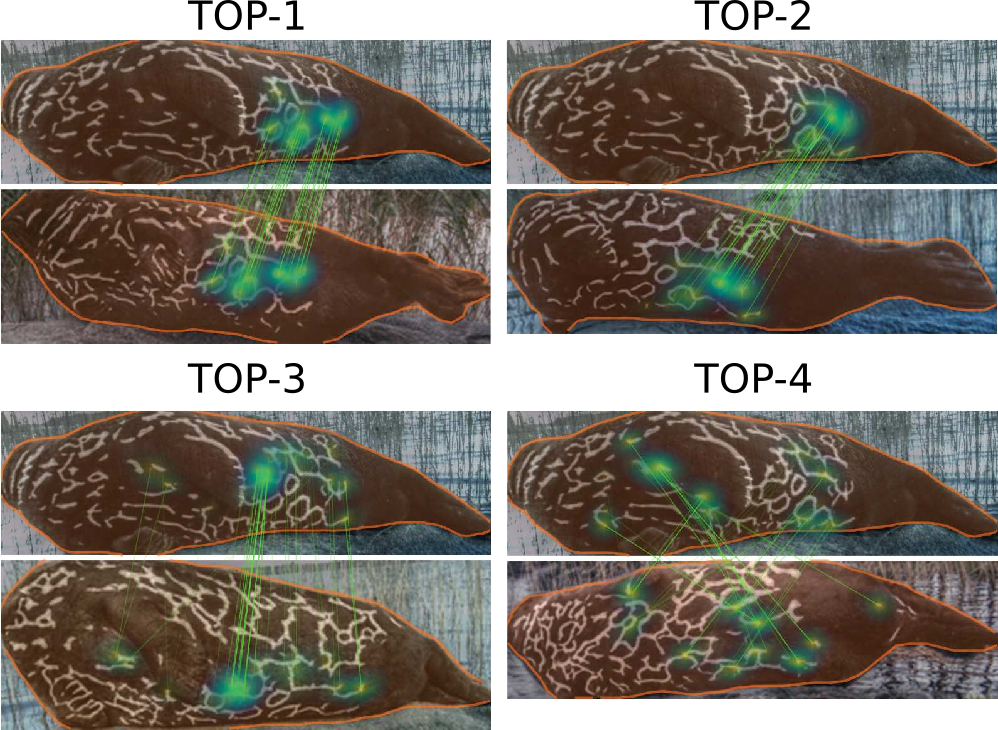}
		\caption{TOP-4 examples for the NORPPA method. First line: query image. Second line: four best matches in a decreasing order of similarity from left to right. Matched hotspots are highlighted in green. TOP-1--TOP-3 matches are correct. TOP-4 is incorrect.}
		\label{fig:fisher_example}
	\end{figure}

\begin{figure*}[ht!]
	\centering
	\subfloat[][]
	{
		\includegraphics[width=0.48\linewidth]{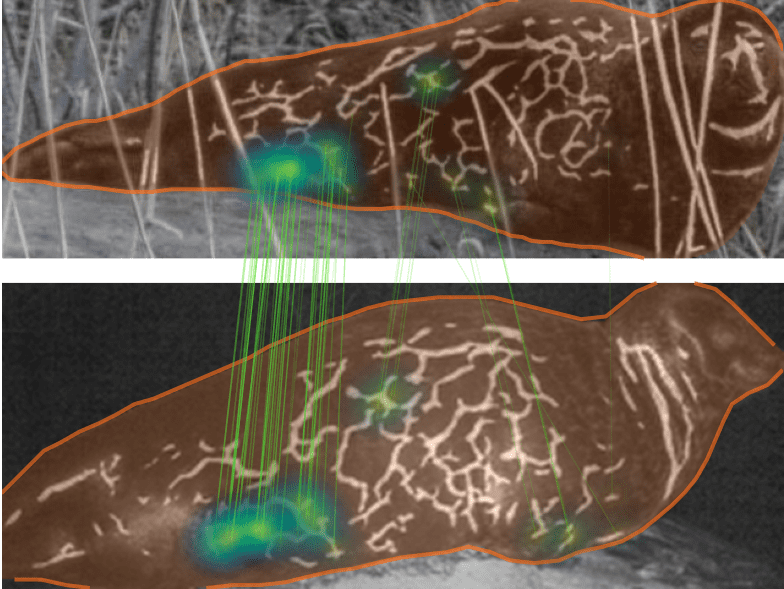}
		\label{subfig:h0}
	}
	\subfloat[][]
	{
		\includegraphics[width=0.48\linewidth]{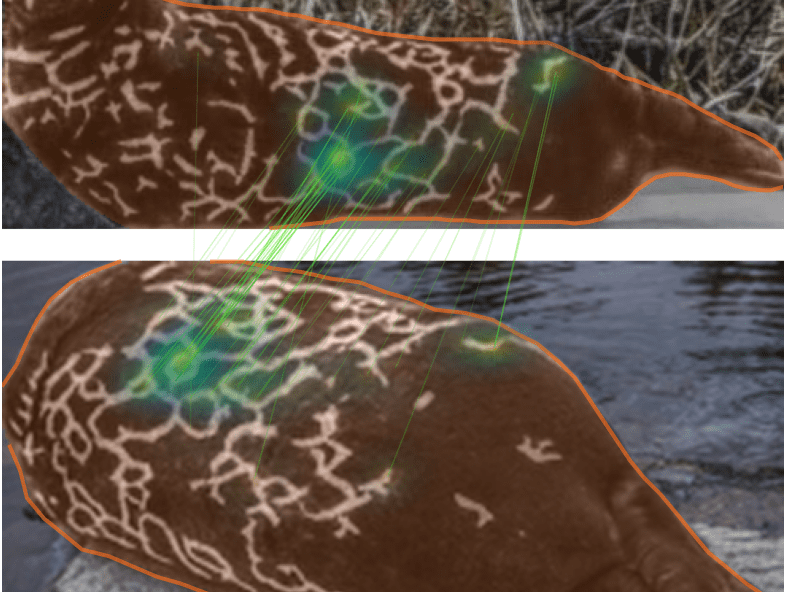}
		\label{subfig:h1}
	}\\
	\subfloat[][]
	{
		\includegraphics[width=0.31\linewidth]{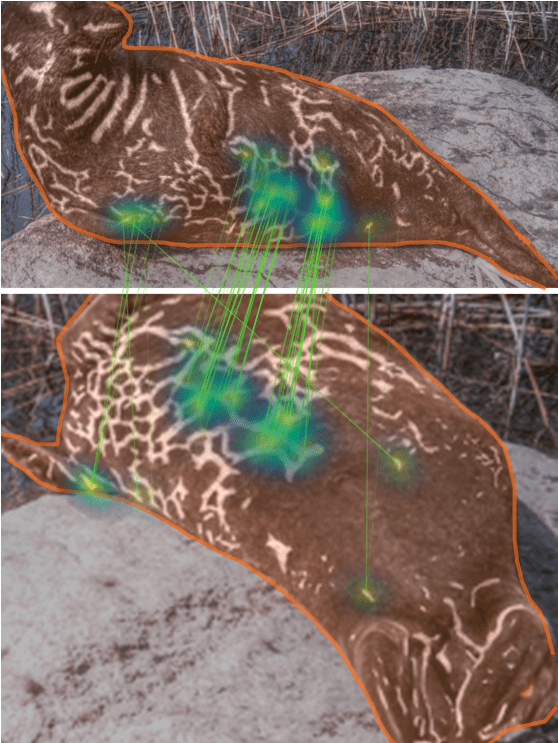}
		\label{subfig:v0}
	}
	\subfloat[][]
	{
		\includegraphics[width=0.31\linewidth]{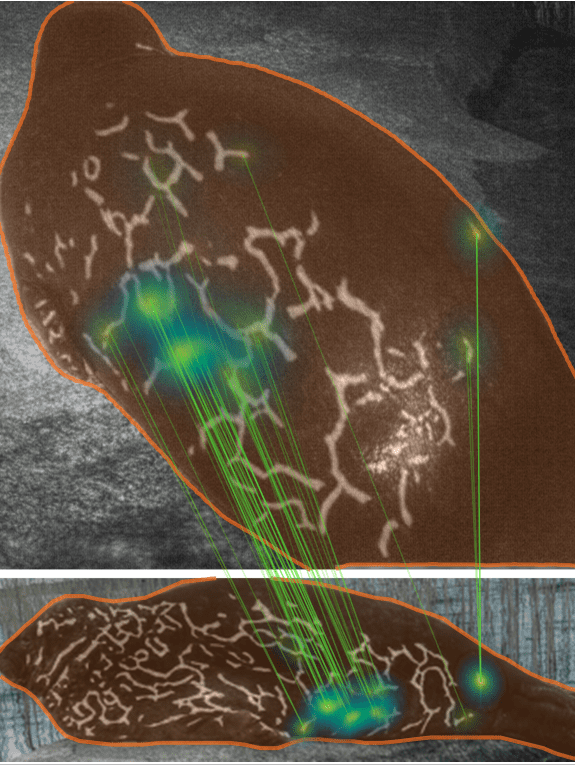}
		\label{subfig:v1}
	}
	\subfloat[][]
	{
		\includegraphics[width=0.31\linewidth]{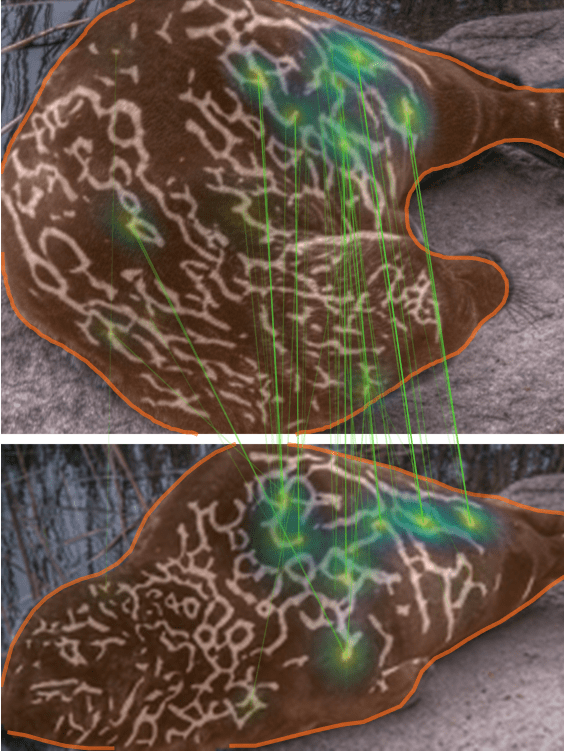}
		\label{subfig:v2}
	}
	
	\caption[]{Examples of some challenging cases. Top images are matched to the bottom images. The seal segmentation is shown in orange. The matching regions are highlighted and connected with green lines, the intensity corresponds to the similarity of a matched pair. The algorithm is able to match patterns even when the pose and point of view change significantly. (\subref{subfig:h0}) shows a successful match in the presence of some foreground obstacles.}
	\label{fig:challenging_cases}
\end{figure*}

Visual examples of the re-identification results for the proposed NORPPA method are presented in Fig.~\ref{fig:fisher_example}. For the final version we use HardNet trained on Brown and HPatches datasets. Upon inspecting the results with highlighted areas, it is evident that the proposed method learns to perform the matching between query and database images based on the characteristics of the pelage pattern. Furthermore, it can be seen that the method is able to find the corresponding regions in the patterns in very challenging cases (Fig.~\ref{fig:challenging_cases}).

\begin{table*}[ht!]
	\caption{Comparison of re-identification results between HotSpotter, NORPPA and previous iterations of the algorithm: SaimaaReID~\citep{nepovinnykh2020siamese} and LadogaReID~\citep{ladoga}.}
	\begin{center}
		\begin{tabular}{lccccc}
			\toprule
			Method & TOP-1 & TOP-2 & TOP-3 & TOP-4 & TOP-5\\
			\midrule
			SaimaaReID~\citep{nepovinnykh2020siamese} & 35.23\% & 41.45\% & 44.61\% & 47.92\% & 60.39\%\\
			
			LadogaReID~\citep{ladoga} & 39.94\% & 45.15\% & 50.06\% & 53.58\% & 56.67\%\\
			
			HotSpotter~\citep{hotspotter} & 61.87\% & 63.09\% & 63.63\% & 63.93\% &  64.42\%\\
			
			NORPPA (ours) & \textbf{77.64\%} & \textbf{80.91\%} & \textbf{82.97\%} & \textbf{84.18\%} & \textbf{85.27\%}\\
			
			\botrule
		\end{tabular}
		\label{tab:hotspotter}
	\end{center}
\end{table*}

\subsection{Quantitative evaluation}
SaimaaReID~\citep{nepovinnykh2020siamese}, LadogaReID~\citep{ladoga} without grouping step and NORPPA seal re-identification methods have been compared to HotSpotter\citep{hotspotter}, which is another method developed for patterned animal re-identification. HotSpotter is species-agnostic, and as such can be applied to Saimaa ringed seals as well. The results of NORPPA and HotSpotter for the Saimaa ringed seal dataset are presented in Table~\ref{tab:hotspotter}. It can be seen that the proposed method clearly outperforms HotSpotter based on TOP-1 accuracy. The difference is even more clear on TOP-5 accuracy, implying that even when NORPPA fails to correctly re-identify the seal, it is often able to provide a high rank for the correct match in the database. This is especially useful when the method is applied in a semi-supervised manner where the algorithm provides a set of possible matches for the expert to verify.

\begin{figure}[H]
	\centering
	\includegraphics[width=1\linewidth]{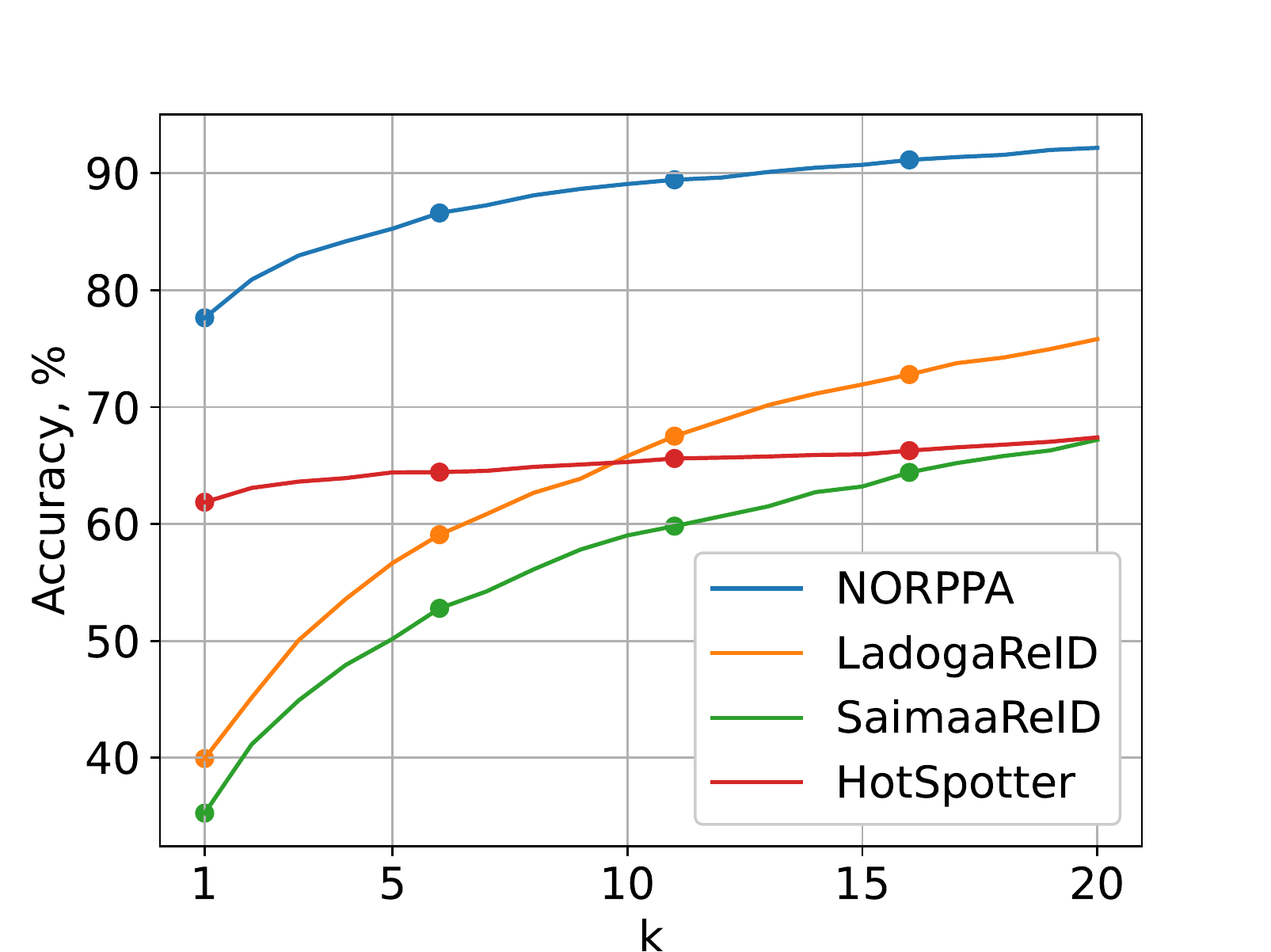}
	\caption{Plot of top-$k$ re-identification accuracy for the proposed NORPPA method relative to $k$.}
	\label{fig:topk_plot}
\end{figure}

By considering a larger number of top matches, it is possible to further increase the chances of finding a correct individual. The plot of the top-$k$ accuracy relative to the $k$ value is presented in Fig.~\ref{fig:topk_plot}. The relationship for the NORPPA, SaimaaReID and LadogaReID methods is logarithmic in nature with fast growth for small $k$ values, which slows down significantly with higher values.  HotSpotter, on the other hand, exhibits almost no improvement after TOP-2 accuracy, with the difference between TOP-1 and TOP-5 accuracy being only about 2\%, while the difference for NORPPA is almost 10\%. The improvement in accuracy is a desirable property for a semi-automatic approach, offering a considerable accuracy improvement in exchange for a relatively small increase in the manual work required (as compared to a fully manual approach). Depending on the final application and available data, the relationship between the top-$k$ accuracy and $k$ can be used to determine the optimal number of matches to be returned by the algorithm.

\section{Conclusion}
A novel method for Saimaa ringed seal re-identification called NOvel Ringed seal re-identification by Pelage Pattern Aggregation (NORPPA) was proposed in this paper. The method utilizes pelage pattern extraction and feature aggregation inspired by content-based image retrieval techniques. The re-identification pipeline consists of image enhancement, seal instance segmentation by Mask R-CNN, U-net based pelage pattern extraction, pattern feature extraction, feature aggregation, and individual re-identification by database search. 
Improved pattern feature embeddings were proposed by employing affine-invariant region of interest detection, CNN based feature descriptors, and Fisher Vector feature aggregation to obtain fixed size embedding vectors with high discriminative power. 
The proposed method was applied to a novel and challenging Saimaa ringed seal dataset and showed superior performance compared to HotSpotter and earlier versions of the Saimaa ringed seal re-identification method by the authors. 
One additional benefit of the proposed method is that it allows features to be aggregated over multiple images. This opens interesting possibilities for further research as sequences of game camera images can be utilized to create a single descriptor for a larger portion of a pelage pattern by filling in the gaps created by obstructions and viewpoints.
While the method was developed for Saimaa ringed seals, it is also possible to apply it on other patterned animal species.

\bmhead{Acknowledgments}

The authors would like to thank Raija ja Ossi Tuuliaisen Säätiö Foundation, the project CoExist (Project ID: KS1549) for funding the research. In addition, authors would like to thank Vincent Biard, Piia Mutka, Marja Niemi, and Mervi Kunnasranta from the Department of Environmental and Biological Sciences at the University of Eastern Finland (UEF) for providing the data of Saimaa ringed seals and their expert knowledge of identifying each individual.

\section*{Declarations}
\subsection*{Funding}
The research is a part of project CoExist (Project ID: KS1549) funded by the European Union, the Russian Federation and the Republic of Finland via The Southeast Finland--Russia CBC 2014-2020 programme for funding the research.

\subsection*{Conflict of interest}
We declare no competing interests.

\subsection*{Ethics approval}
Data collection was done under permits by the Finnish environmental authorities ELY centre (ESAELY/1290/2015, POKELY/1232/2015, KASELY/2014/2015 and POSELY/313/07.01/2012) and Metsähallitus (MH 5813/2013 and MH 6377/2018/05.04.01).

\subsection*{Consent for publication}
All authors consent that the publisher has the author’s permission to publish research findings. All authors guarantee that the research findings have not been previously published.

\subsection*{Availability of data and materials}
All data and materials are publicly available at \burl{https://doi.org/10.23729/0f4a3296-3b10-40c8-9ad3-0cf00a5a4a53}

\subsection*{Code availability}
The codes for the described experiments are available at \burl{https://github.com/kwadraterry/Norppa}

\subsection*{Authors' contributions}

T. Eerola and H. K\"alvi\"ainen were responsible for the supervision of the research, designing methodology, and project administration;
E.Nepovinnykh and I.Chelak implemented the algorithm. E.Nepovinnykh, I.Chelak, T.Eerola, and H. K\"alvi\"ainen prepared the original draft of the manuscript. All the authors gave the final approval for publication.

\bibliography{manuscript}

\begin{thebibliography}{}
\providecommand{\doi}[1]{\url{https://doi.org/#1}}
\bibcommenthead

\bibitem [\protect \citeauthoryear {%
Agarwal%
\ \protect \BOthers {.}}{%
Agarwal%
\ \protect \BOthers {.}}{%
{\protect \APACyear {2019}}%
}]{%
agarwal2019triplet}
\APACinsertmetastar {%
agarwal2019triplet}%
\begin{APACrefauthors}%
Agarwal, M.%
, Sinha, S.%
, Singh, M.%
, Nagpal, S.%
, Singh, R.%
\BCBL {} Vatsa, M.%
\end{APACrefauthors}%
\unskip\
\newblock
\APACrefYearMonthDay{2019}{}{}.
\newblock
{\BBOQ}\APACrefatitle {Triplet Transform Learning for Automated Primate Face
  Recognition} {Triplet transform learning for automated primate face
  recognition}.{\BBCQ}
\newblock
 \APACrefbtitle {{International} {Conference} on {Image} {Processing}
  ({ICIP}).} {{International} {Conference} on {Image} {Processing} ({ICIP}).}
\newblock
\begin{APACrefDOI}\doi{https://doi.org/10.1109/ICIP.2019.8803501}\end{APACrefDOI}
\PrintBackRefs{\CurrentBib}

\bibitem [\protect \citeauthoryear {%
Amari%
\ \BBA {} Nagaoka%
}{%
Amari%
\ \BBA {} Nagaoka%
}{%
{\protect \APACyear {2000}}%
}]{%
amari2000methods}
\APACinsertmetastar {%
amari2000methods}%
\begin{APACrefauthors}%
Amari, S.%
\BCBT {}\ \BBA {} Nagaoka, H.%
\end{APACrefauthors}%
\unskip\
\newblock
\APACrefYear{2000}.
\newblock
\APACrefbtitle {Methods of Information Geometry} {Methods of information
  geometry}.
\newblock
\APACaddressPublisher{}{American Mathematical Society}.
\PrintBackRefs{\CurrentBib}

\bibitem [\protect \citeauthoryear {%
Arandjelovic%
, Gronat%
, Torii%
, Pajdla%
\BCBL {}\ \BBA {} Sivic%
}{%
Arandjelovic%
\ \protect \BOthers {.}}{%
{\protect \APACyear {2016}}%
}]{%
arandjelovic2016netvlad}
\APACinsertmetastar {%
arandjelovic2016netvlad}%
\begin{APACrefauthors}%
Arandjelovic, R.%
, Gronat, P.%
, Torii, A.%
, Pajdla, T.%
\BCBL {} Sivic, J.%
\end{APACrefauthors}%
\unskip\
\newblock
\APACrefYearMonthDay{2016}{}{}.
\newblock
{\BBOQ}\APACrefatitle {{NetVLAD}: {CNN} {Architecture} for {Weakly}
  {Supervised} {Place} {Recognition}} {{NetVLAD}: {CNN} {Architecture} for
  {Weakly} {Supervised} {Place} {Recognition}}.{\BBCQ}
\newblock
 \APACrefbtitle {{Conference} on {Computer} {Vision} and {Pattern}
  {Recognition} ({CVPR}).} {{Conference} on {Computer} {Vision} and {Pattern}
  {Recognition} ({CVPR}).}
\newblock
\begin{APACrefDOI}\doi{https://doi.org/10.1109/CVPR.2016.572}\end{APACrefDOI}
\PrintBackRefs{\CurrentBib}

\bibitem [\protect \citeauthoryear {%
Arandjelovi\'c%
\ \BBA {} Zisserman%
}{%
Arandjelovi\'c%
\ \BBA {} Zisserman%
}{%
{\protect \APACyear {2012}}%
}]{%
arandjelovic2012three}
\APACinsertmetastar {%
arandjelovic2012three}%
\begin{APACrefauthors}%
Arandjelovi\'c, R.%
\BCBT {}\ \BBA {} Zisserman, A.%
\end{APACrefauthors}%
\unskip\
\newblock
\APACrefYearMonthDay{2012}{}{}.
\newblock
{\BBOQ}\APACrefatitle {Three things everyone should know to improve object
  retrieval} {Three things everyone should know to improve object
  retrieval}.{\BBCQ}
\newblock
 \APACrefbtitle {{Conference} on {Computer} {Vision} and {Pattern}
  {Recognition} ({CVPR}).} {{Conference} on {Computer} {Vision} and {Pattern}
  {Recognition} ({CVPR}).}
\newblock
\begin{APACrefDOI}\doi{https://doi.org/10.1109/CVPR.2012.6248018}\end{APACrefDOI}
\PrintBackRefs{\CurrentBib}

\bibitem [\protect \citeauthoryear {%
Araujo%
\ \protect \BOthers {.}}{%
Araujo%
\ \protect \BOthers {.}}{%
{\protect \APACyear {2020}}%
}]{%
araujo2020getting}
\APACinsertmetastar {%
araujo2020getting}%
\begin{APACrefauthors}%
Araujo, G.%
, Ismail, A.%
, McCann, C.%
, McCann, D.%
, Legaspi, C.%
, Snow, S.%
\BDBL {}Ponzo, A.%
\end{APACrefauthors}%
\unskip\
\newblock
\APACrefYearMonthDay{2020}{}{}.
\newblock
{\BBOQ}\APACrefatitle {{Getting the most out of citizen science for endangered
  species such as Whale Shark}} {{Getting the most out of citizen science for
  endangered species such as Whale Shark}}.{\BBCQ}
\newblock
\APACjournalVolNumPages{Journal of Fish Biology}{96}{}{864--867}.
\newblock
\begin{APACrefDOI}\doi{https://doi.org/10.1111/jfb.14254}\end{APACrefDOI}
\PrintBackRefs{\CurrentBib}

\bibitem [\protect \citeauthoryear {%
Babenko%
\ \BBA {} Lempitsky%
}{%
Babenko%
\ \BBA {} Lempitsky%
}{%
{\protect \APACyear {2015}}%
}]{%
babenko2015aggregating}
\APACinsertmetastar {%
babenko2015aggregating}%
\begin{APACrefauthors}%
Babenko, A.%
\BCBT {}\ \BBA {} Lempitsky, V.%
\end{APACrefauthors}%
\unskip\
\newblock
\APACrefYearMonthDay{2015}{}{}.
\newblock
{\BBOQ}\APACrefatitle {{Aggregating Deep Convolutional Features for Image
  Retrieval}} {{Aggregating Deep Convolutional Features for Image
  Retrieval}}.{\BBCQ}
\newblock
 \APACrefbtitle {{International} {Conference} on {Computer} {Vision} ({ICCV}).}
  {{International} {Conference} on {Computer} {Vision} ({ICCV}).}
\newblock
\begin{APACrefDOI}\doi{https://doi.org/10.1109/iccv.2015.150}\end{APACrefDOI}
\PrintBackRefs{\CurrentBib}

\bibitem [\protect \citeauthoryear {%
Babenko%
, Slesarev%
, Chigorin%
\BCBL {}\ \BBA {} Lempitsky%
}{%
Babenko%
\ \protect \BOthers {.}}{%
{\protect \APACyear {2014}}%
}]{%
babenko2014neural}
\APACinsertmetastar {%
babenko2014neural}%
\begin{APACrefauthors}%
Babenko, A.%
, Slesarev, A.%
, Chigorin, A.%
\BCBL {} Lempitsky, V.%
\end{APACrefauthors}%
\unskip\
\newblock
\APACrefYearMonthDay{2014}{}{}.
\newblock
{\BBOQ}\APACrefatitle {{Neural Codes for Image Retrieval}} {{Neural Codes for
  Image Retrieval}}.{\BBCQ}
\newblock
 \APACrefbtitle {European {Conference} on {Computer} {Vision} ({ECCV}).}
  {European {Conference} on {Computer} {Vision} ({ECCV}).}
\newblock
\begin{APACrefDOI}\doi{https://doi.org/10.1007/978-3-319-10590-1_38}\end{APACrefDOI}
\PrintBackRefs{\CurrentBib}

\bibitem [\protect \citeauthoryear {%
Badreldeen Bdawy~Mohamed%
}{%
Badreldeen Bdawy~Mohamed%
}{%
{\protect \APACyear {2021}}%
}]{%
badreldeen2021metric}
\APACinsertmetastar {%
badreldeen2021metric}%
\begin{APACrefauthors}%
Badreldeen Bdawy~Mohamed, O.%
\end{APACrefauthors}%
\unskip\
\newblock
\APACrefYearMonthDay{2021}{}{}.
\newblock
\APACrefbtitle {Metric learning based pattern matching for species agnostic
  animal re-identification.} {Metric learning based pattern matching for
  species agnostic animal re-identification.}
\newblock
\APAChowpublished {Master's thesis, Lappeenranta-Lahti University of Technology
  LUT, Finland}.
\PrintBackRefs{\CurrentBib}

\bibitem [\protect \citeauthoryear {%
Balntas%
, Lenc%
, Vedaldi%
\BCBL {}\ \BBA {} Mikolajczyk%
}{%
Balntas%
\ \protect \BOthers {.}}{%
{\protect \APACyear {2017}}%
}]{%
balntas2019hpatches}
\APACinsertmetastar {%
balntas2019hpatches}%
\begin{APACrefauthors}%
Balntas, V.%
, Lenc, K.%
, Vedaldi, A.%
\BCBL {} Mikolajczyk, K.%
\end{APACrefauthors}%
\unskip\
\newblock
\APACrefYearMonthDay{2017}{}{}.
\newblock
{\BBOQ}\APACrefatitle {HPatches: A benchmark and evaluation of handcrafted and
  learned local descriptors} {Hpatches: A benchmark and evaluation of
  handcrafted and learned local descriptors}.{\BBCQ}
\newblock
 \APACrefbtitle {{Conference} on {Computer} {Vision} and {Pattern}
  {Recognition} ({CVPR}).} {{Conference} on {Computer} {Vision} and {Pattern}
  {Recognition} ({CVPR}).}
\newblock
\begin{APACrefDOI}\doi{https://doi.org/10.1109/TPAMI.2019.2915233}\end{APACrefDOI}
\PrintBackRefs{\CurrentBib}

\bibitem [\protect \citeauthoryear {%
Baumberg%
}{%
Baumberg%
}{%
{\protect \APACyear {2000}}%
}]{%
Baumberg2000}
\APACinsertmetastar {%
Baumberg2000}%
\begin{APACrefauthors}%
Baumberg, A.%
\end{APACrefauthors}%
\unskip\
\newblock
\APACrefYearMonthDay{2000}{}{}.
\newblock
{\BBOQ}\APACrefatitle {Reliable feature matching across widely separated views}
  {Reliable feature matching across widely separated views}.{\BBCQ}
\newblock
 \APACrefbtitle {{Conference} on {Computer} {Vision} and {Pattern}
  {Recognition} ({CVPR}).} {{Conference} on {Computer} {Vision} and {Pattern}
  {Recognition} ({CVPR}).}
\newblock
\begin{APACrefDOI}\doi{https://doi.org/10.1109/CVPR.2000.855899}\end{APACrefDOI}
\PrintBackRefs{\CurrentBib}

\bibitem [\protect \citeauthoryear {%
T.~Berger-Wolf%
\ \protect \BOthers {.}}{%
T.~Berger-Wolf%
\ \protect \BOthers {.}}{%
{\protect \APACyear {2015}}%
}]{%
berger2015ibeis}
\APACinsertmetastar {%
berger2015ibeis}%
\begin{APACrefauthors}%
Berger-Wolf, T.%
, Rubenstein, D.%
, Stewart, C.%
, Holmberg, J.%
, Parham, J.%
\BCBL {} Crall, J.%
\end{APACrefauthors}%
\unskip\
\newblock
\APACrefYearMonthDay{2015}{}{}.
\newblock
{\BBOQ}\APACrefatitle {Ibeis: Image-based ecological information system: From
  pixels to science and conservation} {Ibeis: Image-based ecological
  information system: From pixels to science and conservation}.{\BBCQ}
\newblock
 \APACrefbtitle {{Bloomberg Data for Good Exchange Conference}.} {{Bloomberg
  Data for Good Exchange Conference}.}
\PrintBackRefs{\CurrentBib}

\bibitem [\protect \citeauthoryear {%
T.Y.~Berger-Wolf%
\ \protect \BOthers {.}}{%
T.Y.~Berger-Wolf%
\ \protect \BOthers {.}}{%
{\protect \APACyear {2017}}%
}]{%
berger2017wildbook}
\APACinsertmetastar {%
berger2017wildbook}%
\begin{APACrefauthors}%
Berger-Wolf, T.Y.%
, Rubenstein, D.I.%
, Stewart, C.V.%
, Holmberg, J.A.%
, Parham, J.%
, Menon, S.%
\BDBL {}Joppa, L.%
\end{APACrefauthors}%
\unskip\
\newblock
\APACrefYearMonthDay{2017}{}{}.
\newblock
{\BBOQ}\APACrefatitle {Wildbook: {Crowdsourcing}, computer vision, and data
  science for conservation} {Wildbook: {Crowdsourcing}, computer vision, and
  data science for conservation}.{\BBCQ}
\newblock
\APACjournalVolNumPages{arXiv preprint arXiv:1710.08880}{}{}{}.
\PrintBackRefs{\CurrentBib}

\bibitem [\protect \citeauthoryear {%
Bergstra%
, Bardenet%
, Bengio%
\BCBL {}\ \BBA {} K\'{e}gl%
}{%
Bergstra%
\ \protect \BOthers {.}}{%
{\protect \APACyear {2011}}%
}]{%
bergstra2011algorithms}
\APACinsertmetastar {%
bergstra2011algorithms}%
\begin{APACrefauthors}%
Bergstra, J.%
, Bardenet, R.%
, Bengio, Y.%
\BCBL {} K\'{e}gl, B.%
\end{APACrefauthors}%
\unskip\
\newblock
\APACrefYearMonthDay{2011}{}{}.
\newblock
{\BBOQ}\APACrefatitle {{Algorithms for Hyper-Parameter Optimization}}
  {{Algorithms for Hyper-Parameter Optimization}}.{\BBCQ}
\newblock
 \APACrefbtitle {{Conference on Neural Information Processing Systems
  (NeurIPS)}.} {{Conference on Neural Information Processing Systems
  (NeurIPS)}.}
\PrintBackRefs{\CurrentBib}

\bibitem [\protect \citeauthoryear {%
Bogucki%
\ \protect \BOthers {.}}{%
Bogucki%
\ \protect \BOthers {.}}{%
{\protect \APACyear {2019}}%
}]{%
bogucki2019applying}
\APACinsertmetastar {%
bogucki2019applying}%
\begin{APACrefauthors}%
Bogucki, R.%
, Cygan, M.%
, Khan, C.B.%
, Klimek, M.%
, Milczek, J.K.%
\BCBL {} Mucha, M.%
\end{APACrefauthors}%
\unskip\
\newblock
\APACrefYearMonthDay{2019}{}{}.
\newblock
{\BBOQ}\APACrefatitle {Applying deep learning to right whale photo
  identification} {Applying deep learning to right whale photo
  identification}.{\BBCQ}
\newblock
\APACjournalVolNumPages{Conservation Biology}{33}{}{676--684}.
\newblock
\begin{APACrefDOI}\doi{https://doi.org/10.1111/cobi.13226}\end{APACrefDOI}
\PrintBackRefs{\CurrentBib}

\bibitem [\protect \citeauthoryear {%
Brown%
\ \BBA {} Lowe%
}{%
Brown%
\ \BBA {} Lowe%
}{%
{\protect \APACyear {2007}}%
}]{%
brown_automatic_2007}
\APACinsertmetastar {%
brown_automatic_2007}%
\begin{APACrefauthors}%
Brown, M.%
\BCBT {}\ \BBA {} Lowe, D.G.%
\end{APACrefauthors}%
\unskip\
\newblock
\APACrefYearMonthDay{2007}{}{}.
\newblock
{\BBOQ}\APACrefatitle {Automatic {Panoramic} {Image} {Stitching} using
  {Invariant} {Features}} {Automatic {Panoramic} {Image} {Stitching} using
  {Invariant} {Features}}.{\BBCQ}
\newblock
\APACjournalVolNumPages{International Journal of Computer
  Vision}{74}{}{59--73}.
\newblock
\begin{APACrefDOI}\doi{https://doi.org/10.1007/s11263-006-0002-3}\end{APACrefDOI}
\PrintBackRefs{\CurrentBib}

\bibitem [\protect \citeauthoryear {%
Brust%
\ \protect \BOthers {.}}{%
Brust%
\ \protect \BOthers {.}}{%
{\protect \APACyear {2017}}%
}]{%
brust2017towards}
\APACinsertmetastar {%
brust2017towards}%
\begin{APACrefauthors}%
Brust, C\BHBI A.%
, Burghardt, T.%
, Groenenberg, M.%
, Kading, C.%
, Kuhl, H.S.%
, Manguette, M.L.%
\BCBL {} Denzler, J.%
\end{APACrefauthors}%
\unskip\
\newblock
\APACrefYearMonthDay{2017}{}{}.
\newblock
{\BBOQ}\APACrefatitle {Towards automated visual monitoring of individual
  gorillas in the wild} {Towards automated visual monitoring of individual
  gorillas in the wild}.{\BBCQ}
\newblock
 \APACrefbtitle {{International} {Conference} on {Computer} {Vision} {Workshop}
  ({ICCVW}).} {{International} {Conference} on {Computer} {Vision} {Workshop}
  ({ICCVW}).}
\newblock
\begin{APACrefDOI}\doi{https://doi.org/10.1109/iccvw.2017.333}\end{APACrefDOI}
\PrintBackRefs{\CurrentBib}

\bibitem [\protect \citeauthoryear {%
Cheeseman%
, Johnson%
\BCBL {}\ \BBA {} Muldavin%
}{%
Cheeseman%
\ \protect \BOthers {.}}{%
{\protect \APACyear {2017}}%
}]{%
cheeseman2017happywhale}
\APACinsertmetastar {%
cheeseman2017happywhale}%
\begin{APACrefauthors}%
Cheeseman, T.%
, Johnson, T.%
\BCBL {} Muldavin, N.%
\end{APACrefauthors}%
\unskip\
\newblock
\APACrefYearMonthDay{2017}{}{}.
\newblock
\APACrefbtitle {Happywhale: Globalizing Marine Mammal Photo Identification via
  a Citizen Science Web Platform.} {Happywhale: Globalizing marine mammal photo
  identification via a citizen science web platform.}
\newblock
\APAChowpublished {{Paper SC/67A/PH/02 presented to the Scientific Committee of
  the Report to the International Whaling Commission}}.
\PrintBackRefs{\CurrentBib}

\bibitem [\protect \citeauthoryear {%
Chehrsimin%
\ \protect \BOthers {.}}{%
Chehrsimin%
\ \protect \BOthers {.}}{%
{\protect \APACyear {2018}}%
}]{%
chehrsimin2017automatic}
\APACinsertmetastar {%
chehrsimin2017automatic}%
\begin{APACrefauthors}%
Chehrsimin, T.%
, Eerola, T.%
, Koivuniemi, M.%
, Auttila, M.%
, Lev{\"a}nen, R.%
, Niemi, M.%
\BDBL {}K{\"a}lvi{\"a}inen, H.%
\end{APACrefauthors}%
\unskip\
\newblock
\APACrefYearMonthDay{2018}{}{}.
\newblock
{\BBOQ}\APACrefatitle {{Automatic individual identification of Saimaa ringed
  seals}} {{Automatic individual identification of Saimaa ringed
  seals}}.{\BBCQ}
\newblock
\APACjournalVolNumPages{IET Computer Vision}{12}{}{146--152}.
\newblock
\begin{APACrefDOI}\doi{https://doi.org/10.1049/iet-cvi.2017.0082}\end{APACrefDOI}
\PrintBackRefs{\CurrentBib}

\bibitem [\protect \citeauthoryear {%
Chelak%
, Nepovinnykh%
, Eerola%
, K\"alvi\"ainen%
\BCBL {}\ \BBA {} Belykh%
}{%
Chelak%
\ \protect \BOthers {.}}{%
{\protect \APACyear {2021}}%
}]{%
chelak2021eden}
\APACinsertmetastar {%
chelak2021eden}%
\begin{APACrefauthors}%
Chelak, I.%
, Nepovinnykh, E.%
, Eerola, T.%
, K\"alvi\"ainen, H.%
\BCBL {} Belykh, I.%
\end{APACrefauthors}%
\unskip\
\newblock
\APACrefYearMonthDay{2021}{}{}.
\newblock
{\BBOQ}\APACrefatitle {{EDEN: Deep Feature Distribution Pooling for Saimaa
  Ringed Seals Pattern Matching}} {{EDEN: Deep Feature Distribution Pooling for
  Saimaa Ringed Seals Pattern Matching}}.{\BBCQ}
\newblock
\APACjournalVolNumPages{arXiv preprint arXiv:2105.13979}{}{}{}.
\PrintBackRefs{\CurrentBib}

\bibitem [\protect \citeauthoryear {%
Crall%
, Stewart%
, Berger-Wolf%
, Rubenstein%
\BCBL {}\ \BBA {} Sundaresan%
}{%
Crall%
\ \protect \BOthers {.}}{%
{\protect \APACyear {2013}}%
}]{%
hotspotter}
\APACinsertmetastar {%
hotspotter}%
\begin{APACrefauthors}%
Crall, J.%
, Stewart, C.%
, Berger-Wolf, T.%
, Rubenstein, D.%
\BCBL {} Sundaresan, S.%
\end{APACrefauthors}%
\unskip\
\newblock
\APACrefYearMonthDay{2013}{}{}.
\newblock
{\BBOQ}\APACrefatitle {HotSpotter - Patterned species instance recognition}
  {Hotspotter - patterned species instance recognition}.{\BBCQ}
\newblock
 \APACrefbtitle {{Winter Conference} on {Applications} of {Computer Vision
  (WACV)}.} {{Winter Conference} on {Applications} of {Computer Vision
  (WACV)}.}
\newblock
\begin{APACrefDOI}\doi{https://doi.org/10.1109/2013.6475023}\end{APACrefDOI}
\PrintBackRefs{\CurrentBib}

\bibitem [\protect \citeauthoryear {%
Crouse%
\ \protect \BOthers {.}}{%
Crouse%
\ \protect \BOthers {.}}{%
{\protect \APACyear {2017}}%
}]{%
crouse2017lemur}
\APACinsertmetastar {%
crouse2017lemur}%
\begin{APACrefauthors}%
Crouse, D.%
, Jacobs, R.%
, Richardson, Z.%
, Klum, S.%
, Jain, A.%
, Baden, A.%
\BCBL {} Tecot, S.%
\end{APACrefauthors}%
\unskip\
\newblock
\APACrefYearMonthDay{2017}{}{}.
\newblock
{\BBOQ}\APACrefatitle {LemurFaceID: A face recognition system to facilitate
  individual identification of lemurs} {Lemurfaceid: A face recognition system
  to facilitate individual identification of lemurs}.{\BBCQ}
\newblock
\APACjournalVolNumPages{BMC Zoology}{2}{}{1--14}.
\newblock
\begin{APACrefDOI}\doi{https://doi.org/10.1186/s40850-016-0011-9}\end{APACrefDOI}
\PrintBackRefs{\CurrentBib}

\bibitem [\protect \citeauthoryear {%
Deb%
\ \protect \BOthers {.}}{%
Deb%
\ \protect \BOthers {.}}{%
{\protect \APACyear {2018}}%
}]{%
deb2018face}
\APACinsertmetastar {%
deb2018face}%
\begin{APACrefauthors}%
Deb, D.%
, Wiper, S.%
, Gong, S.%
, Shi, Y.%
, Tymoszek, C.%
, Fletcher, A.%
\BCBL {} Jain, A.K.%
\end{APACrefauthors}%
\unskip\
\newblock
\APACrefYearMonthDay{2018}{}{}.
\newblock
{\BBOQ}\APACrefatitle {Face recognition: Primates in the wild} {Face
  recognition: Primates in the wild}.{\BBCQ}
\newblock
 \APACrefbtitle {{International Conference} on {Biometrics Theory, Applications
  and Systems (BTAS)}.} {{International Conference} on {Biometrics Theory,
  Applications and Systems (BTAS)}.}
\newblock
\begin{APACrefDOI}\doi{https://doi.org/10.1109/btas.2018.8698538}\end{APACrefDOI}
\PrintBackRefs{\CurrentBib}

\bibitem [\protect \citeauthoryear {%
Fischler%
\ \BBA {} Bolles%
}{%
Fischler%
\ \BBA {} Bolles%
}{%
{\protect \APACyear {1981}}%
}]{%
RANSAC}
\APACinsertmetastar {%
RANSAC}%
\begin{APACrefauthors}%
Fischler, M.A.%
\BCBT {}\ \BBA {} Bolles, R.C.%
\end{APACrefauthors}%
\unskip\
\newblock
\APACrefYearMonthDay{1981}{}{}.
\newblock
{\BBOQ}\APACrefatitle {Random sample consensus: a paradigm for model fitting
  with applications to image analysis and automated cartography} {Random sample
  consensus: a paradigm for model fitting with applications to image analysis
  and automated cartography}.{\BBCQ}
\newblock
\APACjournalVolNumPages{Communications of the ACM}{24}{}{381--395}.
\newblock
\begin{APACrefDOI}\doi{https://doi.org/10.1145/358669.358692}\end{APACrefDOI}
\PrintBackRefs{\CurrentBib}

\bibitem [\protect \citeauthoryear {%
Freytag%
\ \protect \BOthers {.}}{%
Freytag%
\ \protect \BOthers {.}}{%
{\protect \APACyear {2016}}%
}]{%
freytag2016Chimpanzee}
\APACinsertmetastar {%
freytag2016Chimpanzee}%
\begin{APACrefauthors}%
Freytag, A.%
, Rodner, E.%
, Simon, M.%
, Loos, A.%
, K\"uhl, H.%
\BCBL {} Denzler, J.%
\end{APACrefauthors}%
\unskip\
\newblock
\APACrefYearMonthDay{2016}{}{}.
\newblock
{\BBOQ}\APACrefatitle {{Chimpanzee Faces in the Wild: Log-Euclidean CNNs for
  Predicting Identities and Attributes of Primates}} {{Chimpanzee Faces in the
  Wild: Log-Euclidean CNNs for Predicting Identities and Attributes of
  Primates}}.{\BBCQ}
\newblock
 \APACrefbtitle {{German Conference on Pattern Recognition (GCPR)}.} {{German
  Conference on Pattern Recognition (GCPR)}.}
\newblock
\begin{APACrefDOI}\doi{https://doi.org/10.1007/978-3-319-45886-1_5}\end{APACrefDOI}
\PrintBackRefs{\CurrentBib}

\bibitem [\protect \citeauthoryear {%
Gong%
, Wang%
, Guo%
\BCBL {}\ \BBA {} Lazebnik%
}{%
Gong%
\ \protect \BOthers {.}}{%
{\protect \APACyear {2014}}%
}]{%
gong2014multi}
\APACinsertmetastar {%
gong2014multi}%
\begin{APACrefauthors}%
Gong, Y.%
, Wang, L.%
, Guo, R.%
\BCBL {} Lazebnik, S.%
\end{APACrefauthors}%
\unskip\
\newblock
\APACrefYearMonthDay{2014}{}{}.
\newblock
{\BBOQ}\APACrefatitle {Multi-scale {Orderless} {Pooling} of {Deep}
  {Convolutional} {Activation} {Features}} {Multi-scale {Orderless} {Pooling}
  of {Deep} {Convolutional} {Activation} {Features}}.{\BBCQ}
\newblock
 \APACrefbtitle {European {Conference} on {Computer} {Vision} ({ECCV}).}
  {European {Conference} on {Computer} {Vision} ({ECCV}).}
\newblock
\begin{APACrefDOI}\doi{https://doi.org/10.1007/978-3-319-10584-0_26}\end{APACrefDOI}
\PrintBackRefs{\CurrentBib}

\bibitem [\protect \citeauthoryear {%
Harris%
\ \BBA {} Stephens%
}{%
Harris%
\ \BBA {} Stephens%
}{%
{\protect \APACyear {1988}}%
}]{%
Harris1988ACC}
\APACinsertmetastar {%
Harris1988ACC}%
\begin{APACrefauthors}%
Harris, C.G.%
\BCBT {}\ \BBA {} Stephens, M.J.%
\end{APACrefauthors}%
\unskip\
\newblock
\APACrefYearMonthDay{1988}{}{}.
\newblock
{\BBOQ}\APACrefatitle {{A Combined Corner and Edge Detector}} {{A Combined
  Corner and Edge Detector}}.{\BBCQ}
\newblock
 \APACrefbtitle {{Alvey Vision Conference}.} {{Alvey Vision Conference}.}
\newblock
\begin{APACrefDOI}\doi{https://doi.org/10.5244/c.2.23}\end{APACrefDOI}
\PrintBackRefs{\CurrentBib}

\bibitem [\protect \citeauthoryear {%
Hartley%
\ \BBA {} Zisserman%
}{%
Hartley%
\ \BBA {} Zisserman%
}{%
{\protect \APACyear {2004}}%
}]{%
DLT}
\APACinsertmetastar {%
DLT}%
\begin{APACrefauthors}%
Hartley, R.%
\BCBT {}\ \BBA {} Zisserman, A.%
\end{APACrefauthors}%
\unskip\
\newblock
\APACrefYear{2004}.
\newblock
\APACrefbtitle {{Multiple View Geometry in Computer Vision}} {{Multiple View
  Geometry in Computer Vision}}.
\newblock
\APACaddressPublisher{}{Cambridge University Press}.
\newblock
\begin{APACrefDOI}\doi{https://doi.org/10.1017/CBO9780511811685}\end{APACrefDOI}
\PrintBackRefs{\CurrentBib}

\bibitem [\protect \citeauthoryear {%
Hartwig%
}{%
Hartwig%
}{%
{\protect \APACyear {2005}}%
}]{%
hartwig2005individual}
\APACinsertmetastar {%
hartwig2005individual}%
\begin{APACrefauthors}%
Hartwig, S.%
\end{APACrefauthors}%
\unskip\
\newblock
\APACrefYearMonthDay{2005}{}{}.
\newblock
{\BBOQ}\APACrefatitle {{Individual acoustic identification as a non-invasive
  conservation tool: An approach to the conservation of the African wild dog
  Lycaon pictus (Temminck, 1820)}} {{Individual acoustic identification as a
  non-invasive conservation tool: An approach to the conservation of the
  African wild dog Lycaon pictus (Temminck, 1820)}}.{\BBCQ}
\newblock
\APACjournalVolNumPages{Bioacoustics The International Journal of Animal Sound
  and its Recording}{15}{}{35--50}.
\newblock
\begin{APACrefDOI}\doi{https://doi.org/10.1080/09524622.2005.9753537}\end{APACrefDOI}
\PrintBackRefs{\CurrentBib}

\bibitem [\protect \citeauthoryear {%
He%
, Gkioxari%
, Doll{\'a}r%
\BCBL {}\ \BBA {} Girshick%
}{%
He%
\ \protect \BOthers {.}}{%
{\protect \APACyear {2017}}%
}]{%
he2017mask}
\APACinsertmetastar {%
he2017mask}%
\begin{APACrefauthors}%
He, K.%
, Gkioxari, G.%
, Doll{\'a}r, P.%
\BCBL {} Girshick, R.%
\end{APACrefauthors}%
\unskip\
\newblock
\APACrefYearMonthDay{2017}{}{}.
\newblock
{\BBOQ}\APACrefatitle {Mask {R-CNN}} {Mask {R-CNN}}.{\BBCQ}
\newblock
 \APACrefbtitle {{International} {Conference} on {Computer} {Vision} ({ICCV}).}
  {{International} {Conference} on {Computer} {Vision} ({ICCV}).}
\newblock
\begin{APACrefDOI}\doi{https://doi.org/10.1109/iccv.2017.322}\end{APACrefDOI}
\PrintBackRefs{\CurrentBib}

\bibitem [\protect \citeauthoryear {%
Holmberg%
, Norman%
\BCBL {}\ \BBA {} Arzoumanian%
}{%
Holmberg%
\ \protect \BOthers {.}}{%
{\protect \APACyear {2009}}%
}]{%
holmberg2009estimating}
\APACinsertmetastar {%
holmberg2009estimating}%
\begin{APACrefauthors}%
Holmberg, J.%
, Norman, B.%
\BCBL {} Arzoumanian, Z.%
\end{APACrefauthors}%
\unskip\
\newblock
\APACrefYearMonthDay{2009}{}{}.
\newblock
{\BBOQ}\APACrefatitle {{Estimating population size, structure, and residency
  time for whale sharks Rhincodon typus through collaborative
  photo-identification}} {{Estimating population size, structure, and residency
  time for whale sharks Rhincodon typus through collaborative
  photo-identification}}.{\BBCQ}
\newblock
\APACjournalVolNumPages{Endangered Species Research}{7}{}{39--53}.
\newblock
\begin{APACrefDOI}\doi{https://doi.org/10.3354/esr00186}\end{APACrefDOI}
\PrintBackRefs{\CurrentBib}

\bibitem [\protect \citeauthoryear {%
Hutchison%
\ \protect \BOthers {.}}{%
Hutchison%
\ \protect \BOthers {.}}{%
{\protect \APACyear {2010}}%
}]{%
perronnin2010improving}
\APACinsertmetastar {%
perronnin2010improving}%
\begin{APACrefauthors}%
Hutchison, D.%
, Kanade, T.%
, Kittler, J.%
, Kleinberg, J.M.%
, Mattern, F.%
, Mitchell, J.C.%
\BDBL {}Mensink, T.%
\end{APACrefauthors}%
\unskip\
\newblock
\APACrefYearMonthDay{2010}{}{}.
\newblock
{\BBOQ}\APACrefatitle {Improving the {Fisher} {Kernel} for {Large}-{Scale}
  {Image} {Classification}} {Improving the {Fisher} {Kernel} for
  {Large}-{Scale} {Image} {Classification}}.{\BBCQ}
\newblock
 \APACrefbtitle {European {Conference} on {Computer} {Vision} ({ECCV}).}
  {European {Conference} on {Computer} {Vision} ({ECCV}).}
\newblock
\begin{APACrefDOI}\doi{https://doi.org/10.1007/978-3-642-15561-1_11}\end{APACrefDOI}
\PrintBackRefs{\CurrentBib}

\bibitem [\protect \citeauthoryear {%
Jaakkola%
\ \BBA {} Haussler%
}{%
Jaakkola%
\ \BBA {} Haussler%
}{%
{\protect \APACyear {1999}}%
}]{%
jaakkola1998exploiting}
\APACinsertmetastar {%
jaakkola1998exploiting}%
\begin{APACrefauthors}%
Jaakkola, T.%
\BCBT {}\ \BBA {} Haussler, D.%
\end{APACrefauthors}%
\unskip\
\newblock
\APACrefYearMonthDay{1999}{}{}.
\newblock
{\BBOQ}\APACrefatitle {{Exploiting Generative Models in Discriminative
  Classifiers}} {{Exploiting Generative Models in Discriminative
  Classifiers}}.{\BBCQ}
\newblock
 \APACrefbtitle {{Conference} on {Neural Information Processing Systems}
  ({NeurIPS}).} {{Conference} on {Neural Information Processing Systems}
  ({NeurIPS}).}
\PrintBackRefs{\CurrentBib}

\bibitem [\protect \citeauthoryear {%
J\'egou%
, Douze%
, Schmid%
\BCBL {}\ \BBA {} P\'erez%
}{%
J\'egou%
\ \protect \BOthers {.}}{%
{\protect \APACyear {2010}}%
}]{%
jegou2010aggregating}
\APACinsertmetastar {%
jegou2010aggregating}%
\begin{APACrefauthors}%
J\'egou, H.%
, Douze, M.%
, Schmid, C.%
\BCBL {} P\'erez, P.%
\end{APACrefauthors}%
\unskip\
\newblock
\APACrefYearMonthDay{2010}{}{}.
\newblock
{\BBOQ}\APACrefatitle {Aggregating local descriptors into a compact image
  representation} {Aggregating local descriptors into a compact image
  representation}.{\BBCQ}
\newblock
 \APACrefbtitle {{Conference} on {Computer} {Vision} and {Pattern}
  {Recognition} ({CVPR}).} {{Conference} on {Computer} {Vision} and {Pattern}
  {Recognition} ({CVPR}).}
\newblock
\begin{APACrefDOI}\doi{https://doi.org/10.1109/CVPR.2010.5540039}\end{APACrefDOI}
\PrintBackRefs{\CurrentBib}

\bibitem [\protect \citeauthoryear {%
Koivuniemi%
, Auttila%
, Niemi%
, Lev{\"a}nen%
\BCBL {}\ \BBA {} Kunnasranta%
}{%
Koivuniemi%
\ \protect \BOthers {.}}{%
{\protect \APACyear {2016}}%
}]{%
koivuniemi2016photo}
\APACinsertmetastar {%
koivuniemi2016photo}%
\begin{APACrefauthors}%
Koivuniemi, M.%
, Auttila, M.%
, Niemi, M.%
, Lev{\"a}nen, R.%
\BCBL {} Kunnasranta, M.%
\end{APACrefauthors}%
\unskip\
\newblock
\APACrefYearMonthDay{2016}{}{}.
\newblock
{\BBOQ}\APACrefatitle {{Photo-ID as a tool for studying and monitoring the
  endangered {Saimaa} ringed seal}} {{Photo-ID as a tool for studying and
  monitoring the endangered {Saimaa} ringed seal}}.{\BBCQ}
\newblock
\APACjournalVolNumPages{Endangered Species Research}{30}{}{29--36}.
\newblock
\begin{APACrefDOI}\doi{https://doi.org/10.3354/esr00723}\end{APACrefDOI}
\PrintBackRefs{\CurrentBib}

\bibitem [\protect \citeauthoryear {%
Koivuniemi%
, Kurkilahti%
, Niemi%
, Auttila%
\BCBL {}\ \BBA {} Kunnasranta%
}{%
Koivuniemi%
\ \protect \BOthers {.}}{%
{\protect \APACyear {2019}}%
}]{%
koivuniemi2019mark}
\APACinsertmetastar {%
koivuniemi2019mark}%
\begin{APACrefauthors}%
Koivuniemi, M.%
, Kurkilahti, M.%
, Niemi, M.%
, Auttila, M.%
\BCBL {} Kunnasranta, M.%
\end{APACrefauthors}%
\unskip\
\newblock
\APACrefYearMonthDay{2019}{}{}.
\newblock
{\BBOQ}\APACrefatitle {{A mark--recapture approach for estimating population
  size of the endangered ringed seal (Phoca hispida saimensis)}} {{A
  mark--recapture approach for estimating population size of the endangered
  ringed seal (Phoca hispida saimensis)}}.{\BBCQ}
\newblock
\APACjournalVolNumPages{PLOS ONE}{14}{}{214--269}.
\newblock
\begin{APACrefDOI}\doi{https://doi.org/10.1371/journal.pone.0214269}\end{APACrefDOI}
\PrintBackRefs{\CurrentBib}

\bibitem [\protect \citeauthoryear {%
Korschens%
\ \BBA {} Denzler%
}{%
Korschens%
\ \BBA {} Denzler%
}{%
{\protect \APACyear {2019}}%
}]{%
korschens2019elpephants}
\APACinsertmetastar {%
korschens2019elpephants}%
\begin{APACrefauthors}%
Korschens, M.%
\BCBT {}\ \BBA {} Denzler, J.%
\end{APACrefauthors}%
\unskip\
\newblock
\APACrefYearMonthDay{2019}{}{}.
\newblock
{\BBOQ}\APACrefatitle {{ELPephants: A Fine-Grained Dataset for Elephant
  Re-Identification}} {{ELPephants: A Fine-Grained Dataset for Elephant
  Re-Identification}}.{\BBCQ}
\newblock
 \APACrefbtitle {{International} {Conference} on {Computer} {Vision} {Workshop}
  ({ICCVW}).} {{International} {Conference} on {Computer} {Vision} {Workshop}
  ({ICCVW}).}
\newblock
\begin{APACrefDOI}\doi{https://doi.org/10.1109/iccvw.2019.00035}\end{APACrefDOI}
\PrintBackRefs{\CurrentBib}

\bibitem [\protect \citeauthoryear {%
Kumar%
\ \protect \BOthers {.}}{%
Kumar%
\ \protect \BOthers {.}}{%
{\protect \APACyear {2018}}%
}]{%
kumar2018deep}
\APACinsertmetastar {%
kumar2018deep}%
\begin{APACrefauthors}%
Kumar, S.%
, Pandey, A.%
, Sai Ram~Satwik, K.%
, Kumar, S.%
, Singh, S.K.%
, Singh, A.K.%
\BCBL {} Mohan, A.%
\end{APACrefauthors}%
\unskip\
\newblock
\APACrefYearMonthDay{2018}{}{}.
\newblock
{\BBOQ}\APACrefatitle {Deep learning framework for recognition of cattle using
  muzzle point image pattern} {Deep learning framework for recognition of
  cattle using muzzle point image pattern}.{\BBCQ}
\newblock
\APACjournalVolNumPages{Measurement}{116}{}{1--17}.
\newblock
\begin{APACrefDOI}\doi{https://doi.org/10.1016/j.measurement.2017.10.064}\end{APACrefDOI}
\PrintBackRefs{\CurrentBib}

\bibitem [\protect \citeauthoryear {%
Kunnasranta%
\ \protect \BOthers {.}}{%
Kunnasranta%
\ \protect \BOthers {.}}{%
{\protect \APACyear {2021}}%
}]{%
kunnasranta2021sealed}
\APACinsertmetastar {%
kunnasranta2021sealed}%
\begin{APACrefauthors}%
Kunnasranta, M.%
, Niemi, M.%
, Auttila, M.%
, Valtonen, M.%
, Kammonen, J.%
\BCBL {} Nyman, T.%
\end{APACrefauthors}%
\unskip\
\newblock
\APACrefYearMonthDay{2021}{}{}.
\newblock
{\BBOQ}\APACrefatitle {Sealed in a lake -- {Biology} and conservation of the
  endangered {Saimaa} ringed seal: {A} review} {Sealed in a lake -- {Biology}
  and conservation of the endangered {Saimaa} ringed seal: {A} review}.{\BBCQ}
\newblock
\APACjournalVolNumPages{Biological Conservation}{253}{}{108908}.
\newblock
\begin{APACrefDOI}\doi{https://doi.org/10.1016/j.biocon.2020.108908}\end{APACrefDOI}
\PrintBackRefs{\CurrentBib}

\bibitem [\protect \citeauthoryear {%
Li%
, Li%
, Tang%
, Qian%
\BCBL {}\ \BBA {} Lin%
}{%
Li%
\ \protect \BOthers {.}}{%
{\protect \APACyear {2020}}%
}]{%
li2019amur}
\APACinsertmetastar {%
li2019amur}%
\begin{APACrefauthors}%
Li, S.%
, Li, J.%
, Tang, H.%
, Qian, R.%
\BCBL {} Lin, W.%
\end{APACrefauthors}%
\unskip\
\newblock
\APACrefYearMonthDay{2020}{}{}.
\newblock
{\BBOQ}\APACrefatitle {{ATRW}: {A} {Benchmark} for {Amur} {Tiger}
  {Re}-identification in the {Wild}} {{ATRW}: {A} {Benchmark} for {Amur}
  {Tiger} {Re}-identification in the {Wild}}.{\BBCQ}
\newblock
 \APACrefbtitle {{ACM} {International} {Conference} on {Multimedia}.} {{ACM}
  {International} {Conference} on {Multimedia}.}
\newblock
\begin{APACrefDOI}\doi{https://doi.org/10.1145/3394171.3413569}\end{APACrefDOI}
\PrintBackRefs{\CurrentBib}

\bibitem [\protect \citeauthoryear {%
Lindeberg%
}{%
Lindeberg%
}{%
{\protect \APACyear {1998}}%
}]{%
lindeberg1998feature}
\APACinsertmetastar {%
lindeberg1998feature}%
\begin{APACrefauthors}%
Lindeberg, T.%
\end{APACrefauthors}%
\unskip\
\newblock
\APACrefYearMonthDay{1998}{}{}.
\newblock
{\BBOQ}\APACrefatitle {{Feature Detection with Automatic Scale Selection}}
  {{Feature Detection with Automatic Scale Selection}}.{\BBCQ}
\newblock
\APACjournalVolNumPages{International Journal of Computer
  Vision}{30}{}{77--116}.
\newblock
\begin{APACrefDOI}\doi{https://doi.org/10.1023/A:1008045108935}\end{APACrefDOI}
\PrintBackRefs{\CurrentBib}

\bibitem [\protect \citeauthoryear {%
C.~Liu%
, Zhang%
\BCBL {}\ \BBA {} Guo%
}{%
C.~Liu%
\ \protect \BOthers {.}}{%
{\protect \APACyear {2019}}%
}]{%
Liu_2019_ICCV_Workshops}
\APACinsertmetastar {%
Liu_2019_ICCV_Workshops}%
\begin{APACrefauthors}%
Liu, C.%
, Zhang, R.%
\BCBL {} Guo, L.%
\end{APACrefauthors}%
\unskip\
\newblock
\APACrefYearMonthDay{2019}{}{}.
\newblock
{\BBOQ}\APACrefatitle {Part-{Pose} {Guided} {Amur} {Tiger}
  {Re}-{Identification}} {Part-{Pose} {Guided} {Amur} {Tiger}
  {Re}-{Identification}}.{\BBCQ}
\newblock
 \APACrefbtitle {{International} {Conference} on {Computer} {Vision} {Workshop}
  ({ICCVW}).} {{International} {Conference} on {Computer} {Vision} {Workshop}
  ({ICCVW}).}
\newblock
\begin{APACrefDOI}\doi{https://doi.org/10.1109/ICCVW.2019.00042}\end{APACrefDOI}
\PrintBackRefs{\CurrentBib}

\bibitem [\protect \citeauthoryear {%
N.~Liu%
, Zhao%
, Zhang%
, Cheng%
\BCBL {}\ \BBA {} Zhu%
}{%
N.~Liu%
\ \protect \BOthers {.}}{%
{\protect \APACyear {2019}}%
}]{%
Liu_2019_ICCV}
\APACinsertmetastar {%
Liu_2019_ICCV}%
\begin{APACrefauthors}%
Liu, N.%
, Zhao, Q.%
, Zhang, N.%
, Cheng, X.%
\BCBL {} Zhu, J.%
\end{APACrefauthors}%
\unskip\
\newblock
\APACrefYearMonthDay{2019}{}{}.
\newblock
{\BBOQ}\APACrefatitle {Pose-{Guided} {Complementary} {Features} {Learning} for
  {Amur} {Tiger} {Re}-{Identification}} {Pose-{Guided} {Complementary}
  {Features} {Learning} for {Amur} {Tiger} {Re}-{Identification}}.{\BBCQ}
\newblock
 \APACrefbtitle {{International} {Conference} on {Computer} {Vision} {Workshop}
  ({ICCVW}).} {{International} {Conference} on {Computer} {Vision} {Workshop}
  ({ICCVW}).}
\newblock
\begin{APACrefDOI}\doi{https://doi.org/10.1109/ICCVW.2019.00038}\end{APACrefDOI}
\PrintBackRefs{\CurrentBib}

\bibitem [\protect \citeauthoryear {%
D.~Lowe%
}{%
D.~Lowe%
}{%
{\protect \APACyear {2004}}%
}]{%
lowe2004distinctive}
\APACinsertmetastar {%
lowe2004distinctive}%
\begin{APACrefauthors}%
Lowe, D.%
\end{APACrefauthors}%
\unskip\
\newblock
\APACrefYearMonthDay{2004}{}{}.
\newblock
{\BBOQ}\APACrefatitle {{Distinctive Image Features from Scale-Invariant
  Keypoints}} {{Distinctive Image Features from Scale-Invariant
  Keypoints}}.{\BBCQ}
\newblock
\APACjournalVolNumPages{International Journal of Computer
  Vision}{60}{}{91--110}.
\newblock
\begin{APACrefDOI}\doi{https://doi.org/10.1023/B:VISI.0000029664.99615.94}\end{APACrefDOI}
\PrintBackRefs{\CurrentBib}

\bibitem [\protect \citeauthoryear {%
D.G.~Lowe%
}{%
D.G.~Lowe%
}{%
{\protect \APACyear {1999}}%
}]{%
lowe1999object}
\APACinsertmetastar {%
lowe1999object}%
\begin{APACrefauthors}%
Lowe, D.G.%
\end{APACrefauthors}%
\unskip\
\newblock
\APACrefYearMonthDay{1999}{}{}.
\newblock
{\BBOQ}\APACrefatitle {{Object Recognition from Local Scale-Invariant
  Features}} {{Object Recognition from Local Scale-Invariant Features}}.{\BBCQ}
\newblock
 \APACrefbtitle {{International} {Conference} on {Computer} {Vision} ({ICCV}).}
  {{International} {Conference} on {Computer} {Vision} ({ICCV}).}
\newblock
\begin{APACrefDOI}\doi{https://doi.org/10.5555/850924.851523}\end{APACrefDOI}
\PrintBackRefs{\CurrentBib}

\bibitem [\protect \citeauthoryear {%
MacQueen%
\ \protect \BOthers {.}}{%
MacQueen%
\ \protect \BOthers {.}}{%
{\protect \APACyear {1967}}%
}]{%
macqueen1967some}
\APACinsertmetastar {%
macqueen1967some}%
\begin{APACrefauthors}%
MacQueen, J.%
\BCBT {}\ \BOthersPeriod {.}
\end{APACrefauthors}%
\unskip\
\newblock
\APACrefYearMonthDay{1967}{}{}.
\newblock
{\BBOQ}\APACrefatitle {Some methods for classification and analysis of
  multivariate observations} {Some methods for classification and analysis of
  multivariate observations}.{\BBCQ}
\newblock
 \APACrefbtitle {{Berkeley Symposium on Mathematical Statistics and
  Probability}.} {{Berkeley Symposium on Mathematical Statistics and
  Probability}.}
\PrintBackRefs{\CurrentBib}

\bibitem [\protect \citeauthoryear {%
Malik%
, Kiranyaz%
\BCBL {}\ \BBA {} Gabbouj%
}{%
Malik%
\ \protect \BOthers {.}}{%
{\protect \APACyear {2021}}%
}]{%
malik2021self}
\APACinsertmetastar {%
malik2021self}%
\begin{APACrefauthors}%
Malik, J.%
, Kiranyaz, S.%
\BCBL {} Gabbouj, M.%
\end{APACrefauthors}%
\unskip\
\newblock
\APACrefYearMonthDay{2021}{}{}.
\newblock
{\BBOQ}\APACrefatitle {Self-organized operational neural networks for severe
  image restoration problems} {Self-organized operational neural networks for
  severe image restoration problems}.{\BBCQ}
\newblock
\APACjournalVolNumPages{Neural Networks}{135}{}{201--211}.
\newblock
\begin{APACrefDOI}\doi{https://doi.org/10.1016/j.neunet.2020.12.014}\end{APACrefDOI}
\PrintBackRefs{\CurrentBib}

\bibitem [\protect \citeauthoryear {%
Mantiuk%
, Myszkowski%
\BCBL {}\ \BBA {} Seidel%
}{%
Mantiuk%
\ \protect \BOthers {.}}{%
{\protect \APACyear {2006}}%
}]{%
mantiuk_perceptual_2006}
\APACinsertmetastar {%
mantiuk_perceptual_2006}%
\begin{APACrefauthors}%
Mantiuk, R.%
, Myszkowski, K.%
\BCBL {} Seidel, H\BHBI P.%
\end{APACrefauthors}%
\unskip\
\newblock
\APACrefYearMonthDay{2006}{}{}.
\newblock
{\BBOQ}\APACrefatitle {A perceptual framework for contrast processing of high
  dynamic range images} {A perceptual framework for contrast processing of high
  dynamic range images}.{\BBCQ}
\newblock
\APACjournalVolNumPages{ACM Transactions on Applied Perception}{3}{}{286--308}.
\newblock
\begin{APACrefDOI}\doi{https://doi.org/10.1145/1166087.1166095}\end{APACrefDOI}
\PrintBackRefs{\CurrentBib}

\bibitem [\protect \citeauthoryear {%
McCoy%
\ \protect \BOthers {.}}{%
McCoy%
\ \protect \BOthers {.}}{%
{\protect \APACyear {2018}}%
}]{%
mccoy2018long}
\APACinsertmetastar {%
mccoy2018long}%
\begin{APACrefauthors}%
McCoy, E.%
, Burce, R.%
, David, D.%
, Aca, E.%
, Hardy, J.%
, Labaja, J.%
\BDBL {}Araujo, G.%
\end{APACrefauthors}%
\unskip\
\newblock
\APACrefYearMonthDay{2018}{}{}.
\newblock
{\BBOQ}\APACrefatitle {{Long-Term Photo-Identification Reveals the Population
  Dynamics and Strong Site Fidelity of Adult Whale Sharks to the Coastal Waters
  of Donsol, Philippines}} {{Long-Term Photo-Identification Reveals the
  Population Dynamics and Strong Site Fidelity of Adult Whale Sharks to the
  Coastal Waters of Donsol, Philippines}}.{\BBCQ}
\newblock
\APACjournalVolNumPages{Frontiers in Marine Science}{5}{}{271}.
\newblock
\begin{APACrefDOI}\doi{https://doi.org/10.3389/fmars.2018.00271}\end{APACrefDOI}
\PrintBackRefs{\CurrentBib}

\bibitem [\protect \citeauthoryear {%
McLachlan%
\ \BBA {} Basford%
}{%
McLachlan%
\ \BBA {} Basford%
}{%
{\protect \APACyear {1988}}%
}]{%
mclachlan1988mixture}
\APACinsertmetastar {%
mclachlan1988mixture}%
\begin{APACrefauthors}%
McLachlan, G.J.%
\BCBT {}\ \BBA {} Basford, K.E.%
\end{APACrefauthors}%
\unskip\
\newblock
\APACrefYear{1988}.
\newblock
\APACrefbtitle {Mixture models: Inference and applications to clustering}
  {Mixture models: Inference and applications to clustering}.
\newblock
\APACaddressPublisher{}{M. Dekker New York}.
\PrintBackRefs{\CurrentBib}

\bibitem [\protect \citeauthoryear {%
Mikolajczyk%
\ \BBA {} Schmid%
}{%
Mikolajczyk%
\ \BBA {} Schmid%
}{%
{\protect \APACyear {2002}}%
}]{%
mikolajczyk2002affine}
\APACinsertmetastar {%
mikolajczyk2002affine}%
\begin{APACrefauthors}%
Mikolajczyk, K.%
\BCBT {}\ \BBA {} Schmid, C.%
\end{APACrefauthors}%
\unskip\
\newblock
\APACrefYearMonthDay{2002}{}{}.
\newblock
{\BBOQ}\APACrefatitle {An affine invariant interest point detector} {An affine
  invariant interest point detector}.{\BBCQ}
\newblock
 \APACrefbtitle {European {Conference} on {Computer} {Vision} ({ECCV}).}
  {European {Conference} on {Computer} {Vision} ({ECCV}).}
\newblock
\begin{APACrefDOI}\doi{https://doi.org/10.1007/3-540-47969-4_9}\end{APACrefDOI}
\PrintBackRefs{\CurrentBib}

\bibitem [\protect \citeauthoryear {%
Mikolajczyk%
\ \BBA {} Schmid%
}{%
Mikolajczyk%
\ \BBA {} Schmid%
}{%
{\protect \APACyear {2004}}%
}]{%
mikolajczyk2004scale}
\APACinsertmetastar {%
mikolajczyk2004scale}%
\begin{APACrefauthors}%
Mikolajczyk, K.%
\BCBT {}\ \BBA {} Schmid, C.%
\end{APACrefauthors}%
\unskip\
\newblock
\APACrefYearMonthDay{2004}{}{}.
\newblock
{\BBOQ}\APACrefatitle {{Scale} {\&} {Affine Invariant Interest Point
  Detectors}} {{Scale} {\&} {Affine Invariant Interest Point
  Detectors}}.{\BBCQ}
\newblock
\APACjournalVolNumPages{International Journal of Computer
  Vision}{60}{}{63--86}.
\newblock
\begin{APACrefDOI}\doi{https://doi.org/10.1023/B:VISI.0000027790.02288.f2}\end{APACrefDOI}
\PrintBackRefs{\CurrentBib}

\bibitem [\protect \citeauthoryear {%
Mishchuk%
, Mishkin%
, Radenovic%
\BCBL {}\ \BBA {} Matas%
}{%
Mishchuk%
\ \protect \BOthers {.}}{%
{\protect \APACyear {2017}}%
}]{%
HardNet2017}
\APACinsertmetastar {%
HardNet2017}%
\begin{APACrefauthors}%
Mishchuk, A.%
, Mishkin, D.%
, Radenovic, F.%
\BCBL {} Matas, J.%
\end{APACrefauthors}%
\unskip\
\newblock
\APACrefYearMonthDay{2017}{}{}.
\newblock
{\BBOQ}\APACrefatitle {Working hard to know your neighbor\textquotesingle s
  margins: Local descriptor learning loss} {Working hard to know your
  neighbor\textquotesingle s margins: Local descriptor learning loss}.{\BBCQ}
\newblock
 \APACrefbtitle {{Conference on Neural Information Processing Systems
  (NeurIPS)}.} {{Conference on Neural Information Processing Systems
  (NeurIPS)}.}
\PrintBackRefs{\CurrentBib}

\bibitem [\protect \citeauthoryear {%
Mishkin%
, Radenovi\'c%
\BCBL {}\ \BBA {} Matas%
}{%
Mishkin%
\ \protect \BOthers {.}}{%
{\protect \APACyear {2018}}%
}]{%
AffNet2018}
\APACinsertmetastar {%
AffNet2018}%
\begin{APACrefauthors}%
Mishkin, D.%
, Radenovi\'c, F.%
\BCBL {} Matas, J.%
\end{APACrefauthors}%
\unskip\
\newblock
\APACrefYearMonthDay{2018}{}{}.
\newblock
{\BBOQ}\APACrefatitle {Repeatability {Is} {Not} {Enough}: {Learning} {Affine}
  {Regions} via {Discriminability}} {Repeatability {Is} {Not} {Enough}:
  {Learning} {Affine} {Regions} via {Discriminability}}.{\BBCQ}
\newblock
 \APACrefbtitle {European {Conference} on {Computer} {Vision} ({ECCV}).}
  {European {Conference} on {Computer} {Vision} ({ECCV}).}
\newblock
\begin{APACrefDOI}\doi{https://doi.org/10.1007/978-3-030-01240-3_18}\end{APACrefDOI}
\PrintBackRefs{\CurrentBib}

\bibitem [\protect \citeauthoryear {%
Moskvyak%
, Maire%
, Dayoub%
, Armstrong%
\BCBL {}\ \BBA {} Baktashmotlagh%
}{%
Moskvyak%
, Maire%
, Dayoub%
, Armstrong%
\BCBL {}\ \BBA {} Baktashmotlagh%
}{%
{\protect \APACyear {2021}}%
}]{%
moskvyak2019robust}
\APACinsertmetastar {%
moskvyak2019robust}%
\begin{APACrefauthors}%
Moskvyak, O.%
, Maire, F.%
, Dayoub, F.%
, Armstrong, A.O.%
\BCBL {} Baktashmotlagh, M.%
\end{APACrefauthors}%
\unskip\
\newblock
\APACrefYearMonthDay{2021}{}{}.
\newblock
{\BBOQ}\APACrefatitle {Robust Re-identification of Manta Rays from Natural
  Markings by Learning Pose Invariant Embeddings} {Robust re-identification of
  manta rays from natural markings by learning pose invariant
  embeddings}.{\BBCQ}
\newblock
 \APACrefbtitle {{International Conference on Digital Image Computing:
  Techniques and Applications (DICTA)}.} {{International Conference on Digital
  Image Computing: Techniques and Applications (DICTA)}.}
\newblock
\begin{APACrefDOI}\doi{https://doi.org/10.1109/DICTA52665.2021.9647359}\end{APACrefDOI}
\PrintBackRefs{\CurrentBib}

\bibitem [\protect \citeauthoryear {%
Moskvyak%
, Maire%
, Dayoub%
\BCBL {}\ \BBA {} Baktashmotlagh%
}{%
Moskvyak%
, Maire%
, Dayoub%
\BCBL {}\ \BBA {} Baktashmotlagh%
}{%
{\protect \APACyear {2021}}%
}]{%
moskvyak2021keypoint}
\APACinsertmetastar {%
moskvyak2021keypoint}%
\begin{APACrefauthors}%
Moskvyak, O.%
, Maire, F.%
, Dayoub, F.%
\BCBL {} Baktashmotlagh, M.%
\end{APACrefauthors}%
\unskip\
\newblock
\APACrefYearMonthDay{2021}{}{}.
\newblock
{\BBOQ}\APACrefatitle {Keypoint-{Aligned} {Embeddings} for {Image} {Retrieval}
  and {Re}-identification} {Keypoint-{Aligned} {Embeddings} for {Image}
  {Retrieval} and {Re}-identification}.{\BBCQ}
\newblock
 \APACrefbtitle {{Winter Conference on Applications of Computer Vision
  (WACV)}.} {{Winter Conference on Applications of Computer Vision (WACV)}.}
\newblock
\begin{APACrefDOI}\doi{https://doi.org/10.1109/48630.2021.00072}\end{APACrefDOI}
\PrintBackRefs{\CurrentBib}

\bibitem [\protect \citeauthoryear {%
Nepovinnykh%
, Chelak%
\BCBL {}\ \protect \BOthers {.}}{%
Nepovinnykh%
, Chelak%
\BCBL {}\ \protect \BOthers {.}}{%
{\protect \APACyear {2022}}%
}]{%
ladoga}
\APACinsertmetastar {%
ladoga}%
\begin{APACrefauthors}%
Nepovinnykh, E.%
, Chelak, I.%
, Lushpanov, A.%
, Eerola, T.%
, K\"alvi\"ainen, H.%
\BCBL {} Chirkova, O.%
\end{APACrefauthors}%
\unskip\
\newblock
\APACrefYearMonthDay{2022}{}{}.
\newblock
{\BBOQ}\APACrefatitle {{Matching individual Ladoga ringed seals across
  short-term image sequences}} {{Matching individual Ladoga ringed seals across
  short-term image sequences}}.{\BBCQ}
\newblock
\APACjournalVolNumPages{{Mammalian Biology}}{}{}{1-16}.
\newblock
\begin{APACrefDOI}\doi{https://doi.org/10.1007/s42991-022-00229-3}\end{APACrefDOI}
\PrintBackRefs{\CurrentBib}

\bibitem [\protect \citeauthoryear {%
Nepovinnykh%
, Eerola%
\BCBL {}\ \protect \BOthers {.}}{%
Nepovinnykh%
, Eerola%
\BCBL {}\ \protect \BOthers {.}}{%
{\protect \APACyear {2022}}%
}]{%
dataset}
\APACinsertmetastar {%
dataset}%
\begin{APACrefauthors}%
Nepovinnykh, E.%
, Eerola, T.%
, Biard, V.%
, Mutka, P.%
, Niemi, M.%
, Kunnasranta, M.%
\BCBL {} K\"{a}lvi\"{a}inen, H.%
\end{APACrefauthors}%
\unskip\
\newblock
\APACrefYearMonthDay{2022}{}{}.
\newblock
{\BBOQ}\APACrefatitle {{SealID: Saimaa ringed seal re-identification database}}
  {{SealID: Saimaa ringed seal re-identification database}}.{\BBCQ}
\newblock
\APACjournalVolNumPages{arXiv preprint arXiv:2206.02260}{}{}{}.
\PrintBackRefs{\CurrentBib}

\bibitem [\protect \citeauthoryear {%
Nepovinnykh%
, Eerola%
\BCBL {}\ \BBA {} K\"alvi\"ainen%
}{%
Nepovinnykh%
\ \protect \BOthers {.}}{%
{\protect \APACyear {2020}}%
}]{%
nepovinnykh2020siamese}
\APACinsertmetastar {%
nepovinnykh2020siamese}%
\begin{APACrefauthors}%
Nepovinnykh, E.%
, Eerola, T.%
\BCBL {} K\"alvi\"ainen, H.%
\end{APACrefauthors}%
\unskip\
\newblock
\APACrefYearMonthDay{2020}{}{}.
\newblock
{\BBOQ}\APACrefatitle {{Siamese Network Based Pelage Pattern Matching for
  Ringed Seal Re-identification}} {{Siamese Network Based Pelage Pattern
  Matching for Ringed Seal Re-identification}}.{\BBCQ}
\newblock
 \APACrefbtitle {{Winter Conference on Applications of Computer Vision
  Workshops (WACVW)}.} {{Winter Conference on Applications of Computer Vision
  Workshops (WACVW)}.}
\newblock
\begin{APACrefDOI}\doi{https://doi.org/10.1109/wacvw50321.2020.9096935}\end{APACrefDOI}
\PrintBackRefs{\CurrentBib}

\bibitem [\protect \citeauthoryear {%
Nepovinnykh%
, Eerola%
, K{\"a}lvi{\"a}inen%
\BCBL {}\ \BBA {} Radchenko%
}{%
Nepovinnykh%
\ \protect \BOthers {.}}{%
{\protect \APACyear {2018}}%
}]{%
nepovinnykh2018identification}
\APACinsertmetastar {%
nepovinnykh2018identification}%
\begin{APACrefauthors}%
Nepovinnykh, E.%
, Eerola, T.%
, K{\"a}lvi{\"a}inen, H.%
\BCBL {} Radchenko, G.%
\end{APACrefauthors}%
\unskip\
\newblock
\APACrefYearMonthDay{2018}{}{}.
\newblock
{\BBOQ}\APACrefatitle {{Identification of Saimaa Ringed Seal Individuals Using
  Transfer Learning}} {{Identification of Saimaa Ringed Seal Individuals Using
  Transfer Learning}}.{\BBCQ}
\newblock
 \APACrefbtitle {{International Conference on Advanced Concepts for Intelligent
  Vision Systems (ACIVS)}.} {{International Conference on Advanced Concepts for
  Intelligent Vision Systems (ACIVS)}.}
\newblock
\begin{APACrefDOI}\doi{https://doi.org/10.1007/978-3-030-01449-0_18}\end{APACrefDOI}
\PrintBackRefs{\CurrentBib}

\bibitem [\protect \citeauthoryear {%
Nipko%
, Holcombe%
\BCBL {}\ \BBA {} Kelly%
}{%
Nipko%
\ \protect \BOthers {.}}{%
{\protect \APACyear {2020}}%
}]{%
nipko2020identifying}
\APACinsertmetastar {%
nipko2020identifying}%
\begin{APACrefauthors}%
Nipko, R.%
, Holcombe, B.%
\BCBL {} Kelly, M.%
\end{APACrefauthors}%
\unskip\
\newblock
\APACrefYearMonthDay{2020}{}{}.
\newblock
{\BBOQ}\APACrefatitle {{Identifying Individual Jaguars and Ocelots via
  Pattern-Recognition Software: Comparing HotSpotter and Wild-ID}}
  {{Identifying Individual Jaguars and Ocelots via Pattern-Recognition
  Software: Comparing HotSpotter and Wild-ID}}.{\BBCQ}
\newblock
\APACjournalVolNumPages{Wildlife Society Bulletin}{44}{}{424-433}.
\newblock
\begin{APACrefDOI}\doi{https://doi.org/10.1002/wsb.1086}\end{APACrefDOI}
\PrintBackRefs{\CurrentBib}

\bibitem [\protect \citeauthoryear {%
Norouzzadeh%
\ \protect \BOthers {.}}{%
Norouzzadeh%
\ \protect \BOthers {.}}{%
{\protect \APACyear {2018}}%
}]{%
norouzzadeh2018automatically}
\APACinsertmetastar {%
norouzzadeh2018automatically}%
\begin{APACrefauthors}%
Norouzzadeh, M.S.%
, Nguyen, A.%
, Kosmala, M.%
, Swanson, A.%
, Palmer, M.S.%
, Packer, C.%
\BCBL {} Clune, J.%
\end{APACrefauthors}%
\unskip\
\newblock
\APACrefYearMonthDay{2018}{}{}.
\newblock
{\BBOQ}\APACrefatitle {Automatically identifying, counting, and describing wild
  animals in camera-trap images with deep learning} {Automatically identifying,
  counting, and describing wild animals in camera-trap images with deep
  learning}.{\BBCQ}
\newblock
\APACjournalVolNumPages{Proceedings of the National Academy of
  Sciences}{115}{}{5716--5725}.
\newblock
\begin{APACrefDOI}\doi{https://doi.org/10.1073/pnas.1719367115}\end{APACrefDOI}
\PrintBackRefs{\CurrentBib}

\bibitem [\protect \citeauthoryear {%
Parham%
, Crall%
, Stewart%
, Berger-Wolf%
\BCBL {}\ \BBA {} Rubenstein%
}{%
Parham%
\ \protect \BOthers {.}}{%
{\protect \APACyear {2017}}%
}]{%
parham2017animal}
\APACinsertmetastar {%
parham2017animal}%
\begin{APACrefauthors}%
Parham, J.R.%
, Crall, J.%
, Stewart, C.%
, Berger-Wolf, T.%
\BCBL {} Rubenstein, D.%
\end{APACrefauthors}%
\unskip\
\newblock
\APACrefYearMonthDay{2017}{}{}.
\newblock
{\BBOQ}\APACrefatitle {Animal {Population} {Censusing} at {Scale} with
  {Citizen} {Science} and {Photographic} {Identification}} {Animal {Population}
  {Censusing} at {Scale} with {Citizen} {Science} and {Photographic}
  {Identification}}.{\BBCQ}
\newblock
 \APACrefbtitle {{AAAI} {Spring} {Symposium} {Series}.} {{AAAI} {Spring}
  {Symposium} {Series}.}
\PrintBackRefs{\CurrentBib}

\bibitem [\protect \citeauthoryear {%
Perronnin%
\ \BBA {} Dance%
}{%
Perronnin%
\ \BBA {} Dance%
}{%
{\protect \APACyear {2007}}%
}]{%
perronnin2007fisher}
\APACinsertmetastar {%
perronnin2007fisher}%
\begin{APACrefauthors}%
Perronnin, F.%
\BCBT {}\ \BBA {} Dance, C.%
\end{APACrefauthors}%
\unskip\
\newblock
\APACrefYearMonthDay{2007}{}{}.
\newblock
{\BBOQ}\APACrefatitle {{Fisher Kernels on Visual Vocabularies for Image
  Categorization}} {{Fisher Kernels on Visual Vocabularies for Image
  Categorization}}.{\BBCQ}
\newblock
 \APACrefbtitle {{Conference} on {Computer} {Vision} and {Pattern}
  {Recognition} ({CVPR}).} {{Conference} on {Computer} {Vision} and {Pattern}
  {Recognition} ({CVPR}).}
\newblock
\begin{APACrefDOI}\doi{https://doi.org/10.1109/CVPR.2007.383266}\end{APACrefDOI}
\PrintBackRefs{\CurrentBib}

\bibitem [\protect \citeauthoryear {%
Perronnin%
, Liu%
, S\'anchez%
\BCBL {}\ \BBA {} Poirier%
}{%
Perronnin%
\ \protect \BOthers {.}}{%
{\protect \APACyear {2010}}%
}]{%
perronnin2010large}
\APACinsertmetastar {%
perronnin2010large}%
\begin{APACrefauthors}%
Perronnin, F.%
, Liu, Y.%
, S\'anchez, J.%
\BCBL {} Poirier, H.%
\end{APACrefauthors}%
\unskip\
\newblock
\APACrefYearMonthDay{2010}{}{}.
\newblock
{\BBOQ}\APACrefatitle {{Large-scale image retrieval with compressed Fisher
  vectors}} {{Large-scale image retrieval with compressed Fisher
  vectors}}.{\BBCQ}
\newblock
 \APACrefbtitle {{Conference} on {Computer} {Vision} and {Pattern}
  {Recognition} ({CVPR}).} {{Conference} on {Computer} {Vision} and {Pattern}
  {Recognition} ({CVPR}).}
\newblock
\begin{APACrefDOI}\doi{https://doi.org/10.1109/CVPR.2010.5540009}\end{APACrefDOI}
\PrintBackRefs{\CurrentBib}

\bibitem [\protect \citeauthoryear {%
Pruchova%
, Jaška%
\BCBL {}\ \BBA {} Linhart%
}{%
Pruchova%
\ \protect \BOthers {.}}{%
{\protect \APACyear {2017}}%
}]{%
pruchova2017cues}
\APACinsertmetastar {%
pruchova2017cues}%
\begin{APACrefauthors}%
Pruchova, A.%
, Jaška, P.%
\BCBL {} Linhart, P.%
\end{APACrefauthors}%
\unskip\
\newblock
\APACrefYearMonthDay{2017}{}{}.
\newblock
{\BBOQ}\APACrefatitle {{Cues to individual identity in songs of songbirds:
  testing general song characteristics in Chiffchaffs Phylloscopus collybita}}
  {{Cues to individual identity in songs of songbirds: testing general song
  characteristics in Chiffchaffs Phylloscopus collybita}}.{\BBCQ}
\newblock
\APACjournalVolNumPages{Journal of Ornithology}{158}{}{911--924}.
\newblock
\begin{APACrefDOI}\doi{https://doi.org/10.1007/s10336-017-1455-6}\end{APACrefDOI}
\PrintBackRefs{\CurrentBib}

\bibitem [\protect \citeauthoryear {%
Radenovi{\'c}%
, Tolias%
\BCBL {}\ \BBA {} Chum%
}{%
Radenovi{\'c}%
\ \protect \BOthers {.}}{%
{\protect \APACyear {2018}}%
}]{%
radenovic2018fine}
\APACinsertmetastar {%
radenovic2018fine}%
\begin{APACrefauthors}%
Radenovi{\'c}, F.%
, Tolias, G.%
\BCBL {} Chum, O.%
\end{APACrefauthors}%
\unskip\
\newblock
\APACrefYearMonthDay{2018}{}{}.
\newblock
{\BBOQ}\APACrefatitle {{Fine-tuning CNN image retrieval with no human
  annotation}} {{Fine-tuning CNN image retrieval with no human
  annotation}}.{\BBCQ}
\newblock
\APACjournalVolNumPages{IEEE Transactions on Pattern Analysis and Machine
  Intelligence}{41}{}{1655--1668}.
\newblock
\begin{APACrefDOI}\doi{https://doi.org/10.1109/tpami.2018.2846566}\end{APACrefDOI}
\PrintBackRefs{\CurrentBib}

\bibitem [\protect \citeauthoryear {%
Razavian%
, Sullivan%
, Carlsson%
\BCBL {}\ \BBA {} Maki%
}{%
Razavian%
\ \protect \BOthers {.}}{%
{\protect \APACyear {2016}}%
}]{%
razavian2016visual}
\APACinsertmetastar {%
razavian2016visual}%
\begin{APACrefauthors}%
Razavian, A.S.%
, Sullivan, J.%
, Carlsson, S.%
\BCBL {} Maki, A.%
\end{APACrefauthors}%
\unskip\
\newblock
\APACrefYearMonthDay{2016}{}{}.
\newblock
{\BBOQ}\APACrefatitle {{Visual} {Instance} {Retrieval} with {Deep}
  {Convolutional} {Networks}} {{Visual} {Instance} {Retrieval} with {Deep}
  {Convolutional} {Networks}}.{\BBCQ}
\newblock
\APACjournalVolNumPages{ITE Transactions on Media Technology and
  Applications}{4}{}{251--258}.
\newblock
\begin{APACrefDOI}\doi{https://doi.org/10.3169/mta.4.251}\end{APACrefDOI}
\PrintBackRefs{\CurrentBib}

\bibitem [\protect \citeauthoryear {%
Ronneberger%
, Fischer%
\BCBL {}\ \BBA {} Brox%
}{%
Ronneberger%
\ \protect \BOthers {.}}{%
{\protect \APACyear {2015}}%
}]{%
ronneberger2015u}
\APACinsertmetastar {%
ronneberger2015u}%
\begin{APACrefauthors}%
Ronneberger, O.%
, Fischer, P.%
\BCBL {} Brox, T.%
\end{APACrefauthors}%
\unskip\
\newblock
\APACrefYearMonthDay{2015}{}{}.
\newblock
{\BBOQ}\APACrefatitle {U-{Net}: {Convolutional} {Networks} for {Biomedical}
  {Image} {Segmentation}} {U-{Net}: {Convolutional} {Networks} for {Biomedical}
  {Image} {Segmentation}}.{\BBCQ}
\newblock
 \APACrefbtitle {International Conference on Medical Image Computing and
  Computer Assisted Intervention ({MICCAI}).} {International conference on
  medical image computing and computer assisted intervention ({MICCAI}).}
\newblock
\begin{APACrefDOI}\doi{https://doi.org/10.1007/978-3-319-24574-4_28}\end{APACrefDOI}
\PrintBackRefs{\CurrentBib}

\bibitem [\protect \citeauthoryear {%
S\'anchez%
, Perronnin%
, Mensink%
\BCBL {}\ \BBA {} Verbeek%
}{%
S\'anchez%
\ \protect \BOthers {.}}{%
{\protect \APACyear {2013}}%
}]{%
sanchez2013image}
\APACinsertmetastar {%
sanchez2013image}%
\begin{APACrefauthors}%
S\'anchez, J.%
, Perronnin, F.%
, Mensink, T.%
\BCBL {} Verbeek, J.%
\end{APACrefauthors}%
\unskip\
\newblock
\APACrefYearMonthDay{2013}{}{}.
\newblock
{\BBOQ}\APACrefatitle {Image {Classification} with the {Fisher} {Vector}:
  {Theory} and {Practice}} {Image {Classification} with the {Fisher} {Vector}:
  {Theory} and {Practice}}.{\BBCQ}
\newblock
\APACjournalVolNumPages{International Journal of Computer
  Vision}{105}{}{222--245}.
\newblock
\begin{APACrefDOI}\doi{https://doi.org/10.1007/s11263-013-0636-x}\end{APACrefDOI}
\PrintBackRefs{\CurrentBib}

\bibitem [\protect \citeauthoryear {%
Schneider%
, Taylor%
\BCBL {}\ \BBA {} Kremer%
}{%
Schneider%
\ \protect \BOthers {.}}{%
{\protect \APACyear {2020}}%
}]{%
schneider2020similarity}
\APACinsertmetastar {%
schneider2020similarity}%
\begin{APACrefauthors}%
Schneider, S.%
, Taylor, G.%
\BCBL {} Kremer, S.%
\end{APACrefauthors}%
\unskip\
\newblock
\APACrefYearMonthDay{2020}{}{}.
\newblock
{\BBOQ}\APACrefatitle {Similarity {Learning} {Networks} for {Animal}
  {Individual} {Re}-{Identification} - {Beyond} the {Capabilities} of a {Human}
  {Observer}} {Similarity {Learning} {Networks} for {Animal} {Individual}
  {Re}-{Identification} - {Beyond} the {Capabilities} of a {Human}
  {Observer}}.{\BBCQ}
\newblock
 \APACrefbtitle {{Winter} {Applications} of {Computer} {Vision} {Workshops}
  ({WACVW}).} {{Winter} {Applications} of {Computer} {Vision} {Workshops}
  ({WACVW}).}
\newblock
\begin{APACrefDOI}\doi{https://doi.org/10.1109/WACVW50321.2020.9096925}\end{APACrefDOI}
\PrintBackRefs{\CurrentBib}

\bibitem [\protect \citeauthoryear {%
Schneider%
, Taylor%
, Linquist%
\BCBL {}\ \BBA {} Kremer%
}{%
Schneider%
\ \protect \BOthers {.}}{%
{\protect \APACyear {2019}}%
}]{%
schneider2019past}
\APACinsertmetastar {%
schneider2019past}%
\begin{APACrefauthors}%
Schneider, S.%
, Taylor, G.W.%
, Linquist, S.%
\BCBL {} Kremer, S.C.%
\end{APACrefauthors}%
\unskip\
\newblock
\APACrefYearMonthDay{2019}{}{}.
\newblock
{\BBOQ}\APACrefatitle {Past, present and future approaches using computer
  vision for animal re-identification from camera trap data} {Past, present and
  future approaches using computer vision for animal re-identification from
  camera trap data}.{\BBCQ}
\newblock
\APACjournalVolNumPages{Methods in Ecology and Evolution}{10}{}{461--470}.
\newblock
\begin{APACrefDOI}\doi{https://doi.org/10.1111/2041-210x.13133}\end{APACrefDOI}
\PrintBackRefs{\CurrentBib}

\bibitem [\protect \citeauthoryear {%
Sch\"olkopf%
, Smola%
\BCBL {}\ \BBA {} Müller%
}{%
Sch\"olkopf%
\ \protect \BOthers {.}}{%
{\protect \APACyear {1998}}%
}]{%
scholkopf_nonlinear_1998}
\APACinsertmetastar {%
scholkopf_nonlinear_1998}%
\begin{APACrefauthors}%
Sch\"olkopf, B.%
, Smola, A.%
\BCBL {} Müller, K\BHBI R.%
\end{APACrefauthors}%
\unskip\
\newblock
\APACrefYearMonthDay{1998}{}{}.
\newblock
{\BBOQ}\APACrefatitle {{Nonlinear Component Analysis as a Kernel Eigenvalue
  Problem}} {{Nonlinear Component Analysis as a Kernel Eigenvalue
  Problem}}.{\BBCQ}
\newblock
\APACjournalVolNumPages{Neural Computation}{10}{}{1299--1319}.
\newblock
\begin{APACrefDOI}\doi{https://doi.org/10.1162/089976698300017467}\end{APACrefDOI}
\PrintBackRefs{\CurrentBib}

\bibitem [\protect \citeauthoryear {%
Sivic%
\ \BBA {} Zisserman%
}{%
Sivic%
\ \BBA {} Zisserman%
}{%
{\protect \APACyear {2003}}%
}]{%
sivic2003video}
\APACinsertmetastar {%
sivic2003video}%
\begin{APACrefauthors}%
Sivic%
\BCBT {}\ \BBA {} Zisserman.%
\end{APACrefauthors}%
\unskip\
\newblock
\APACrefYearMonthDay{2003}{}{}.
\newblock
{\BBOQ}\APACrefatitle {{Video Google: a text retrieval approach to object
  matching in videos}} {{Video Google: a text retrieval approach to object
  matching in videos}}.{\BBCQ}
\newblock
 \APACrefbtitle {{International} {Conference} on {Computer} {Vision} ({ICCV}).}
  {{International} {Conference} on {Computer} {Vision} ({ICCV}).}
\newblock
\begin{APACrefDOI}\doi{https://doi.org/10.1109/ICCV.2003.1238663}\end{APACrefDOI}
\PrintBackRefs{\CurrentBib}

\bibitem [\protect \citeauthoryear {%
Smeulders%
, Worring%
, Santini%
, Gupta%
\BCBL {}\ \BBA {} Jain%
}{%
Smeulders%
\ \protect \BOthers {.}}{%
{\protect \APACyear {2000}}%
}]{%
smeulders2000content}
\APACinsertmetastar {%
smeulders2000content}%
\begin{APACrefauthors}%
Smeulders, A.%
, Worring, M.%
, Santini, S.%
, Gupta, A.%
\BCBL {} Jain, R.%
\end{APACrefauthors}%
\unskip\
\newblock
\APACrefYearMonthDay{2000}{}{}.
\newblock
{\BBOQ}\APACrefatitle {Content-based image retrieval at the end of the early
  years} {Content-based image retrieval at the end of the early years}.{\BBCQ}
\newblock
\APACjournalVolNumPages{IEEE Transactions on Pattern Analysis and Machine
  Intelligence}{22}{}{1349--1380}.
\newblock
\begin{APACrefDOI}\doi{https://doi.org/10.1109/34.895972}\end{APACrefDOI}
\PrintBackRefs{\CurrentBib}

\bibitem [\protect \citeauthoryear {%
Thompson%
\ \protect \BOthers {.}}{%
Thompson%
\ \protect \BOthers {.}}{%
{\protect \APACyear {2019}}%
}]{%
thompson2019finfindr}
\APACinsertmetastar {%
thompson2019finfindr}%
\begin{APACrefauthors}%
Thompson, J.%
, Zero, V.%
, Schwacke, L.%
, Speakman, T.%
, Quigley, B.%
, Morey, J.%
\BCBL {} McDonald, T.%
\end{APACrefauthors}%
\unskip\
\newblock
\APACrefYearMonthDay{2019}{}{}.
\newblock
{\BBOQ}\APACrefatitle {{finFindR: Computer-assisted Recognition and
  Identification of Bottlenose Dolphin Photos in R}} {{finFindR:
  Computer-assisted Recognition and Identification of Bottlenose Dolphin Photos
  in R}}.{\BBCQ}
\newblock
\APACjournalVolNumPages{bioRxiv}{}{}{825661}.
\newblock
\begin{APACrefDOI}\doi{https://doi.org/10.1101/825661}\end{APACrefDOI}
\PrintBackRefs{\CurrentBib}

\bibitem [\protect \citeauthoryear {%
Titterington%
, Afm%
, Smith%
, Makov%
\BCBL {}\ \protect \BOthers {.}}{%
Titterington%
\ \protect \BOthers {.}}{%
{\protect \APACyear {1985}}%
}]{%
titterington1985statistical}
\APACinsertmetastar {%
titterington1985statistical}%
\begin{APACrefauthors}%
Titterington, D.M.%
, Afm, S.%
, Smith, A.F.%
, Makov, U.%
\BCBL {}\ \BOthersPeriod {.}\end{APACrefauthors}%
\unskip\
\newblock
\APACrefYear{1985}.
\newblock
\APACrefbtitle {Statistical Analysis of Finite Mixture Distributions}
  {Statistical analysis of finite mixture distributions}.
\newblock
\APACaddressPublisher{}{John Wiley \& Sons Incorporated}.
\PrintBackRefs{\CurrentBib}

\bibitem [\protect \citeauthoryear {%
Tolias%
, Sicre%
\BCBL {}\ \BBA {} J{\'e}gou%
}{%
Tolias%
\ \protect \BOthers {.}}{%
{\protect \APACyear {2016}}%
}]{%
tolias2015particular}
\APACinsertmetastar {%
tolias2015particular}%
\begin{APACrefauthors}%
Tolias, G.%
, Sicre, R.%
\BCBL {} J{\'e}gou, H.%
\end{APACrefauthors}%
\unskip\
\newblock
\APACrefYearMonthDay{2016}{}{}.
\newblock
{\BBOQ}\APACrefatitle {{Particular Object Retrieval With Integral Max-Pooling
  of CNN Activations}} {{Particular Object Retrieval With Integral Max-Pooling
  of CNN Activations}}.{\BBCQ}
\newblock
 \APACrefbtitle {International Conference on Learning Representations
  ({ICLR}).} {International conference on learning representations ({ICLR}).}
\PrintBackRefs{\CurrentBib}

\bibitem [\protect \citeauthoryear {%
Vidal%
, Wolf%
, Rosenberg%
, Harris%
\BCBL {}\ \BBA {} Mathis%
}{%
Vidal%
\ \protect \BOthers {.}}{%
{\protect \APACyear {2021}}%
}]{%
vidal2021perspectives}
\APACinsertmetastar {%
vidal2021perspectives}%
\begin{APACrefauthors}%
Vidal, M.%
, Wolf, N.%
, Rosenberg, B.%
, Harris, B.%
\BCBL {} Mathis, A.%
\end{APACrefauthors}%
\unskip\
\newblock
\APACrefYearMonthDay{2021}{}{}.
\newblock
{\BBOQ}\APACrefatitle {{Perspectives on Individual Animal Identification from
  Biology and Computer Vision}} {{Perspectives on Individual Animal
  Identification from Biology and Computer Vision}}.{\BBCQ}
\newblock
\APACjournalVolNumPages{Integrative and Comparative Biology}{61}{}{900-916}.
\newblock
\begin{APACrefDOI}\doi{https://doi.org/10.1093/icb/icab107}\end{APACrefDOI}
\PrintBackRefs{\CurrentBib}

\bibitem [\protect \citeauthoryear {%
Weideman%
\ \protect \BOthers {.}}{%
Weideman%
\ \protect \BOthers {.}}{%
{\protect \APACyear {2017}}%
}]{%
weideman2017integral}
\APACinsertmetastar {%
weideman2017integral}%
\begin{APACrefauthors}%
Weideman, H.J.%
, Jablons, Z.M.%
, Holmberg, J.%
, Flynn, K.%
, Calambokidis, J.%
, Tyson, R.B.%
\BDBL {}others%
\end{APACrefauthors}%
\unskip\
\newblock
\APACrefYearMonthDay{2017}{}{}.
\newblock
{\BBOQ}\APACrefatitle {Integral curvature representation and matching
  algorithms for identification of dolphins and whales} {Integral curvature
  representation and matching algorithms for identification of dolphins and
  whales}.{\BBCQ}
\newblock
 \APACrefbtitle {{International} {Conference} on {Computer} {Vision} {Workshop}
  ({ICCVW}).} {{International} {Conference} on {Computer} {Vision} {Workshop}
  ({ICCVW}).}
\newblock
\begin{APACrefDOI}\doi{https://doi.org/10.1109/iccvw.2017.334}\end{APACrefDOI}
\PrintBackRefs{\CurrentBib}

\bibitem [\protect \citeauthoryear {%
Yeleshetty%
, Spreeuwers%
\BCBL {}\ \BBA {} Li%
}{%
Yeleshetty%
\ \protect \BOthers {.}}{%
{\protect \APACyear {2020}}%
}]{%
yeleshetty20203d}
\APACinsertmetastar {%
yeleshetty20203d}%
\begin{APACrefauthors}%
Yeleshetty, D.%
, Spreeuwers, L.%
\BCBL {} Li, Y.%
\end{APACrefauthors}%
\unskip\
\newblock
\APACrefYearMonthDay{2020}{}{}.
\newblock
{\BBOQ}\APACrefatitle {{3D Face Recognition For Cows}} {{3D Face Recognition
  For Cows}}.{\BBCQ}
\newblock
 \APACrefbtitle {{International Conference of the Biometrics Special Interest
  Group (BIOSIG)}.} {{International Conference of the Biometrics Special
  Interest Group (BIOSIG)}.}
\PrintBackRefs{\CurrentBib}

\bibitem [\protect \citeauthoryear {%
Yu%
\ \protect \BOthers {.}}{%
Yu%
\ \protect \BOthers {.}}{%
{\protect \APACyear {2021}}%
}]{%
yu2021ap}
\APACinsertmetastar {%
yu2021ap}%
\begin{APACrefauthors}%
Yu, H.%
, Xu, Y.%
, Zhang, J.%
, Zhao, W.%
, Guan, Z.%
\BCBL {} Tao, D.%
\end{APACrefauthors}%
\unskip\
\newblock
\APACrefYearMonthDay{2021}{}{}.
\newblock
{\BBOQ}\APACrefatitle {{AP}-10K: A Benchmark for Animal Pose Estimation in the
  Wild} {{AP}-10k: A benchmark for animal pose estimation in the wild}.{\BBCQ}
\newblock
 \APACrefbtitle {{Conference on Neural Information Processing Systems (NeurIPS)
  Datasets and Benchmarks Track}.} {{Conference on Neural Information
  Processing Systems (NeurIPS) Datasets and Benchmarks Track}.}
\PrintBackRefs{\CurrentBib}

\bibitem [\protect \citeauthoryear {%
Zavialkin%
}{%
Zavialkin%
}{%
{\protect \APACyear {2020}}%
}]{%
zavialkin2020cnn}
\APACinsertmetastar {%
zavialkin2020cnn}%
\begin{APACrefauthors}%
Zavialkin, D.%
\end{APACrefauthors}%
\unskip\
\newblock
\APACrefYearMonthDay{2020}{}{}.
\newblock
\APACrefbtitle {{CNN-based ringed seal pelage pattern extraction}.} {{CNN-based
  ringed seal pelage pattern extraction}.}
\newblock
\APAChowpublished {Master's thesis, Lappeenranta-Lahti University of Technology
  LUT, Finland}.
\PrintBackRefs{\CurrentBib}

\bibitem [\protect \citeauthoryear {%
Zhelezniakov%
\ \protect \BOthers {.}}{%
Zhelezniakov%
\ \protect \BOthers {.}}{%
{\protect \APACyear {2015}}%
}]{%
zhelezniakov2015segmentation}
\APACinsertmetastar {%
zhelezniakov2015segmentation}%
\begin{APACrefauthors}%
Zhelezniakov, A.%
, Eerola, T.%
, Koivuniemi, M.%
, Auttila, M.%
, Lev{\"a}nen, R.%
, Niemi, M.%
\BDBL {}K{\"a}lvi{\"a}inen, H.%
\end{APACrefauthors}%
\unskip\
\newblock
\APACrefYearMonthDay{2015}{}{}.
\newblock
{\BBOQ}\APACrefatitle {{Segmentation of Saimaa ringed seals for identification
  purposes}} {{Segmentation of Saimaa ringed seals for identification
  purposes}}.{\BBCQ}
\newblock
 \APACrefbtitle {{International Symposium on Visual Computing (ISVC)}.}
  {{International Symposium on Visual Computing (ISVC)}.}
\newblock
\begin{APACrefDOI}\doi{https://doi.org/10.1007/978-3-319-27863-6_21}\end{APACrefDOI}
\PrintBackRefs{\CurrentBib}

\end{thebibliography}


\end{document}